\PassOptionsToPackage{table,prologue}{xcolor}
\documentclass[acmsmall, screen]{acmart}
\usepackage{siunitx}
\usepackage{subcaption}
\usepackage{multirow}
\usepackage{tcolorbox}
\usepackage{fancybox}
\usepackage[inline]{enumitem}

\usepackage{listings}

\AtBeginDocument{%
  \providecommand\BibTeX{{%
    \normalfont B\kern-0.5em{\scshape i\kern-0.25em b}\kern-0.8em\TeX}}}

\begin{document}

\title[System Safety Monitoring of Learned Components Using Temporal Metric Forecasting]{System Safety Monitoring of Learned Components Using Temporal Metric Forecasting}

\author{Sepehr Sharifi}
\email{s.sharifi@uottawa.ca}
\orcid{0000-0002-2088-9930}
\affiliation{%
  \institution{EECS, University of Ottawa}
  \streetaddress{800 King Edward}
  \city{Ottawa}
  \country{Canada}
  \postcode{K1N 6N5}
  }
\author{Andrea Stocco}
\email{andrea.stocco@tum.de}
\orcid{0000-0001-8956-3894}
\affiliation{%
  \institution{Technical University of Munich}
  \streetaddress{Boltzmannstra{\ss}e 3}
  \city{Munich}
  \country{Germany}
  \postcode{85748}
}
\affiliation{%
  \institution{fortiss GmbH}
  \streetaddress{Guerickestra{\ss}e 25}
  \city{Munich}
  \state{Bayern}
  \country{Germany}
  \postcode{80805}
}
\author{Lionel C. Briand}
\orcid{0000-0002-1393-1010}
\email{lbriand@uottawa.ca}
\affiliation{%
  \institution{University of Ottawa, Canada, and Lero SFI Centre for Software Research, University of Limerick}
  \streetaddress{Tierney building}
  \city{Limerick}
  \country{Ireland}
  \postcode{V94 NYD3}
}

\renewcommand{\shortauthors}{Sharifi et al.}

\begin{abstract}
In learning-enabled autonomous systems, safety monitoring of learned components is crucial to ensure their outputs do not lead to system safety violations, given the operational context of the system. However, developing a safety monitor for practical deployment in real-world applications is challenging. This is due to limited access to internal workings and training data of the learned component. Furthermore, safety monitors should predict safety violations with low latency, while consuming a reasonable computation resource amount.

To address the challenges, we propose a safety monitoring method based on probabilistic time series forecasting. Given the learned component outputs and an operational context, we empirically investigate different Deep Learning (DL)-based probabilistic forecasting to predict the objective measure capturing the satisfaction or violation of a safety requirement (\emph{safety metric}).
We empirically evaluate safety metric and violation prediction accuracy, and inference latency and resource usage of four state-of-the-art models, with varying horizons, using autonomous aviation
and autonomous driving
case studies.
Our results suggest that probabilistic forecasting of safety metrics, given learned component outputs and scenarios, is effective for safety monitoring.
Furthermore, for both case studies, the Temporal Fusion Transformer (TFT) was the most accurate model for predicting imminent safety violations, with acceptable latency and resource consumption.
\end{abstract}

\begin{CCSXML}
<ccs2012>
   <concept>
       <concept_id>10011007.10010940.10011003.10011114</concept_id>
       <concept_desc>Software and its engineering~Software safety</concept_desc>
       <concept_significance>500</concept_significance>
       </concept>
   <concept>
       <concept_id>10010147.10010178</concept_id>
       <concept_desc>Computing methodologies~Artificial intelligence</concept_desc>
       <concept_significance>300</concept_significance>
       </concept>
   <concept>
       <concept_id>10010520.10010553.10010554</concept_id>
       <concept_desc>Computer systems organization~Robotics</concept_desc>
       <concept_significance>500</concept_significance>
       </concept>
 </ccs2012>
\end{CCSXML}

\ccsdesc[500]{Software and its engineering~Software safety}
\ccsdesc[300]{Computing methodologies~Artificial intelligence}
\ccsdesc[500]{Computer systems organization~Robotics}

\keywords{ML-enabled Autonomous System, Learned Component, System Safety Monitoring, Probabilistic Time Series Forecasting}

\renewcommand*{\figureautorefname}{Figure}
\renewcommand*{\sectionautorefname}{Section}
\renewcommand*{\subsectionautorefname}{Section}
\renewcommand*{\subsubsectionautorefname}{Section}

\newcommand{\SSS}[1]{{\color{violet} [SS: #1]}}
\newcommand{\AS}[1]{{\color{red} [AS: #1]}}
\newcommand{\LB}[1]{{\color{green} [LB: #1]}}
\newcommand{\Rev}[1]{{\color{blue} #1}}

\maketitle

\section{Introduction}\label{sec:intro}

Autonomous systems are increasingly being empowered using learned components to perform perception, prediction, planning, and control tasks~\cite{SOORI2023RoboticMLReview}.
Since such components' behaviour is learned through training, as opposed to being expressed in source code or specification, ensuring the reliability of such systems through conventional software engineering practices is inadequate.
These risks are particularly acute when autonomous systems are employed in safety-critical applications, e.g., autonomous driving~\cite{nvidia-dave2}, autonomous aviation~\cite{Julian}, medical diagnosis~\cite{zhang2018medical} or disease prediction~\cite{Zhao-nature}, as failures could directly jeopardize human safety.
Recently, methods have been proposed to make reliable, robust, and accurate learned components through novel testing methods~\cite{huang2020survey,zhang2020machine}.
Nevertheless, such components are never perfect and even systems comprised of reliable components are still prone to accidents~\cite{Leveson2012Engineering}. 
For instance, some accidents are caused by unsafe component interactions~\cite{albee2000report,Leveson2012Engineering}.
Thus, the impact of ML components on safety can only be studied in the context of the system they are integrated into and in a specific operational context~\cite{black2009system, Leveson2012Engineering}.

The specialized nature of learned components, i.e., trained on necessarily limited training data, necessitates the use of runtime assurance mechanisms~\cite{skoog2020leveraging}, i.e., safety monitors~\cite{sharifi2023mlcshe}.
Runtime safety monitors observe the system, its operational context, and the inputs and outputs of a component that cannot be fully trusted, such as a learned component, predicting if its outputs may lead the system toward a safety requirement violation.
In such cases, a warning is raised by the safety monitor to prevent the outputs of the learned component from propagating to the rest of the system. For example, safety recovery measures include falling back on a less efficient but trustworthy component~\cite{skoog2020leveraging} or taking pre-designed safety recovery measures, such as an emergency stop in autonomous vehicles (AVs).
Safety monitors must know the operational context of the system to determine whether the component might contribute to a hazard. 
For instance, a misclassification by an AV object detection component can lead to non-hazardous outcomes under certain system contexts, e.g., when an AV misidentifies a horse-drawn carriage in its front as a truck and maintains a safe distance from it.
Thus, runtime {monitoring} of {both the operational context and the }learned components {outputs} is crucial in developing effective safety monitors that can identify transitions of the system from safe to hazard states, which can lead to safety requirement violations.

However, monitoring the impact of a learned component on learning-enabled system safety poses several significant challenges.
First, many safety-critical learning-enabled autonomous systems are developed by system integrators who are developing the system using various components, including learned components, many of which are developed by third parties.
Thus, system integrators often do not have access to the training or test data of the learned component, nor to white-box information such as their architecture or neuron weights.
Second, safety monitors should be able to monitor not only the outputs of the learned component over time but also the operational context, which typically includes static parameters such as weather, and may also include dynamic parameters such as the trajectory of other vehicles in proximity to an AV.
Third, in a safety-critical context, the safety monitor must predict a safety violation early enough to allow the system or a user sufficient time to mitigate it. As such, efficiency is a key requisite, which translates to the necessity of developing monitors that exhibit a low reaction latency and do not exceed the practical limits of onboard computing units, as opposed to cloud-based alternatives.

Currently, existing methods fail to address all of the above challenges as they monitor for learned component mispredicitons, as opposed to system safety violations~\cite{Stocco2020SelfOracle,Stocco2023ThirdEye,2024-Grewal-ICST,Henriksson,deeproad,dissector,hell2021vae}.
Furthermore, many of the proposed methods rely on internal information sources from the learned component~\cite{Hendrycks2017softmax, Ovadia2019uncertainty, Stocco2023ThirdEye}.

To address the above challenges, we propose a safety monitoring method based on the idea of predicting the near-future values of a safety metric, i.e., the objective measure used to determine the satisfaction or violation of a safety requirement~\cite{asaadi2020UA2}, given the history of learned component outputs and the operational context of the system.

Given the safety-criticality of learning-enabled autonomous systems, where the cost of not predicting safety violations at runtime is very high, instead of relying on single forecast values for each timestep, our method predicts the probability distribution of the safety metric and relies on its tail-end values to conservatively predict safety violations.
We leverage Deep Learning (DL) based probabilistic time series forecasters and empirically evaluate state-of-the-art models in terms of prediction accuracy as well as average inference latency and runtime computation resource usage.

\paragraph{Contributions}
The contributions of this paper are as follows:
\begin{itemize}
    \item A safety monitoring method that leverages time series forecasting of a safety metric to identify learned component behavior and system context that lead to system safety violations at runtime.
    \item {An application of the safety monitoring method to widely used case studies in autonomous aviation~\cite{asaadi2019UA1, pasareanu2023act, katz2022verification, Kadron2022act, Asaadi2020ACT}, i.e., Autonomous Centerline Tracking (ACT), and in autonomous driving, i.e., an Autonomous Driving System (ADS)~\cite{Stocco2023ThirdEye}, including a dataset generated from system-in-the-loop simulations.}
    \item A large-scale empirical evaluation ({$7500+$ GPU hours and $42$ calendar days of computation}) with state-of-the-art DL-based probabilistic forecasting models targeting safety metric and safety violation prediction accuracy, inference latency, and runtime resource usage.
\end{itemize}

\paragraph{Key Findings.}
The key findings of our empirical evaluation as follows:
\begin{itemize}
    \item Overall, the results of our study suggest that probabilistic forecasting of safety metrics, given learned component outputs and scenarios, is effective for safety monitoring.
    \item For our ACT case study, DL-based probabilistic forecasting methods, especially those with sequence-to-sequence architecture, yield low inference latency while consuming feasible computing resources in terms of model size and peak memory usage during inference.
    \item Using Temporal Fusion Transformers (TFT) for predicting \emph{imminent} safety violations---where the hazard forecast horizon is equal to the minimum reaction time, for all lookback horizons---leads to the most accurate predictions with acceptable inference latency and reasonable computational resource usage.
\end{itemize}

\paragraph{Paper Structure}
\autoref{sec:background} provides the necessary background on time series forecasting models and the main DL-based architectures.
\autoref{sec:prob-chall} formally defines the safety metric forecasting problem and details its challenges.
\autoref{sec:related-work} discusses related work.
\autoref{sec:solution} presents our proposed safety monitoring method in detail.
\autoref{sec:evaluation} provides an empirical evaluation of our method and discusses the results.
\autoref{sec:conclusion} concludes the paper and suggests future directions for research and improvement.

\section{Background}\label{sec:background}

In this section, we discuss the main characteristics of time series forecasting methods as well as the main Deep Learning (DL)-based time series forecasting architectures.

\subsection{Time-series Forecasting}
\label{sec:background-tsf}
Time series forecasting aims at predicting the future values of a time series.
As described in Januschowski et al.~\cite{JANUSCHOWSKI2020criteria}, we can distinguish among forecasting methods along a number of dimensions such as \emph{global vs. local},
\emph{probablistic vs. point},
\emph{computational complexity and costs},
and \emph{data-driven vs. model-based}.

\paragraph{Global vs. Local Forecasting}
Local methods involve estimating model parameters independently for each time series, while global methods estimate parameters jointly using all available time series~\cite{JANUSCHOWSKI2020criteria,BENIDIS2022DL4TS}.
This distinction is concerned with how model parameters are estimated and does not necessarily imply a specific dependency structure between the time series.
For instance, a global model can still assume independence between forecasts for different time series for computational efficiency reasons, even though it estimates parameters jointly~\cite{JANUSCHOWSKI2020criteria}.
While traditional statistical methods often adopt local approaches, global methods have been utilized in both the statistics and machine learning (ML) communities.
{Recent trends show that Deep Neural Networks (DNNs) which are trained as global models, surpass all forecasting models when used as local models~\cite{BENIDIS2022DL4TS,makridakis2023forecasting}.
}

\paragraph{Probabilistic vs. Point Forecasting}
Forecasting techniques can also be broadly categorized into probabilistic and point forecasting methods. 
While point forecasts offer a single best prediction, probabilistic forecasting methods quantify predictive uncertainty, allowing decision-makers to consider this uncertainty when using the forecast~\cite{JANUSCHOWSKI2020criteria}.
{
For a safety-critical application, e.g., predicting whether a system will experience a safety violation in the future and taking recovery actions, it is vital to be able to take the uncertainty associated with the forecasts into account.
For instance, one can use the tail-end values of the predicted probability distribution to take into account the worst-case predictions as a basis for decision making.
}

Methods for handling uncertainty include Bayesian approaches
and frequentist approaches like model ensembles and bootstrap sampling~\cite{BENIDIS2022DL4TS}.
The use of Bayesian approaches in estimating parameter and model uncertainty is well studied in ML literature (for introductory and recent work references refer to the study by \citet{JANUSCHOWSKI2020criteria}).
The predictive uncertainty for a time series is fully described by the predictive distribution, but probabilistic forecasting methods differ in how they enable users to access this distribution, often providing pointwise predictive intervals or Monte Carlo sample paths~\cite{BENIDIS2022DL4TS}.
Some methods assume a parametric form of the distribution and return its parameters~\cite{BENIDIS2022DL4TS, SALINAS2020DeepAR}.
Modern ML methods handle uncertainty by
estimating quantile functions directly~\cite{JANUSCHOWSKI2020criteria}.
The results of the M4 competition have demonstrated the accuracy of prediction intervals obtained from ML methods, even though they may lack theoretical underpinnings~\cite{JANUSCHOWSKI2020criteria, MAKRIDAKIS202054}.
This highlights the effectiveness of ML approaches in handling uncertainty in forecasting.

\paragraph{Data-driven vs. Model-based Forecasting}
Methods commonly associated with machine learning, such as deep neural networks, are characterized by their data-driven nature.
These approaches excel at capturing intricate patterns from data without relying on strong structural assumptions.
However, their flexibility comes at the cost of requiring large amounts of data to effectively tune the multitude of parameters they possess.
For instance, recurrent neural networks (RNNs) can discern complex nonlinear patterns from data, as exemplified by their ability to predict time series with oscillating variance amplitudes~\cite{JANUSCHOWSKI2020criteria}.
Nevertheless, the risk of overfitting arises due to their capacity to memorize patterns, a challenge that regularization techniques like Dropout~\cite{srivastava2014dropout}, aim to mitigate.
In contrast, statistical models like AutoRegressive Integrated Moving Average (ARIMA) models and Generalized Linear Models (GLMs) are characterized by their parsimonious parameterization and reliance on assumptions to model patterns~\cite{box2015time}.
These models require less data to accurately estimate their parameters but are inherently more rigid due to the limitations imposed by their structural assumptions~\cite{JANUSCHOWSKI2020criteria}.
Furthermore, a study by \citet{kolassa2016} shows that simpler models can sometimes outperform complex, correctly specified ones, showcasing the intricacies of model-driven approaches~\cite{JANUSCHOWSKI2020criteria}.
Furthermore, model-driven approaches require meticulous feature engineering and model specification.
Conversely, data-driven models are often preferred for forecasting tasks that involve a large number of time series, from which complex patterns can be extracted~{\cite{makridakis2023forecasting}}.
{Moreover, DL-based forecasting models can often be trained on large datasets without the need for problem-specific feature engineering~\cite{JANUSCHOWSKI2020criteria}.}

\subsection{DL-based Forecasting Architectures}

DL-based forecasting models can be categorized into two main categories of architectures, namely \emph{iterative} and \emph{sequence-to-sequence}.
The iterative architecture generates forecasts step by step, where the model predicts a one-time step based on the previous hidden state and current available information~\cite{BENIDIS2022DL4TS}.
The process is repeated until the desired forecast horizon is reached.
Iterative models can easily be applied to any forecast horizon length.
However, since the generated forecast at each time step has an error, the recursive structure of iterative models can potentially lead to large errors being accumulated over long forecast horizons~\cite{lim2021time}. 
RNN models such as long short-term memory networks (LSTMs) and gated recurrent units (GRUs) are commonly employed in iterative architectures~\cite{BENIDIS2022DL4TS}.
On the other hand, sequence-to-sequence architectures operate by mapping an input sequence to an output sequence, potentially of different lengths.
This architecture consists of two main components: an encoder and a decoder.
The encoder transforms the input sequence into a fixed-size context vector, which is then used by the decoder to generate the output sequence of a predetermined length.
A typical training instance in this approach includes the target and covariate ({static and time-series features or embeddings}~\cite{lim2021time}) values up to a specific time point \textit{t} as input, while the neural network generates a set number of target values beyond time \textit{t}.

\section{Problem and Challenges}\label{sec:prob-chall}

In this section we cast the learned component safety monitoring problem as a safety metric forecasting problem and discuss its challenges.

\subsection{Problem Definition}\label{sec:problem}

Safety-critical systems such as autonomous vehicles (AVs) or Unmanned Aircraft Systems (UASs) use learned components such as Deep Neural Networks (DNNs) to automate and inform perception, localization, and planning tasks.
In this paper, we use as a running example an Autonomous Centerline Tracking (ACT) software, which is used to ensure accurate and safe UAS taxiing on a runway, by detecting and following certain reference points or a designated path without human intervention.
The distance between the system position to such reference point or centerline is called Centerline Track Error ($cte$).
The ACT uses a DNN to estimate the $cte$ from camera images and steer the physical system, e.g., a vehicle or an aircraft, towards the centerline where $cte=0$.

During its operation, the system must satisfy certain safety requirements such as {\itshape ``the system shall stay within 5 meters of the centerline''}. Although the learned component and the system have to be thoroughly tested and validated before going into operation, during certain challenging or unexpected execution scenarios, the learned component could contribute to the system violating the safety requirement, with potentially life-critical consequences.
Therefore, early run-time prediction of a safety violation is an important endeavor and a prerequisite for developing fallback measures and mitigation strategies~\cite{skoog2020leveraging}, that include blocking the output of the learned component from being broadcast throughout the rest of the system.
To measure the degree of satisfaction or violation of a safety requirement, safety metrics are used.
For example, from the above requirement, the safety metric can be defined as the difference between the actual $cte$ (measured by calculating the difference between the system and centerline GPS locations) of the system and a maximum safe $cte$ threshold of \SI{5}{m}.
Note that the safety metric value varies over time and is therefore calculated at each time step.

More concretely, let $s$ be the ACT system including the learned component $m$ for image-based $cte$ estimation, operating in its environment under an operational scenario $x$. The latter is represented by static and dynamic properties that exist during system operation, e.g., the angle of the sun, cloud cover, runway properties, or the initial position of the aircraft. 
For each time step $t$, the system takes an input $in_{s,t}$ from the camera, thus capturing the state of the environment, and provides a pre-processed (e.g., by drivers or information fusion) image $in_{m,t}$ ready to be consumed by $m$.
$m$ produces a real number $out_{m,t}$ which represents the $cte$ estimate. $s$ processes $out_{m,t}$, generates a steering command $out_{s,t}$, and applies it to the system. 
The state of the environment relative to $s$ changes based on $out_{s,t}$,
and the entire process repeats during the operation of $s$, whereas the next learned component inputs are partially determined by previously learned component outputs.

For a safety requirement $r$ (e.g., \textit{``the system shall not deviate from the centerline more than \SI{5}{m}''}), we can measure the degree of safety violation of $s$ at time $t$, denoted by $y_{r,t}$, with a continuous function $f_r(t)$ which determines at time $t$ whether $r$ has been violated ($y_{r,t}=f_r(t)\geq0$) or how close it has come to violating it ($y_{r,t}=f_r(t)<0$).
Note, that the exact definition of $f_r$ is context-dependent and varies based on the system and the safety requirement of interest.
For the ACT system and the safety requirement above, we define $f_r$ as denoted in Equation~\ref{eq:sm-fcn-def}.

\begin{equation}
    y_{r,t}=f_r(t)\coloneqq|cte_{act}|-|cte_{thr}|
\label{eq:sm-fcn-def}
\end{equation}

Whereas, $cte_{act}$ and $cte_{thr}$ are the actual and safety violation threshold values of the centerline track error, respectively.
Based on the above context, let 
$x^{(n)}$ be a given set of environmental conditions in the space of all possible conditions, also referred to as \emph{operational scenarios}.
Let
$o_{m,t-k:t}^{(n)}$
and $y_{t-k:t}^{(n)}$ be the sequence of observed $m$'s
outputs
and safety metric values from time $t-k$ to $t$ (where $k$ denotes the \textit{lookback} horizon), given $x^{(n)}$.

Given a \textit{hazard forecast horizon} $h$\footnote{{Hazard forecast horizon is the number of timesteps in the future~\cite{Haben2023TSConcepts}, over which we want to predict the values of a safety metric such that safety violations (\emph{hazards}) can be predicted and mitigated or avoided.}}
, we want to predict the sequence of safety metric values from time $t+1$ to $t+h$, i.e., $\hat{y}_{t+1:t+h}^{(n)}$, using a prediction model $g$, as expressed in \autoref{eq:prob-def}, as accurately as possible.

\begin{equation}
    \hat{y}_{t+1:t+h}^{(n)}=g(h,y_{t-k:t}^{(n)},o_{m,t-k:t}^{(n)},x^{(n)})
\label{eq:prob-def}
\end{equation}

{As mentioned in \autoref{sec:intro}, aside from the ACT system mentioned above, this paper additionally targets a second cases study, i.e., an Autonomous Driving System (ADS) which performs the \emph{lane keeping} functionality autonomously, while relying only on image inputs from the camera.
The formal problem definition of the problem provided in this section, especially \autoref{eq:prob-def}, equally applies to the ADS case study.
We provide, in \autoref{sec:eval-subj}, complete details for both the ACT and ADS case studies evaluated in this paper.
}

\subsection{Challenges}\label{sec:challenges}

Given the context and the problem definition provided in \autoref{sec:problem}, we observe multiple challenges.
First, the development of learned components is often outsourced to third parties{~\cite{statista2023artificial, renieris2023building}}, which are later integrated into the main autonomous system. Thus, the limited or lack of access to the learned component details inhibits the application of white-box methods for safety monitoring~\cite{Stocco2020SelfOracle}.
Such details include the training data and the model's architecture, weights, activation patterns, or gradients during the feed-forward pass.

Second, as mentioned in \autoref{sec:intro}, evaluating the safety of a learning-enabled autonomous system relies both on the \emph{static} operational context data (\emph{scenario}) and \emph{dynamic} time-series data related to the behavior of the learned component and the history of safety metric values over time.
Thus, the safety monitor should be able to utilize both types of data to provide an accurate forecast of the safety metric values over the hazard forecast horizon.

Third, safety monitors are often developed for safety-critical cyber-physical systems with limited computation capabilities. Thus, it is paramount that the safety monitor introduces low latency and memory overhead to the system.
Although many safety monitoring methods that rely on white-box confidence estimation techniques~\cite{Stocco2023ThirdEye,2024-Grewal-ICST} are more accurate than their black-box counterparts, their memory and computing overhead makes their adoption in resource-constrained settings impractical~\cite{https://doi.org/10.1002/smr.2386}.

To address the above challenges, {henceforth denoted C1 through C3, respectively,} in this paper we evaluate time series forecasting methods, especially the ones based on DL, to forecast the safety metric values of a learning-enabled system over the hazard forecast horizon. 
DL methods have been shown to provide accurate forecasts while being amenable to multiple data types (static and time series) and a large number of samples~\cite{BENIDIS2022DL4TS, makridakis2023forecasting, JANUSCHOWSKI2020criteria}.
In \autoref{sec:solution}, we provide further details on the proposed safety metric forecasting solutions, whereas in \autoref{sec:evaluation} we discuss the experimental evaluation of different state-of-the-art forecasting models for the ACT case study.

\section{Related Work}\label{sec:related-work}

This section discusses existing studies related to the problem of learned component safety monitoring.
Some surveys~\cite{Rahman2021SM-Survey, mohseni2021practical, Bogdoll_2022_CVPR, luo2021survey},
distinguish between
Out-Of-Distribution (OOD) detection and uncertainty estimation (quantification) methods.
The former focuses on identifying learned component inputs that are not within its training distribution, while the latter estimates the uncertainty associated with the learned component outputs.
Since these methods aim, at a high level, at a similar goal, i.e., identifying inputs that lead to uncertain and thus untrustworthy outputs, they should therefore be discussed here.
Next, we discuss the main safety monitoring techniques in the literature, primarily categorized based on the type of system information access they assume, namely black-box and white-box approaches.

\subsection{Black-box Methods}\label{sec:blackbox-methods}
Black-box methods use information such as learned component inputs and outputs, as well as its training and test datasets,\footnote{The survey conducted by \citet{riccio2020testing} categorizes the methods that require access to learned component train and test dataset as \textit{data-box} methods. However, to avoid confusion, we categorize them as black-box methods here as they do not use any internal information from the model itself.} to identify the shift in the distribution of inputs observed during operation from the training input distribution, which can lead to mispredictions during operation~\cite{zolfagharian2023smarla}.

For example, \citet{Zhang2018DeepRoad} proposed DeepRoad which was mainly designed for testing AV learned components by validating single input images according to their minimum distance from the training set based on the embeddings generated according to VGGNet~\cite{simonyan2014VGGNet} features.
SelfOracle~\cite{Stocco2020SelfOracle} is a black-box failure predictor that uses an autoencoder and time series-based anomaly detection to reconstruct the input images observed by the learned component and to use reconstruction loss to detect OOD inputs.
Similar methods that utilize variational autoencoders (VAEs) to measure an anomaly score have also been proposed by other studies~\cite{hussain2022deepguard,hell2021vae,borg2023ergo}.
DeepGuard, proposed by \citet{hussain2022deepguard}, uses the VAE reconstruction error to prevent roadside collisions with other vehicles,
\citet{borg2023ergo} proposed an OOD detector based on VAEs combined with object detection for an automated emergency braking system.

Moreover, some black-box methods quantify the uncertainty of the learned component outputs,
to help practitioners identify the learned component inputs leading to unreliable outputs,
by estimating probability distributions of the outputs given past system executions.
These methods leverage Bayesian networks or their approximations~\cite{Mackay1992bnn, asaadi2019UA1, gal2016ba}, allowing them to incorporate expert domain knowledge in their Bayesian network models.
In the context of autonomous aviation systems (similar to our ACT example), \citet{asaadi2019UA1} used a non-parametric Bayesian-based uncertainty quantifier, i.e., Gaussian Process (GP) regressor, trained on a subset of the learned component training data, to estimate the uncertainty in learned component outputs given its inputs.

\subsection{White-box Methods}\label{sec:whitebox-methods}

Unlike black-box methods, white-box methods take advantage of internal information sources from the learned component, e.g., model confidence~\cite{Hendrycks2017softmax}, neuron activation patterns~\cite{xiao2021selfchecker} or gradients~\cite{Stocco2023ThirdEye}, comparing their observations at runtime (during the learned component operation) against design-time (during the learned component training).

For example, \citet{Lakshminarayanan2017ensemble1} proposed the use of an ensemble of neural networks (\textit{Deep Ensembles}) to effectively predict the uncertainty of  perception component outputs at runtime. 
\citet{Kendall2017ua} proposed a Bayesian deep learning framework that captures uncertainties associated with both the learned component inputs, also referred to as \textit{aleatoric} uncertainty, as well as the model itself, also known as \textit{epistemic} uncertainty, for a perception component (image segmentation and depth regression).
In the context of autonomous driving, Grewal et al.~\cite{2024-Grewal-ICST} evaluate different uncertainty quantification methods for the misbehavior prediction of failures.
\citet{Hendrycks2017softmax} used the learned component's own confidence, i.e., its softmax probability distributions, to measure uncertainty in learned component outputs.
However, since learned components are prone to generating incorrect outputs (misprediction) with high confidence~\cite{Ovadia2019uncertainty}, many methods have leveraged other information sources to estimate uncertainty.
ThirdEye~\cite{Stocco2023ThirdEye} uses an eXplainable AI (XAI) technique, namely attention maps, to generate a confidence score for the learned component (in this case a DNN) based on input images and gradients of the DNN.
The generated confidence score is then used to predict a failure by comparing it with a failure threshold learned from past system executions (simulations).

\subsection{Limitations of Existing Methods}\label{sec:related-work-shortcomings}
Although the white-box and black-box methods mentioned above
are effective at evaluating the inputs to the learned component, they do not consider the effect of learned component outputs on system safety.
Learned component inputs that can lead to inaccurate outputs (i.e., mispredictions) may not lead to system safety violations, depending on the system's operational context.
As discussed in \autoref{sec:challenges}, a safety monitoring method must be able to predict the combinations of system context and learned component outputs that can lead to system-level safety violations.

In terms of information requirements, methods such as
DeepRoad~\cite{Zhang2018DeepRoad} and the Bayesian method proposed by \citet{asaadi2019UA1}, require access to training and test datasets.
Furthermore, as mentioned in \autoref{sec:whitebox-methods}, to identify safety-violating inputs, white-box methods rely on internal information of the model~\cite{Hendrycks2017softmax, xiao2021selfchecker, Stocco2023ThirdEye}.
As discussed in \autoref{sec:challenges}, system integrators often do not have access to such information, nor training and test datasets, as they are frequently developed by third parties.

Finally, both the black-box and white-box methods discussed in Sections~\ref{sec:blackbox-methods} and~\ref{sec:whitebox-methods}, respectively,
were not evaluated in terms of their inference latency and computation resource usage at runtime~\cite{asaadi2019UA1, Zhang2018DeepRoad, Mackay1992bnn, gal2016ba,Stocco2020SelfOracle,xiao2021selfchecker, Stocco2023ThirdEye}.

Different from the described black-box and white approaches, we evaluate time-series DL methods for safety monitoring, a previously unexplored topic. 
We empirically evaluate the effectiveness of such methods when using both the operational context of the system and learned component behavior while being computationally feasible for runtime monitoring.
We then predict when system context and learned component behavior together lead to system-level safety violations.

\section{Temporal Forecasting of Safety Metrics}\label{sec:solution}

{In this work, as mentioned in \autoref{sec:prob-chall}, we have cast the safety metric prediction problem as a safety metric forecasting problem.}
{Recall that, as described in \autoref{sec:problem} and \autoref{eq:prob-def}, 
the inputs to the forecasting model are 
static operational scenario data ($x$),
dynamic (time-dependent) learned component behavior ($o_{m,t-k:t}$),
and past safety metric data of learning-enabled autonomous systems ($y_{t-k:t})$.
While the output of the forecasting model}
\footnote{Also referred to as \emph{target variable} in time series forecasting literature~\cite{BENIDIS2022DL4TS}.}
{is the safety metric forecasts over the hazard forecast horizon ($\hat{y}_{t+1:t+h}$).
In the rest of this section, we evaluate existing time-series forecasting methods and propose potentially suitable candidates for the problem of predicting the value of the safety metric of a learning-enabled autonomous system over a hazard forecast horizon, as introduced in \autoref{sec:problem}.
Note that the hazard forecast horizon is set by the system developers and safety engineers according to the system properties, its intended mission and its corresponding set of operational contexts, also referred to as
its Operational Design Domain (ODD)~\cite{sharifi2023mlcshe,J3016_202104}.
}

{
We assume that for training the forecasting model, a dataset containing numerous samples from historical system executions under various scenarios has been collected through simulation testing, a common practice for learning-enabled autonomous systems deployed in safety-critical contexts~\cite{lou2022adsSurvey, araujo2023sysreview}.
}
As mentioned in \autoref{sec:background},
model-based time-series forecasting models such as fitting ARIMA models are not suitable for forecasting when the dataset are large, high dimensional, or contains non-linear relationships (between features and target), as the time required to fit them to data considerably increases and their prediction performance degrades. 
Classical machine learning (ML) models, especially tree-based methods (e.g., gradient-boosted tree methods), have been used widely in time series forecasting as they provide superior prediction performance to their statistical counterparts~\cite{BENIDIS2022DL4TS}.
While they can be trained on a large number of samples, these models require substantial feature engineering, thus requiring expert knowledge of the system~\cite{BENIDIS2022DL4TS}, which consumes significant time and effort.

Recently, deep learning time-series forecasting models (DL forecasters) have shown great potential for challenging forecasting problems.  
{As discussed in \autoref{sec:background}, DL forecasters can be trained as global models on large time-series samples without requiring white-box knowledge of the system under test, or specific feature engineering.
Thus, addressing challenge C1 described in \autoref{sec:challenges} by relying only on black-box information related to the system under test.
Moreover, as surveyed by \citet{BENIDIS2022DL4TS}, some DL forecasters can handle samples that contain both static and dynamic data types with complex and non-linear relationships~\cite{BENIDIS2022DL4TS}, addressing challenge C2 stated in \autoref{sec:challenges}.
Last, DL forecasters, given real-valued times series inputs, consume limited computation resources and enable low inference latency, addressing challenge C3 discussed in \autoref{sec:challenges}}
\footnote{We will empirically measure the inference latency as well as the computation resource usage of state-of-the-art DL forecasters in \autoref{sec:evaluation}.}.

Finally, due to the safety-critical nature of learning-enabled systems, it is important to account for the uncertainty associated with predicted safety metric values.
As discussed in \autoref{sec:background}, DL-based probabilistic forecasting methods account for such uncertainty by predicting the values of the safety metric probability distribution.
Knowing the values at the tail-end of the predicted probability distribution of the safety metric allows us to rely on such values for worst-case metric predictions.

Thus, to address the challenges outlined in \autoref{sec:prob-chall}, we propose training a DL-based probabilistic forecaster, given historical execution data of the system and its safety metric, such that it provides forecasts of the safety metric value over the hazard forecast horizon.
Note that the weights of the DL model are selected and remain the same for all the different scenarios and time-series data that are used for training, i.e., it is a \emph{global} model.
Concretely, the DL-based probabilistic forecaster takes the times series of the safety metric values and learned component outputs, as well as scenario parameters as input, and returns time series predictions of the safety metric probability distribution.
Note that the duration of the input time series is equal to the lookback horizon, while the duration of the predicted time series is equal to the forecast horizon.

\paragraph{Safety Violation Prediction}\label{sec:sol-viol-pred}

As discussed earlier in Section~\ref{sec:challenges}, one of the main goals, aside from knowing the value of the safety metric at each timestep, which is crucial for safety-critical decision-making, is to predict safety violations that might occur in the near future, i.e., over the hazard forecast horizon.
Therefore, we further describe how a safety metric forecaster can be used to predict safety violations.

Recall that a safety metric forecaster, at timestep $t$, provides predictions of the safety metric value from timestep $t+1$ to $t+h$.
Further recall that we have assumed that the function measuring the safety metric is defined such that non-negative values imply safety violation, similar to \autoref{eq:sm-fcn-def}.
Thus, if the \emph{maximum} of the predicted safety metric values over the hazard forecast horizon is non-negative, we can say that a safety violation has been predicted.

More specifically, given predicted safety metric values and a hazard forecast horizon $h$, we can define the safety violation function $v(h)$ as detailed in \autoref{eq:viol-pred}.

\begin{equation}
    v(h) = \mathit{sign}\big({max}\{y_{t+1:t+h}\}\big)
\label{eq:viol-pred}
\end{equation}

{Note that the sign function $\mathit{sign}(i)$ used in \autoref{eq:viol-pred} returns $+1$ when $i\geq0$ and returns $-1$ otherwise~\cite{BELLOTTI2014115, cristianini2000svm}. Based on the definition provided in \autoref{eq:viol-pred}, a safety violation is detected over the hazard forecast horizon, i.e., from timestep $t+1$ to $t+h$, if $v(h)=1$.}

\begin{figure*}[t]
\centering
\includegraphics[width=\linewidth]{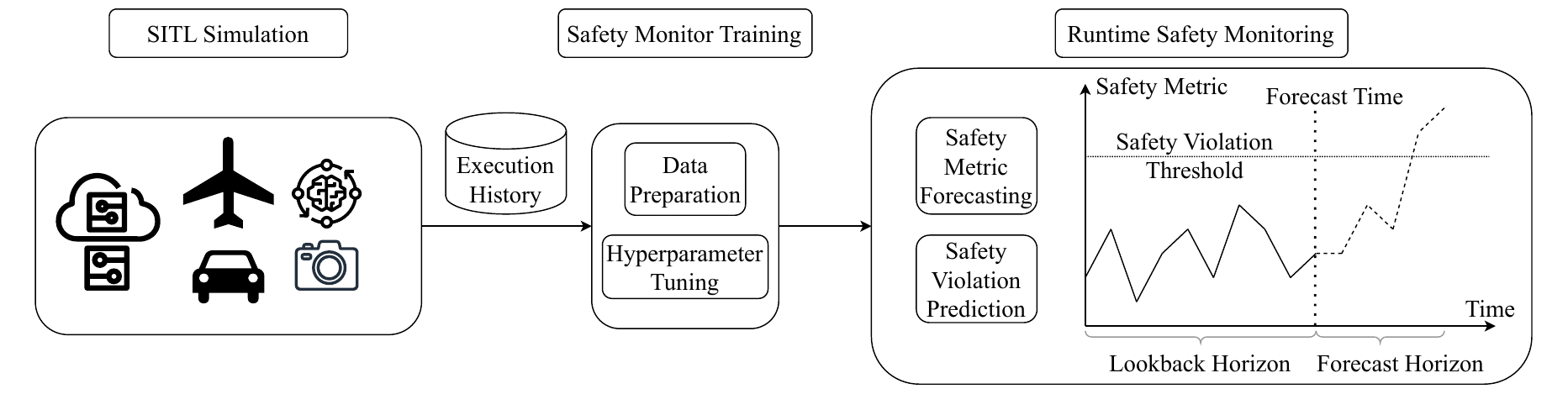}
\Description{A workflow starting with SITL simulation of the AI-enabled system to generate the execution history required for training the safety monitor which entails data preparation and hyperparameter tuning. After training, the safety monitor is used at runtime for safety metric forecasting and safety violation prediction. A diagram for safety metric forecasting with safety metric as the y-axis and time as the x-axis is shown.}
\caption{The overall process for training the temporal safety metric forecaster for safety monitoring.
}
\label{fig:sm-process}
\end{figure*}

\autoref{fig:sm-process} depicts the overall process for training and deploying the safety monitor.
The process starts with data generation using System-in-the-Loop (SITL) simulation where the required data to train the safety monitor, as discussed above, is generated.
Then, in the training stage, we preprocess the execution history, tune the hyperparameters of the safety metric prediction model, and train the best model on the complete dataset.
Finally, the trained model is deployed during system operation where the future values of the safety metric are predicted, which is in turn used for safety violation prediction.

As surveyed by Benidis et al. \cite{BENIDIS2022DL4TS}, various DL models with different architectures have been proposed and applied to time-series forecasting.
The number and variety of the proposed models make the problem of selecting the appropriate DL model for safety metric and violation prediction an important challenge, which can only be addressed through empirical investigation.
To this end, we have empirically evaluated state-of-the-art DL-based time-series forecasting models in our specific application context. %

\section{Empirical Evaluation}\label{sec:evaluation}

In this section, we report the empirical evaluation of time series-based safety monitors applied to an ACT and an ADS system.
We aim to answer the following research questions:

\begin{description}
    \item[RQ\textsubscript{1} (Safety Metric Prediction Accuracy)]
    How do different forecasting models score and compare in terms of safety metric prediction accuracy?
    \item[RQ\textsubscript{2} (Safety Violation Prediction Accuracy)]
    How do different forecasting models perform and compare in terms of safety violation prediction accuracy?
    \item[RQ\textsubscript{3} (Accuracy Sensitivity Analysis)]
    What is the impact of varying lookback and hazard horizon window sizes on safety metrics and safety violation prediction accuracy?
    \item[RQ\textsubscript{4} (Resource Overhead Sensitivity Analysis)]
    How do different forecasting models compare in terms of the memory and time overhead of making predictions?
\end{description}

RQ\textsubscript{1} and RQ\textsubscript{2} are motivated by the wide variety of potentially applicable forecasting models, which raises the need for experimental evaluation to determine which one scores best in terms of safety metric forecast and safety violation prediction accuracy, respectively. RQ\textsubscript{2} is particularly relevant in scenarios where a safety monitor does not have a particularly high accuracy in predicting safety metric values, yet its predictions sufficiently contribute to accurate safety violation predictions.

Note that hazard forecast horizon and lookback window sizes are design choices for system developers.
Increasing the hazard forecast horizon is expected to decrease safety metric prediction accuracy, as indicated by previous studies~\cite{Challu2023NHits}.
Conversely, increasing the lookback window size is expected to enhance accuracy.
However, we also expect such changes to impact models' runtime performance, as they impact the number of model parameters,
influencing factors like latency and memory overhead.
Given that the forecasters are destined for deployment in resource-constrained safety-critical systems, understanding the consequences of altering window configurations---specified by hazard forecast horizon window size and lookback to forecast window size ratio parameters---on prediction accuracy and runtime performance is crucial for system developers in practice.
Consequently, our evaluation further explores the effects of different window configurations on prediction accuracy (RQ\textsubscript{3}) and runtime performance (RQ\textsubscript{4}).

\subsection{Evaluation Subjects}\label{sec:eval-subj}

{We evaluate the DL-based forecasting models by applying them to two case studies related to an autonomous centerline tracking system for autonomous taxiing (ACT), and an autonomous driving system (ADS) focused on lane keeping.
In this section, we provide for each case study, an overview of the subject system, explain the details of the evaluation dataset and the simulation workflow used to generate it.
}

\subsubsection{ACT Case Study}\label{sec:ACT-details}
\paragraph{Subject System and Simulation Platform}\label{sec:ACT-subj-sys}
We used an open-source ACT system~\cite{NASA-ULI}, similar to previous studies that had ACT-related case studies~\cite{Cofer2020ACT, Corina2023ACT, asaadi2020UA2}.
{
Note that the ACT system is crucial for safe taxiing operation of autonomous aviation systems.
As reported by the U.S. Department of Transportation, National Transportation Safety Board (NTSB)~\cite{ntsbReport}, as well as major commercial aircraft manufacturers~\cite{BoeingReport, AirbusReport}, fatalities, loss of aircraft, and other substantial damages have occurred during the taxi stage of flight.
}

\begin{figure}
\centering
\begin{subfigure}
        {0.49\textwidth}
        \centering
        \includegraphics[width=\textwidth]{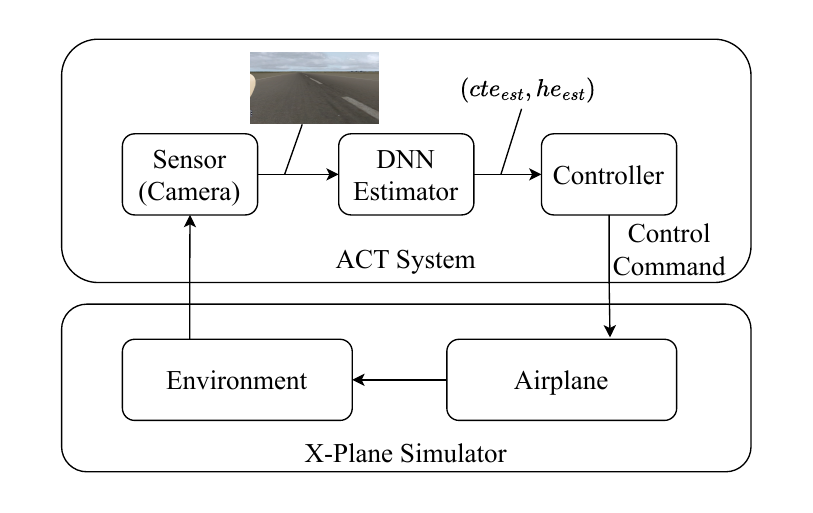}
        \caption{{ACT.}}
        \label{fig:sitl-act}
    \end{subfigure}
    \hfill
    \begin{subfigure}
        {0.49\textwidth}
        \centering
        \includegraphics[width=\textwidth]{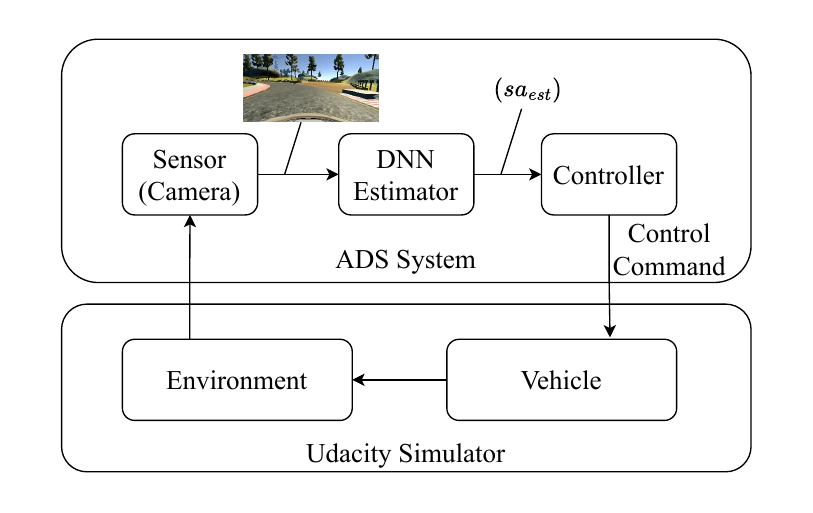}
        \caption{{ADS.}}
        \label{fig:sitl-ads}
    \end{subfigure}
    \Description{A block diagram of the ACT system and the X-Plane simulator. The former includes camera as sensor, a DNN estimator, and a controller. The flow of control commands from the controller to the airplane is shown which then flows to the environment. The environment provides the information for the next state which is again captured by the camera.}
    \caption{{System-in-the-loop simulation of the ACT system and the X-Plane simulator (a), and the ADS system and the Udacity simulator (b).
    }
    }
    \label{fig:data-generation}
\end{figure}

As illustrated in \autoref{fig:sitl-act}, the ACT system consists of a camera, a learned component (i.e., a DNN estimator that outputs cross-track error $cte$ and heading error $he$ estimates given an image input), and a proportional controller that generates control commands steering the aircraft.

Previous studies that used ACT as a case study used the TaxiNet model~\cite{Cofer2020ACT, Corina2023ACT, asaadi2020UA2}, a DNN developed by Boeing for ACT applications, as their learned component.
Since we were not granted access to TaxiNet,
we relied on the open source version called TinyTaxiNet~\cite{NASA-ULI},\footnote{To the best of our knowledge, this is the only open-source DNN model trained for ACT.} which has a lower number of deep layers and input vector size.
We used X-Plane 11~\cite{xplane11}, a high-fidelity flight simulator--- which is used for training pilots~\cite{xplane11} in all flight phases including taxiing---to control various aircraft and environmental parameters.
Based on the simulator's controllable elements, in line with previous studies~\cite{NASA-ULI}, we considered the following four scenario elements:
period of the day,
cloud cover,
starting cross-track error, and
starting heading error.
A detailed explanation of the scenario elements and their value ranges, and the TinyTaxiNet model 
can be found in the supporting material (see \autoref{sec:data-availability}).

Considering the capability of X-Plane in controlling environmental parameters and the major functionality of our target learned component (i.e., ACT), we focus on the following {two} safety requirement{s:
\begin{enumerate*}
    \item \textit{``The aircraft shall have a distance ($|cte_{act}|$) no more than $cte_{thr}$ from the centerline, while taxiing on the runway.''}
    \item \textit{``The aircraft shall have a heading angle ($|he_{act}|$) no more than $he_{thr}$ from the centerline, while taxiing on the runway.''}
\end{enumerate*}
}
To compute the safety metric related to the above requirements, we measure the actual distance of the center of the aircraft from the centerline ($|cte_{act}|$){, as well as the actual angle between the longitudinal axis of the aircraft and the centerline ($|he_{act}|$),} at every time step.
Given our definition of the safety metric provided in \autoref{sec:problem} (\autoref{eq:sm-fcn-def}), negative values of the computed safety metric ($|cte_{act}|<cte_{thr}$) imply that the {$cte$} safety requirement is \emph{not} violated, whereas zero and positive values ($|cte_{act}|\geq cte_{thr}$) indicate a safety violation.
{Note that the same applies to the $he$ safety requirement.}
Based on the size of the aircraft and its taxiing speed, we set $cte_{thr}$ to \SI{5}{m} {and $he_{thr}$ to \SI{5}{\degree},} in line with other ACT-related studies~\cite{asaadi2019UA1, Corina2023ACT, asaadi2020assured}.

\paragraph{Evaluation Dataset}\label{sec:ACT-dataset}
Given the input and output of the safety metric forecaster, discussed in \autoref{sec:solution}, the dataset used to train and evaluate the DL-based forecasters must contain time series values of the learned component's output and the safety metric, given a specific scenario.
Thus, we generated a dataset based on the autonomous taxiing aircraft case study (\autoref{sec:ACT-subj-sys}, inspired by the simulation setup proposed by \cite{NASA-ULI}).

{Concretely, we used the Latin Hypercube Sampling (LHS) method to generate 1,996 unique scenarios to ensure coverage of the scenario parameter search space~\cite{Mckay2000LHS}.
LHS is a sampling method that is used to generate near-random samples from a multidimensional distribution, while ensuring uniform coverage of the simulation input space (i.e., \emph{scenario space}) by stratifying each input dimension~\cite{Mckay2000LHS, sanches2005LHS}.
Given that the high-fidelity simulations that we require to generate the dataset are computationally expensive, LHS allows us to obtain diverse scenarios with a limited number (1996) of simulations.
}
We executed each scenario using the ACT simulation stack until the aircraft reaches to its destination on the runway (taking an average duration of \SI{200}{s}), whereby time series data of TinyTaxiNet outputs ($cte_{est}$, $he_{est}$) are recorded, as well as the safety metric{s} (computed based on \autoref{eq:sm-fcn-def}, with $cte_{thr}=\SI{5}{m}$ {and $he_{thr}=\SI{5}{\degree}$,} as discussed in \autoref{sec:ACT-subj-sys}).
{Thanks to our sampling strategy,
$52\%$ and $21.9\%$ of the scenarios
recorded in the dataset for the $cte$ and $he$ safety requirements, respectively,
include safety violations in their test set (the division of the dataset is discussed in \autoref{sec:RQ1-method}).
We therefore generate a large number of diverse $cte$ and $he$ safety violations in the test sets.
In many simulations, the ACT system misidentifies an imaginary line, which is a parallel offset of the runway centerline by more than $\SI{5}{m}$, as the actual centerline and continues taxiing along it.
This can explain why the dataset includes more $cte$ safety violations than $he$ safety violations.}

Finally, following best practice, similar to the guidelines provided by previous studies on training and evaluating DL-based time series forecasting models~\cite{LIM2021TFT, SALINAS2020DeepAR, Wen2017}, we normalized the time series data, i.e., $cte_{est}$, $he_{est}$ and the safety metric{s} values using Z-score normalization, thereby reducing model bias caused by differences in time series magnitudes among various parameters and scenarios~\cite{LIMA2023100407}.
Further details regarding our generated dataset can be found in the supporting material (see \autoref{sec:data-availability}).

\subsubsection{ADS Case Study}\label{sec:ADS-details}
{
For the ADS case study, we relied on the artifacts available in the paper by \citet{Stocco2023ThirdEye}, where the authors tested a lane-keeping ADS in a driving simulator~\cite{Stocco2023ThirdEye}.
The dataset includes not only the time series of the learned component outputs and the scenario parameters but also the time series of the safety metric, which is crucial for our safety monitoring method.
Over the rest of this section, we provide an overview of the ADS subject system and the simulation platform used to generate the dataset.
We then discuss the details of the raw dataset, as well as our preprocessing steps leading to the final dataset used by our study.
}

\paragraph{Subject System and Simulation Platform}\label{sec:ADS-subj-sys}
{
The ADS case study involves a lane keeping ADS widely used in previous studies~\cite{Stocco2020SelfOracle, Stocco2023ThirdEye, Humbatova2021deepCrime, Jahangirova2021metrics, riccio2020testing}.
Note that the lane keeping ADS is critical for safe operation of AVs.
As described in the original paper, in the U.S., run-off-road crashes are one the most important types of road accidents in terms of frequency and cost.

As depicted in \autoref{fig:sitl-ads}, the ADS consists of a camera, a learned component (i.e., a DNN which estimates the required \textit{steering angle} $sa_{est}$ to keep the vehicle within the lane given an image input), and a controller that issues the control commands to the vehicle.
The learning component is based on the NVIDIA Dave-2 model architecture~\cite{nvidia-dave2}, a DNN-based steering angle estimator that is trained with a set of images collected while a human driver is driving a vehicle.
The simulator used to evaluate the ADS case study was the Udacity simulator for self-driving cars~\cite{udacity}, a driving simulator which has been widely used in the ADS testing literature~\cite{Jahangirova2021metrics, Stocco2020SelfOracle,Jahangirova2021metrics,Stocco2023ThirdEye,Humbatova2021deepCrime,https://doi.org/10.1002/smr.2386}.
Udacity provides closed-loop tracks (circuits) to simulate an ADS driving under various scenario conditions.
Based on the controllable elements of the Udacity simulator, the original paper considered the following
two
scenario elements: weather conditions (i.e., clear, fog, rain, and snow), and weather intensity.\footnote{Note that we are only mentioning the parts of the study conducted by \citet{Stocco2023ThirdEye} that are relevant to our study. We refer the reader to the study itself for comprehensive details on all the evaluations conducted by the authors.}
Since the authors record the time series of $cte_{act}$, we use the recorded value at the beginning of each episode as our third scenario element, i.e., the starting cross-track error, which indicates the initial position of the vehicle.
A detailed explanation of the value ranges for the scenario parameters can be found in the supporting material (\autoref{sec:data-availability}).

Considering the main functionality of the target learned component, i.e., lane keeping, the following safety requirement is considered:
\textit{``The vehicle shall have a distance ($|cte_{act}|$) no more than $cte_{thr}$ from the centerline, while driving on track''.}
To calculate the safety metric, it is necessary to obtain measurements of $cte_{act}$, which are provided in the original dataset.
Similar to the $cte$ safety requirement for the ACT case study, the safety requirement is violated when $|cte_{act}|\geq cte_{thr}$ and not violated otherwise.
Given that the total width of the track set in the simulator,
the size and the speed of the vehicle, we set $cte_{thr}$ to $\SI{5}{m}$, which is reasonable as it provides the vehicle less than $\SI{0.5}{m}$ of side clearance from the edge of the track.
}

\paragraph{Evaluation Dataset}\label{sec:ADS-dataset}

{
The dataset generated by \citet{Stocco2023ThirdEye}  contains a time series of the Dave-2 outputs (estimated steering angles), a time series of $cte_{act}$ (which we used to compute the time series of the safety metric, as described above), and the scenario parameters.

Concretely, the authors executed the ADS in the Udacity simulator, i.e., let it drive for one lap around the track under various weather conditions with intensity increments of $10\%$.
Therefore, to cover a diverse range of the scenario space, the authors recorded 31 one-lap simulations
($1\times\text{clear} + 10\times\text{fog} + 10\times\text{rain} + 10\times\text{snow} = 31$)
which include the time series of $cte_{act}$ measurements~\cite{Stocco2023ThirdEye}.
Note that in some scenarios, the ADS drives the car out of the track, in which case the car is reset on the next waypoint on the track.
This reset during the execution of the scenario leads to a discontinuity during the execution of the learned component output and safety metric time series.
Therefore, we divide the executions at the points where the vehicle has gone out of the track completely into separate episodes, to handle the discontinuity in the time series data.
However, this has led to having diverse execution lengths, e.g., from $\SI{5}{s}$ to more than $\SI{100}{s}$.
To make the size of the episodes more uniform, we discarded the very short episodes, i.e., less than $\SI{15}{s}$, as they do not contain the minimum number of timesteps required to train and test the DL-forecasting models, and divided the larger episodes into shorter chunks.
The resulting dataset contains 175 episodes with an average duration of $\SI{17.5}{s}$.
Note that the size of the ADS dataset is a fraction of (approximately $0.8\%$) the size of the dataset we generated for the ACT case study.
As we will see, this will have an impact on our results and conclusions.
Despite the ADS dataset size, we observe that $33.7\%$ of the episodes include safety violations in their test set.
Therefore, the ADS dataset includes a considerable number of diverse safety violations thanks to the sampling strategy used to search the scenario space, as discussed above.

Finally, similar to the ACT case study (\autoref{sec:ACT-dataset}), we normalized the time series data, i.e., estimated steering angle and the safety metric values using Z-score normalization.
We have included more details about the raw and the preprocessed datasets in our supporting material (\autoref{sec:data-availability}).
}

\subsection{Models Under Evaluation}\label{sec:EvalModels}
In this section, we outline the chosen forecasting models for evaluation and discuss the hyperparameter tuning process applied to optimize the selected models.

As discussed in \autoref{sec:solution}, we are interested in global univariate probabilistic forecasting models that take in input both dynamic time series and static scenario data and provide probabilistic forecasts of the safety metric.
We selected the models for evaluation from the GluonTS library~\cite{gluonts_jmlr}, a widely used probabilistic DL-based forecasting Python library, containing the implementations of many state-of-the-art models.
From the list of available models in the library, we selected four models, three of which satisfy all the requirements of the safety metric forecasting problem (i.e., a global univariate probabilistic forecasting model capable of processing both static and dynamic inputs), whereas the fourth model acts as a competitive baseline, even though it does not fully satisfy all requirements.

\begin{itemize}
    \item MQCNN: a sequence-to-sequence model which is the CNN-based variant of Multi Quantile Recurrent Forecaster, using a CNN encoder instead of an RNN~\cite{Wen2017}.
    \item Temporal Fusion Transformer (TFT): a sequence-to-sequence transformer-based model~\cite{LIM2021TFT}.
    \item Seq2Seq: a vanilla sequence-to-sequence model with a CNN encoder and an MLP decoder~\cite{gluonts_jmlr}.
    \item DeepAR: an iterative model which utilizes both RNNs and autoregressive techniques to iteratively capture temporal dependencies~\cite{SALINAS2020DeepAR}.
    Due to DeepAR's architecture, it only takes as input the static scenario parameters and time series of the target variable, i.e., safety metric.
    Despite DeepAR architecture's lack of ability to process the time series of the learned component output, we have included it in our evaluation as a competitive baseline, due to its high performance in the forecasting benchmarks~\cite{makridakis2023forecasting} and wide use in industry~\cite{BENIDIS2022DL4TS}.
\end{itemize}

\paragraph{Hyperparameter Tuning}\label{sec:hp-tuning}
We fine-tuned the hyperparameters of each considered model before answering the research questions, considering the values (fixed or range) retrieved from the original publications.
{We have tuned the hyperparameters for each safety requirement separately, as they lead to different datasets for the models to be trained and tested on.}
{The relevant hyperparameters and their value ranges for each model are as follows:}

\begin{itemize}
    \item\textit{MQCNN}~\cite{Wen2017}.
    \emph{Number of layers in the MLP decoder (or dim)} was selected in the range \{2, 4, 8\}.
    \emph{Number of neurons in the hidden layer} was selected in the range \{20, 40, 80\}.
    \emph{Number of channels per layer of the CNN encoder} was chosen in the range \{20, 40, 80\}.
    \item\textit{TFT}~\cite{LIM2021TFT}. 
    We set the \emph{dropout rate} to values ranging from 0.1 to 0.3 in steps of 0.1.
    We took the values of \emph{number of attention heads} and \emph{state size} in the ranges \{1, 4\} and \{80, 160, 320\}, respectively.
    We kept the loss function the same as the original paper, i.e., \emph{quantile (pinball) loss}~\cite{LIM2021TFT}.
    \item\textit{Seq2Seq}.
    \emph{Number of layers} was chosen from \{2, 4, 8\}.
    \emph{Number of neurons per layer} was selected from \{10, 20, 40\}.
    \item\textit{DeepAR}~\cite{SALINAS2020DeepAR}.
    The \emph{number of RNN layers} was set to 3 as in the original study~\cite{SALINAS2020DeepAR}.
    The \emph{number of RNN nodes per layer} was selected in the range \{40, 100\}.
    We selected the type of RNN nodes in each layer as being one among \{LSTM, GRU\}.
    We selected the \emph{dropout rate} in the range \{0.1, 0.2, 0.3\}.
    The loss function \emph{negative log likelihood} was used, in line with the original study~\cite{SALINAS2020DeepAR}.
    \item\textit{DL Training}.
    For all the deep learning models above, we selected the hyper-parameters related to training as follows.
    We chose \emph{learning rate} in the range \{0.0001, 0.001, 0.01\}.
    We selected \emph{max gradient norm} in the range \{0.01, 1.0, 100.0\}.
    We evaluated \emph{batch size} in the range \{64, 128, 256\}.
   The Adam optimizer~\cite{Kingma2014AdamAM} was used for training all the models, as per the original paper implementations or that of the GluonTS library.
\end{itemize}

For other hyperparameters, we relied on the suggested values used in the original studies or the default value in their implementation (additional details on the parameter settings are available in the supporting material in \autoref{sec:data-availability}).

We trained each model configuration (defined by a combination of hyperparameters) on the training set (i.e., 70\% of the dataset) and evaluated it on the validation dataset (i.e., 10\% of the dataset), 5 times, to account for randomness, for example, due to
random seeds.
Similar to RQ\textsubscript{1}, we compared the models based on their q-Risk (\autoref{eq:q-risk}) values at quantiles considered in RQ\textsubscript{1}.
The details of the evaluation metric, i.e., q-Risk and the quantiles under consideration, are presented in \autoref{sec:RQ1-method}.
\autoref{tab:HyperParams} summarizes the hyperparameters for each model, their possible values, and the selected hyperparameters{, for $cte$ (ACT$_{cte}$) and $he$ (ACT$_{he}$) safety requirements of the ACT case study and the $cte$ (ADS$_{cte}$) safety requirement of the ADS case study, respectively}.

\begin{table}
  \caption{Hyperparameters of the models under evaluation{, for $cte$ and $he$ safety requirements of the ACT case study, and the $cte$ requirement of the ADS case study}.}
  \label{tab:HyperParams}
  \resizebox{0.75\textwidth}{!}{
  \begin{tabular}{ccccc}
    \toprule
    Hyperparameter&Value Range&{ACT$_{cte}$}&{ACT$_{he}$}&{ADS$_{cte}$}\\
    \midrule
    \multicolumn{4}{c}{Seq2Seq}\\
    \midrule
    Batch Size&64, 128, 256&{128}&{64}&{64}\\
    Learning Rate&1e-4, 1e-3, 1e-2&{1e-3}&{1e-4}&{1e-4}\\
    Gradient Clipping Value&1e-2, 1.0, 1e+2&{1.0}&{1e-2}&{1e+2}\\
    Number of MLP Decoder Layers& 1, 2, 4&{2}&{2}&{2}\\
    Number of Neurons per MLP Layer& 20, 80&{80}&{20}&{80}\\
    \midrule
    \multicolumn{4}{c}{DeepAR}\\
    \midrule
    Batch Size&64, 128, 256&{64}&{64}&{64}\\
    Learning Rate&1e-4, 1e-3, 1e-2&{1e-2}&{1e-3}&{1e-3}\\
    Gradient Clipping Value&1e-2, 1.0, 1e+2&{1e-2}&{1.0}&{1.0}\\
    RNN Node Type& LSTM, GRU&{GRU}&{GRU}&{LSTM}\\
    Number of RNN Nodes&40, 100&{40}&{40}&{100}\\
    Dropout Rate&0.1, 0.2, 0.3&{0.1}&{0.1}&{0.1}\\
    \midrule
    \multicolumn{4}{c}{TFT}\\
    \midrule
    Batch Size&64, 128, 256&{256}&{256}&{128}\\
    Learning Rate&1e-4, 1e-3, 1e-2&{1e-3}&{1e-3}&{1e-2}\\
    Gradient Clipping Value&1e-2, 1.0, 1e+2&{1e-2}&{1.0}&{1e+2}\\
    State Size&40 , 80, 160&{160}&{160}&{160}\\
    Number of Attention Heads&1, 4&{4}&{4}&{4}\\
    Dropout Rate&0.1, 0.2, 0.3&{0.1}&{0.1}&{0.1}\\
    \midrule
    \multicolumn{4}{c}{MQCNN}\\
    \midrule
    Batch Size&64, 128, 256&{256}&{64}&{128}\\
    Learning Rate&1e-4, 1e-3, 1e-2&{1e-3}&{1e-4}&{le-4}\\
    Gradient Clipping Value&1e-2, 1.0, 1e+2&{1.0}&{1e-2}&{1.0}\\
    Number of MLP Decoder Layers& 1, 2, 4&{2}&{2}&{2}\\
    Number of Neurons per MLP Layer& 20, 80&{20}&{20}&{80}\\
    Number of Channels&20, 40&{20}&{20}&{20}\\
  \bottomrule
\end{tabular}
}
\end{table}

\paragraph{Evaluation Hardware}\label{sec:eval-hardware}
To train each configuration of the models under investigation on the evaluation dataset and evaluate them, we used the following compute resources: 1x NVIDIA V100 GPU with 32GB HBM2 memory, 16 cores of Intel Silver 4216 Cascade Lake 2.1GHz CPU, and 128GB of RAM. 

\subsection{\texorpdfstring{RQ\textsubscript{1}}{RQ1}: Safety Metric Forecast Accuracy}\label{sec:RQ1}

{In this section, first we provide the details of our evaluation methodology to answer RQ\textsubscript{1} (\autoref{sec:RQ1-method}).
Then, we present the results for the autonomous taxiing (ACT) case study (\autoref{sec:RQ1-act-results}), followed by the results of the autonomous driving (ADS) case study (\autoref{sec:RQ1-ads-results}).
Finally, we draw conclusions from both case studies (\autoref{sec:RQ1-discussion}) and present our answer to RQ\textsubscript{1}.}

\subsubsection{Methodology}\label{sec:RQ1-method}

{
To answer RQ\textsubscript{1}, we divide each dataset (detailed in \autoref{sec:ACT-dataset} and \autoref{sec:ADS-dataset}) into training, validation, and test datasets, which correspond to 70\%, 10\% and 20\% of the dataset, respectively.
The training set is used to train the time series forecasting models, whereas the validation dataset is used for hyper-parameter tuning (\autoref{sec:hp-tuning}).
Finally, we generated predictions using the trained models on the test dataset, which is disjoint from the training and validation datasets.
To avoid \emph{look-ahead bias}~\cite{Iyer2019lookahead}, we used time-based splitting~\cite{peixeiro2022ts, BENIDIS2022DL4TS, lim2021time}, such that all the samples in the test dataset occur after the validation dataset, whose samples occur after the training dataset.
Our evaluation method, also referred to as \emph{rolling-horizon out-of-sample testing}~\cite{TASHMAN2000437}, provides an evaluation of the model forecasting accuracy on multiple rolling-window samples in each time series that are not seen by the model during training, and aggregates them over all time series in the dataset (see \autoref{eq:q-risk})
\footnote{This method is also in line with the evaluation method used by the reference studies of the models evaluated in paper~\cite{Wen2017,LIM2021TFT,SALINAS2020DeepAR,BENIDIS2022DL4TS}}.
Note that, the evaluation of the model is conducted over multiple subsequent samples in the test set of each time series, thus providing an accuracy measure of the forecasting model over time.
}

The most accurate model resulting from the hyper-parameter tuning phase, in terms of the loss function, was selected and retrained on the union of training and validation datasets. The hyper-parameters for each optimized model selected for evaluation, are listed in \autoref{tab:HyperParams}.
The retrained model was evaluated against the test dataset and its corresponding evaluation metric was computed.
To account for randomness (in the training process), we repeated the above process, i.e., training the best model on the joint training and validation set and evaluating it on the test set, 30 times and reported descriptive statistics of the evaluation metric.
To evaluate the statistical significance of the difference in accuracy metrics of different DL-based safety metric forecasters, we used the Mann-Whitney U test~\cite{mann1947test}.
To measure the effect size of the differences, we measured Vargha and Delaney's $\hat{A}_{AB}$, where $0 \leq\hat{A}_{AB} \leq 1$~\cite{vargha2000effect}. Generally, the value of $\hat{A}_{AB}$ indicates a small, medium, and large difference (effect size) between populations $A$ and $B$ when it is higher than 0.56, 0.64, and 0.71, respectively.

We investigated RQ\textsubscript{1} while considering, as the window size for the hazard forecast horizon, the minimum reaction time required for a human to take over control of the system in case of a hazard.
Considering that each time step in the ACT dataset corresponds to one second and the minimum reaction time for a human with vehicles traveling at \SI{30}{mi\per h} is \SI{3}{s}~\cite{Stocco2023ThirdEye}, we set the minimum hazard forecast horizon to 3 timesteps.
Furthermore, the lookback to forecast horizon ratio was set to 3 times, since it has been frequently considered in previous studies~\cite{LIM2021TFT, SALINAS2020DeepAR}.

{Similarly, for the ADS dataset, we selected the hazard forecast horizon of 3 timesteps equaling to \SI{3}{s}, which is in line with the minimum reaction time suggested by the original study~\cite{Stocco2023ThirdEye}.
However, due to the small size of the dataset, as described in \autoref{sec:ADS-dataset}, we could only select the lookback to a forecast horizon ratio of 1, i.e., \SI{3}{s}.
Note that larger lookbacks to forecast horizon ratios significantly increase the total window size and reduce the number of samples available to train and test the model.}

\paragraph{Evaluation Metric}
As discussed in \autoref{sec:solution}, a probabilistic forecast is better suited {than a point forecast} for critical applications, such as predicting a safety violation, as it provides forecast intervals with attached probabilities.
Similar to previous studies in other application domains, where the performance of probabilistic forecasting models has been reported~\cite{LIM2021TFT, SALINAS2020DeepAR}, we report the \emph{q-Risk} metric at multiple quantiles.
Equation~\ref{eq:q-risk} provides the definition of q-Risk.
Intuitively, \emph{q-Risk} measures the quantile loss~\cite{Wen2017} (\autoref{eq:QL}) across the entire hazard forecast horizon, normalized over the length of the horizon and over all samples in the test set.
Thus, it allows us to compare the safety metric prediction accuracy of models under evaluation at each prediction quantile.
Formally, q-Risk is defined as follows:

\begin{equation}
    q\text{-Risk}=\frac{
    2{\sum_{\mathit{y_t}\in\mathit{\Tilde{\Omega}}}}{\sum^\mathit{\tau_{max}}_{\tau=1}}\mathit{QL}\big(\mathit{y_t},~\mathit{\hat{y}}(q, t-\tau, \tau),~q
    \big)
    }
    {
    {\sum_{\mathit{y}_\mathit{t}\in\mathit{\Tilde{\Omega}}}}{\sum^{\tau_{max}}_{\tau=1}|\mathit{y_t}|}
    },
\label{eq:q-risk}
\end{equation}
where $\mathit{\Tilde{\Omega}}$ is the test set, $q$ is the quantile, $\tau=1,\dots,\tau_\mathit{max}$ is the time step counter of the hazard forecast horizon\footnote{$\tau=1$ and $\tau=\tau_\mathit{max}$ correspond to $\mathit{t}+1$ and $\mathit{t}+h$ in Equation~\ref{eq:prob-def}, respectively.} and $\mathit{QL}$ is the quantile loss function, which is defined in Equation~\ref{eq:QL}.

\begin{equation}
    \mathit{QL}(\mathit{y}, \mathit{\hat{y}}, q)=q(\mathit{y}-\mathit{\hat{y}})_+ +(1-q)(\mathit{\hat{y}}-\mathit{y})_+,
\label{eq:QL}
\end{equation}
where $(.)_+=\mathit{max}(0,.)$.

Given the safety-critical nature of the decisions that need to be made based on the predicted safety metric values, we reported the q-risk at quantiles that correspond to tail-end values of the prediction interval, namely $90\%$ (q=0.05, 0.95), $95\%$ (q=0.025, 0.975), $99\%$ (q=0.005, 0.995), as well as the median (q=0.5) of the prediction distribution.\footnote{Note that the median of the predicted probability distribution often corresponds to the single value predicted by \emph{point} forecasting methods~\cite{BENIDIS2022DL4TS}.}
{
Recall that, according the safety metric function defined in \autoref{sec:problem} (\autoref{eq:sm-fcn-def}\footnote{{As mentioned in \autoref{sec:ACT-subj-sys}, substituting $cte$ with $he$ in \autoref{eq:viol-pred}, provides us with the safety metric function definition for the $he$ safety requirement.}}),
negative values of the safety metric imply no violation of the safety requirement while non-negative safety metric values imply a violation.
Further, the higher the negative values, the closer the system is to a safety violation.
Moreover, note that safety metric values predicted at higher prediction quantiles ($q>0.5$) provide the upper bounds of the predicted safety metric value, which are more conservative estimates based on the definition of the safety metric function.
Therefore, given their safety-critical application, we expect the predictions that our proposed safety monitors generate at quantiles $q>0.5$ to be more useful for predicting safety violations, than predictions for other quantiles.
}

\subsubsection{{ACT Case Study Results}}\label{sec:RQ1-act-results}
\autoref{tab:qRisk-values} reports the achieved q-Risk values for Seq2Seq, DeepAR, MQCNN, and TFT over 30 repetitions at different quantiles, where the best
value for each quantile is written in \textbf{bold}.

\begin{table}[t]
\caption{q-Risk values for different models and quantiles{, for the $cte$ and $he$ safety requirements of the ACT case study and the $cte$ safety requirement of the ADS case study, respectively}.}
\centering
\resizebox{\textwidth}{!}{
\begin{tabular}{lccccccc}
\toprule
\multirow{2}{*}{Model}&\multicolumn{7}{c}{Average q-Risk~$\pm~0.5\times CI_{0.95}$}\\
\cmidrule(lr){2-8}
& $q=0.005$ & $q=0.025$ & $q=0.05$ & $q=0.5$ & $q=0.95$ & $q=0.975$ & $q=0.995$ \\
\midrule
&\multicolumn{7}{c}{{ACT$_{cte}$}}\\
\cmidrule(lr){2-8}
Seq2Seq & $0.038 \pm 0.0076$ & $0.046 \pm 0.0056$ & $0.052 \pm 0.0028$ & $0.040 \pm 0.0013$ & $0.026 \pm 0.0014$ & $0.024 \pm 0.0019$ & $0.020 \pm 0.0026$\\
DeepAR & $0.004 \pm 0.0006$ & $0.012 \pm 0.0012$ & $0.020 \pm 0.0019$ & $0.062 \pm 0.0035$ & $0.016 \pm 0.0008$ & $0.009 \pm 0.0005$ & $0.003 \pm 0.0002$\\
MQCNN & $0.012 \pm 0.0015$ & $0.015 \pm 0.0011$ & $0.018 \pm 0.0016$ & $0.030 \pm 0.0010$ & $0.026 \pm 0.0025$ & $0.023 \pm 0.0022$ & $0.017 \pm 0.0026$\\
TFT & $\mathbf{0.001 \pm 0.0001}$ & $\mathbf{0.003 \pm 0.0001}$ & $\mathbf{0.004 \pm 0.0002}$ & $\mathbf{0.012 \pm 0.0002}$ & $\mathbf{0.005 \pm 0.0001}$ & $\mathbf{0.003 \pm 0.0001}$ & $\mathbf{0.001 \pm 0.0001}$\\
\midrule
&\multicolumn{7}{c}{{ACT$_{he}$}}\\
\cmidrule(lr){2-8}
{Seq2Seq} & {$0.208 \pm 0.0420$} & {$0.300 \pm 0.0363$} & {$0.420 \pm 0.0370$} & {$0.841 \pm 0.0204$} & {$0.541 \pm 0.0390$} & {$0.458 \pm 0.0314$} & {$0.330 \pm 0.0351$}\\
{DeepAR} & {$\mathbf{0.041 \pm 0.0022}$} & {$0.118 \pm 0.0040$} & {$0.199 \pm 0.0056$} & {$0.732 \pm 0.0119$} & {$0.296 \pm 0.0092$} & {$0.196 \pm 0.0077$} & {$0.086 \pm 0.0053$}\\
{MQCNN} & {$0.092 \pm 0.0197$} & {$0.171 \pm 0.0161$} & {$0.260 \pm 0.0151$} & {$0.705 \pm 0.0170$} & {$0.418 \pm 0.0372$} & {$0.316 \pm 0.0215$} & {$0.180 \pm 0.0273$}\\
{TFT} & {$\mathbf{0.041 \pm 0.0026}$} & {$\mathbf{0.096 \pm 0.0037}$} & {$\mathbf{0.140 \pm 0.0041}$} & {$\mathbf{0.379 \pm 0.0025}$} & {$\mathbf{0.158 \pm 0.0036}$} & {$\mathbf{0.112 \pm 0.0034}$} & {$\mathbf{0.049 \pm 0.0025}$}\\
\midrule
&\multicolumn{7}{c}{{ADS$_{cte}$}}\\
\cmidrule(lr){2-8}
{Seq2Seq} & {$0.019 \pm 0.0025$} & {$0.056 \pm 0.0064$} & {$0.074 \pm 0.0065$} & {$0.222 \pm 0.0033$} & {$0.137 \pm 0.0073$} & {$0.097 \pm 0.0089$} & {$0.045 \pm 0.0127$}\\
{DeepAR} & {$\mathbf{0.007 \pm 0.0004}$} & {$\mathbf{0.023 \pm 0.0008}$} & {$\mathbf{0.042 \pm 0.0011}$} & {$\mathbf{0.204 \pm 0.0031}$} & {$0.124 \pm 0.0049$} & {$0.090 \pm 0.0045$} & {$0.048 \pm 0.0041$}\\
{MQCNN} & {$0.015 \pm 0.0021$} & {$0.052 \pm 0.0076$} & {$0.070 \pm 0.0080$} & {$0.214 \pm 0.0031$} & {$0.119 \pm 0.0068$} & {$0.077 \pm 0.0060$} & {$0.028 \pm 0.0068$}\\
{TFT} & {$0.023 \pm 0.0040$} & {$0.045 \pm 0.0023$} & {$0.069 \pm 0.0035$} & {$0.226 \pm 0.0066$} & {$\mathbf{0.098 \pm 0.0049}$} & {$\mathbf{0.065 \pm 0.0039}$} & {$\mathbf{0.023 \pm 0.0006}$}\\
\bottomrule
\end{tabular}
}
\label{tab:qRisk-values}
\end{table}

Overall, we observe that TFT consistently outperforms the other models at all reported quantiles{, except in the case of the $he$ safety requirement when $q<0.025$, where TFT and DeepAR both have the lowest q-Risk value}.
Furthermore, {for the $cte$ safety requirement,} DeepAR is the second best at very high and low ends of the quantile spectrum (specifically, when $q<0.025$ or $q\geq0.95$), while yielding the worst accuracy at the median of the forecast probability distribution ($q=0.5$).
{Similarly, for the $he$ safety requirement, DeepAR is the second best at the ends of the quantile spectrum (except when $q<0.025$, as mentioned above), while being second to last at the median ($q=0.5$).
}
We suspect that the fact that DeepAR is an iterative forecasting model, as opposed to the other three models which are sequence-to-sequence forecasting models, could explain the large variability in q-Risk values over quantiles.
Since iterative forecasting models, such as DeepAR, only predict the target value for the next timestep and use the predicted value to predict the timestep after that (as explained in \autoref{sec:background}), they are prone to accumulating forecasting errors from previous forecast timesteps.
DeepAR's error accumulation is more extreme when predicting at quantiles closer to the median ($\text{q}=0.5$), where other models also have higher q-Risk values than for other quantiles.

\begin{figure}
    \begin{subfigure}
        {0.49\textwidth}
        \centering
        \includegraphics[width=\textwidth]{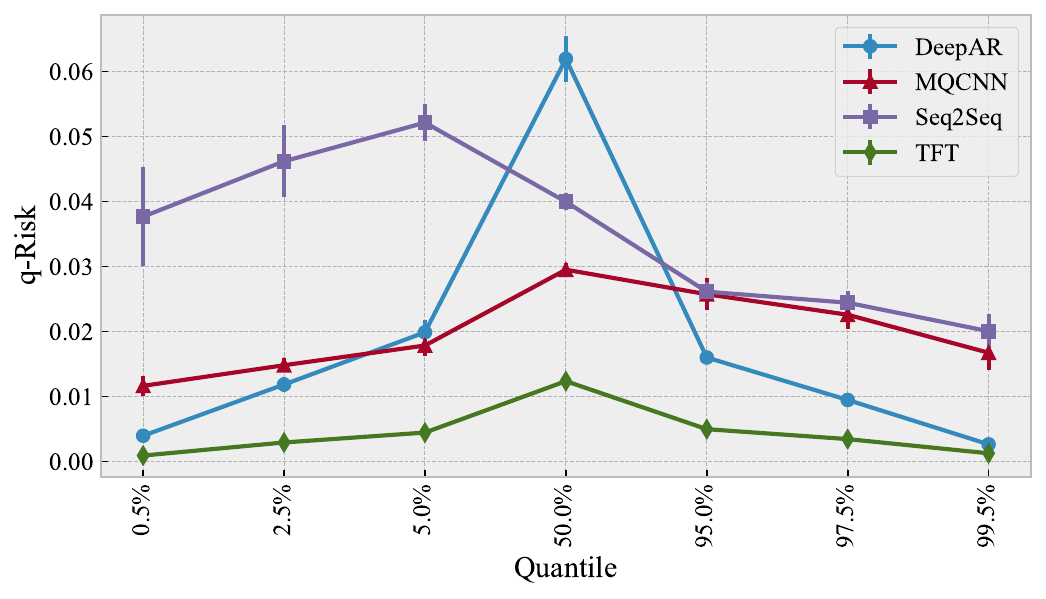}
        \caption{{ACT$_{cte}$}}
        \label{fig:rq1-act_cte}
    \end{subfigure}
    \hfill
    \begin{subfigure}
        {0.49\textwidth}
        \centering
        \includegraphics[width=\textwidth]{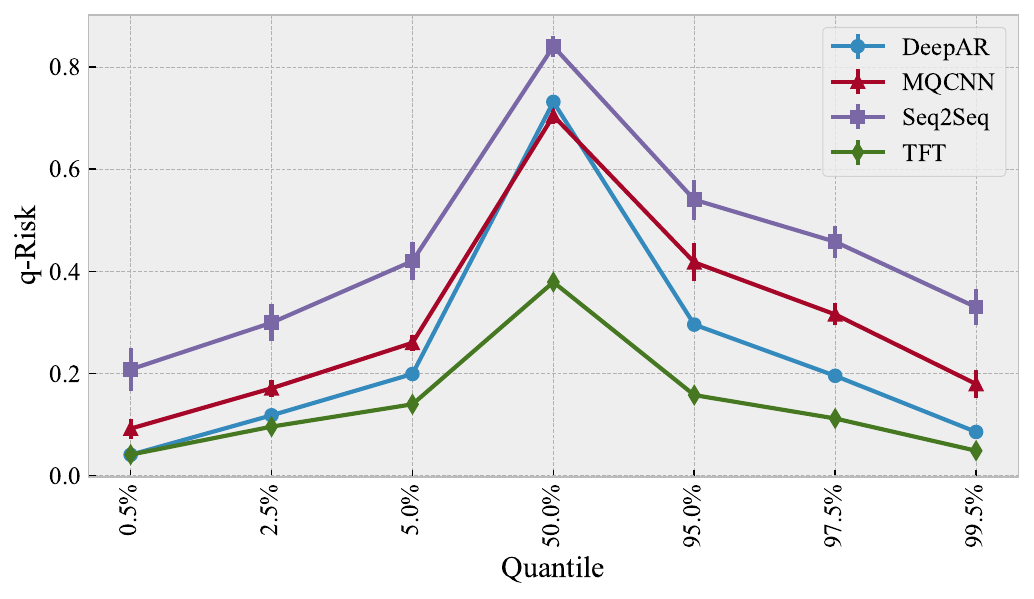}
        \caption{{ACT$_{he}$}}
        \label{fig:rq1-act_he}
    \end{subfigure}
    \hfill
    \begin{subfigure}
        {0.49\textwidth}
        \centering
        \includegraphics[width=\textwidth]{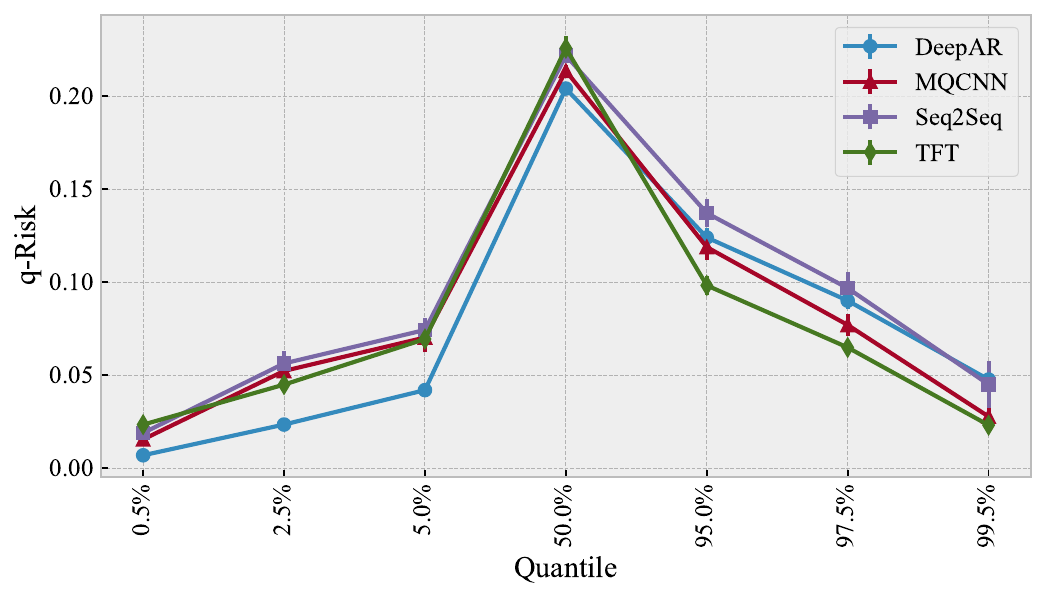}
        \caption{{ADS$_{cte}$}}
        \label{fig:rq1-av_cte}
    \end{subfigure}
    \Description{A diagram of q-Risk versus quantile for the four evaluated methods. TFT shows the lowest q-Risk over all quantiles.}
    \caption{Average q-Risk values and their corresponding 95\% confidence interval (CI\textsubscript{0.95}) for all models over all reported quantiles{, for the $cte$ and $he$ safety requirements of the autonomous taxiing (ACT) case study and the $cte$ safety requirement of the autonomous driving (ADS) case study, respectively}. Note that the x-axis is \emph{not} drawn to scale, in favor of a more readable presentation.
    }
    \label{fig:rq1-q-risk}
\end{figure}

\autoref{fig:rq1-act_cte}, which depicts the q-Risk average values and their 95\% confidence interval for different models at all measured quantiles, illustrates the large change in performance (average q-Risk values) of DeepAR.

\begin{table*}
  \centering
  \small
  \caption{Statistical comparison of q-Risk values for different DL-based forecasters at different quantiles q.}
  \label{tab:rq1-stat-test}
  \resizebox{\textwidth}{!}{
  \begin{tabular}{cccccccccccccccc}
  \toprule
    \multicolumn{2}{c}{Comparison}&\multicolumn{14}{c}{q-Risk}
    \\
    \cmidrule(lr){1-2}\cmidrule(lr){3-16}
    \multirow{2}{2em}{$A$}&\multirow{2}{2em}{$B$}&\multicolumn{2}{c}{$q=0.005$}&\multicolumn{2}{c}{$q=0.025$}&\multicolumn{2}{c}{$q=0.05$}&\multicolumn{2}{c}{$q=0.5$}&\multicolumn{2}{c}{$q=0.95$}&\multicolumn{2}{c}{$q=0.975$}&\multicolumn{2}{c}{$q=0.995$}
    \\
    \cmidrule(lr){3-4}\cmidrule(lr){5-6}\cmidrule(lr){7-8}\cmidrule(lr){9-10}\cmidrule(lr){11-12}\cmidrule(lr){13-14}\cmidrule(lr){15-16}
    &&$p$&$\hat{A}_{AB}$&$p$&$\hat{A}_{AB}$&$p$&$\hat{A}_{AB}$&$p$&$\hat{A}_{AB}$&$p$&$\hat{A}_{AB}$&$p$&$\hat{A}_{AB}$&$p$&$\hat{A}_{AB}$\\

    \midrule
    &&\multicolumn{14}{c}{{ACT$_{cte}$}}\\
    \cmidrule(lr){3-16}
    Seq2Seq & DeepAR & $\num{8.15e-11}$ & $\num{0.99}$ & $\num{6.72e-10}$ & $\num{0.96}$ & $\num{3.34e-11}$ & $\num{1.00}$ & $\num{3.02e-11}$ & $\num{0.00}$ & $\num{1.33e-10}$ & $\num{0.98}$ & $\num{3.34e-11}$ & $\num{1.00}$ & $\num{3.02e-11}$ & $\num{1.00}$\\
    Seq2Seq & MQCNN & $\num{2.00e-05}$ & $\num{0.82}$ & $\num{1.56e-08}$ & $\num{0.93}$ & $\num{3.02e-11}$ & $\num{1.00}$ & $\num{7.39e-11}$ & $\num{0.99}$ & $\num{4.04e-01}$ & $\num{0.56}$ & $\num{9.33e-02}$ & $\num{0.63}$ & $\num{4.51e-02}$ & $\num{0.65}$\\
    Seq2Seq & TFT & $\num{3.02e-11}$ & $\num{1.00}$ & $\num{3.02e-11}$ & $\num{1.00}$ & $\num{3.02e-11}$ & $\num{1.00}$ & $\num{3.02e-11}$ & $\num{1.00}$ & $\num{3.02e-11}$ & $\num{1.00}$ & $\num{3.02e-11}$ & $\num{1.00}$ & $\num{3.02e-11}$ & $\num{1.00}$\\
    DeepAR & MQCNN & $\num{2.87e-10}$ & $\num{0.03}$ & $\num{1.41e-04}$ & $\num{0.21}$ & $\num{7.48e-02}$ & $\num{0.63}$ & $\num{3.02e-11}$ & $\num{1.00}$ & $\num{3.20e-09}$ & $\num{0.05}$ & $\num{3.02e-11}$ & $\num{0.00}$ & $\num{3.02e-11}$ & $\num{0.00}$\\
    DeepAR & TFT & $\num{3.02e-11}$ & $\num{1.00}$ & $\num{3.02e-11}$ & $\num{1.00}$ & $\num{3.02e-11}$ & $\num{1.00}$ & $\num{3.02e-11}$ & $\num{1.00}$ & $\num{3.02e-11}$ & $\num{1.00}$ & $\num{3.02e-11}$ & $\num{1.00}$ & $\num{3.02e-11}$ & $\num{1.00}$\\
    MQCNN & TFT & $\num{3.02e-11}$ & $\num{1.00}$ & $\num{3.02e-11}$ & $\num{1.00}$ & $\num{3.02e-11}$ & $\num{1.00}$ & $\num{3.02e-11}$ & $\num{1.00}$ & $\num{3.02e-11}$ & $\num{1.00}$ & $\num{3.02e-11}$ & $\num{1.00}$ & $\num{3.02e-11}$ & $\num{1.00}$\\

    \midrule
    &&\multicolumn{14}{c}{{ACT$_{he}$}}\\
    \cmidrule(lr){3-16}
    {Seq2Seq} & {DeepAR} & {$\num{8.48e-09}$} & {$\num{0.93}$} & {$\num{4.50e-11}$} & {$\num{1.00}$} & {$\num{3.02e-11}$} & {$\num{1.00}$} & {$\num{4.62e-10}$} & {$\num{0.97}$} & {$\num{3.02e-11}$} & {$\num{1.00}$} & {$\num{3.02e-11}$} & {$\num{1.00}$} & {$\num{3.02e-11}$} & {$\num{1.00}$}\\
    {Seq2Seq} & {MQCNN} & {$\num{2.60e-05}$} & {$\num{0.82}$} & {$\num{1.60e-07}$} & {$\num{0.89}$} & {$\num{1.29e-09}$} & {$\num{0.96}$} & {$\num{9.92e-11}$} & {$\num{0.99}$} & {$\num{4.64e-05}$} & {$\num{0.81}$} & {$\num{1.56e-08}$} & {$\num{0.93}$} & {$\num{1.87e-07}$} & {$\num{0.89}$}\\
    {Seq2Seq} & {TFT} & {$\num{8.48e-09}$} & {$\num{0.93}$} & {$\num{3.02e-11}$} & {$\num{1.00}$} & {$\num{3.02e-11}$} & {$\num{1.00}$} & {$\num{3.02e-11}$} & {$\num{1.00}$} & {$\num{3.02e-11}$} & {$\num{1.00}$} & {$\num{3.02e-11}$} & {$\num{1.00}$} & {$\num{3.02e-11}$} & {$\num{1.00}$}\\
    {DeepAR} & {MQCNN} & {$\num{8.15e-05}$} & {$\num{0.20}$} & {$\num{8.20e-07}$} & {$\num{0.13}$} & {$\num{8.48e-09}$} & {$\num{0.07}$} & {$\num{1.70e-02}$} & {$\num{0.68}$} & {$\num{3.08e-08}$} & {$\num{0.08}$} & {$\num{6.72e-10}$} & {$\num{0.04}$} & {$\num{2.44e-09}$} & {$\num{0.05}$}\\
    {DeepAR} & {TFT} & {$\num{8.77e-01}$} & {$\num{0.49}$} & {$\num{4.18e-09}$} & {$\num{0.94}$} & {$\num{3.02e-11}$} & {$\num{1.00}$} & {$\num{3.02e-11}$} & {$\num{1.00}$} & {$\num{3.02e-11}$} & {$\num{1.00}$} & {$\num{3.02e-11}$} & {$\num{1.00}$} & {$\num{3.69e-11}$} & {$\num{1.00}$}\\
    {MQCNN} & {TFT} & {$\num{8.66e-05}$} & {$\num{0.80}$} & {$\num{3.16e-10}$} & {$\num{0.97}$} & {$\num{3.02e-11}$} & {$\num{1.00}$} & {$\num{3.02e-11}$} & {$\num{1.00}$} & {$\num{3.02e-11}$} & {$\num{1.00}$} & {$\num{3.02e-11}$} & {$\num{1.00}$} & {$\num{3.02e-11}$} & {$\num{1.00}$}\\
    \midrule
    &&\multicolumn{14}{c}{{ADS$_{cte}$}}\\
    \cmidrule(lr){3-16}
    {Seq2Seq} & {DeepAR} & {$\num{4.98e-11}$} & {$\num{0.99}$} & {$\num{3.02e-11}$} & {$\num{1.00}$} & {$\num{3.02e-11}$} & {$\num{1.00}$} & {$\num{2.39e-08}$} & {$\num{0.92}$} & {$\num{3.67e-03}$} & {$\num{0.72}$} & {$\num{5.40e-01}$} & {$\num{0.55}$} & {$\num{3.39e-02}$} & {$\num{0.34}$}\\
    {Seq2Seq} & {MQCNN} & {$\num{4.36e-02}$} & {$\num{0.65}$} & {$\num{2.12e-01}$} & {$\num{0.59}$} & {$\num{9.05e-02}$} & {$\num{0.63}$} & {$\num{1.25e-04}$} & {$\num{0.79}$} & {$\num{5.56e-04}$} & {$\num{0.76}$} & {$\num{1.17e-03}$} & {$\num{0.74}$} & {$\num{2.50e-03}$} & {$\num{0.73}$}\\
    {Seq2Seq} & {TFT} & {$\num{2.06e-01}$} & {$\num{0.40}$} & {$\num{2.07e-02}$} & {$\num{0.67}$} & {$\num{5.79e-01}$} & {$\num{0.54}$} & {$\num{8.53e-01}$} & {$\num{0.49}$} & {$\num{2.44e-09}$} & {$\num{0.95}$} & {$\num{3.96e-08}$} & {$\num{0.91}$} & {$\num{1.58e-01}$} & {$\num{0.61}$}\\
    {DeepAR} & {MQCNN} & {$\num{8.99e-11}$} & {$\num{0.01}$} & {$\num{3.02e-11}$} & {$\num{0.00}$} & {$\num{3.02e-11}$} & {$\num{0.00}$} & {$\num{2.01e-04}$} & {$\num{0.22}$} & {$\num{6.35e-02}$} & {$\num{0.64}$} & {$\num{5.56e-04}$} & {$\num{0.76}$} & {$\num{1.75e-05}$} & {$\num{0.82}$}\\
    {DeepAR} & {TFT} & {$\num{3.02e-11}$} & {$\num{0.00}$} & {$\num{3.02e-11}$} & {$\num{0.00}$} & {$\num{3.02e-11}$} & {$\num{0.00}$} & {$\num{2.38e-07}$} & {$\num{0.11}$} & {$\num{9.26e-09}$} & {$\num{0.93}$} & {$\num{7.12e-09}$} & {$\num{0.94}$} & {$\num{3.34e-11}$} & {$\num{1.00}$}\\
    {MQCNN} & {TFT} & {$\num{1.77e-03}$} & {$\num{0.26}$} & {$\num{6.00e-01}$} & {$\num{0.54}$} & {$\num{1.54e-01}$} & {$\num{0.39}$} & {$\num{1.24e-03}$} & {$\num{0.26}$} & {$\num{2.88e-06}$} & {$\num{0.85}$} & {$\num{1.17e-03}$} & {$\num{0.74}$} & {$\num{7.73e-02}$} & {$\num{0.37}$}\\
    \bottomrule
\end{tabular}}
\end{table*}

Our visual observations are supported by the statistical comparison results reported in \autoref{tab:rq1-stat-test}.
Columns $A$ and $B$ indicate the DL-forecasting models being compared.
Columns $p$ and $\hat{A}_{AB}$ indicate statistical significance and effect size (as described in \autoref{sec:RQ1-method}), respectively, when comparing A and B in terms of q-Risk at different quantiles q.
Given a significance level of $\alpha=0.01$, for the $cte$ safety requirement,
we observe that the differences between the best model (TFT) in all quantiles, and the other models are significant.
Furthermore, for all quantiles, $\hat{A}_{AB}$ is greater than $0.71$ when $B=\text{TFT}$, indicating that the difference between TFT and other models is large.
{For the $he$ safety requirement, TFT and DeepAR are equally the best models, when $q<0.025$.
In this case, we observe that TFT is significantly better than the second-best model, i.e., MQCNN, with a large difference, though that is not the case for DeepAR.}

\subsubsection{{ADS Case Study Results}}\label{sec:RQ1-ads-results}
{
The q-Risk values achieved by Seq2Seq, DeepAR, MQCNN, and TFT over 30 repetitions at different quantiles are reported in \autoref{tab:qRisk-values}.
Note that the best value for each quantile is written in \textbf{bold}.

Overall, we observe that at each quantile, the difference between the most accurate and least accurate models are less than what is observed for the ACT case study results at similar quantiles (compare \autoref{fig:rq1-av_cte} vs \autoref{fig:rq1-act_cte} and \autoref{fig:rq1-act_he}).
We believe that this is due to the sample size of the ADS dataset which is significantly lower than the size of the ACT dataset, as discussed in \autoref{sec:ACT-dataset}, where a model like TFT is expected to suffer the most, as it is a transformer-based model, which has been shown to require significantly more training data than other DL-based models such as CNNs in vision tasks~\cite{dosovitskiy2021imageworth16x16words}.
Nonetheless, we observe that TFT achieves the lowest q-Risk values (most accurate predictions) when $q>0.5$.
Whereas, for $q\leq0.5$, DeepAR outperforms other models.

Our statistical test results (\autoref{tab:rq1-stat-test}), confirm that TFT significantly outperforms other models when $0.5<q<0.995$, with a large effect size as the corresponding $\hat{A}_{AB}$ values are greater than $0.71$.
However, at $q=0.995$, using the Mann-Whitney U-test, when $A\in\{\text{Seq2Seq}, \text{MQCNN}\}$ and $B\in\{\text{TFT}\}$, indicate that their difference
is not statistically significant (with a confidence level of $95\%$), as p-values are larger than $0.05$.
Therefore, at $q=0.995$, we conclude that sequence-to-sequence models, i.e., TFT, MQCNN and Seq2Seq, all equally yield the best safety metric prediction accuracy.
Finally, we observe that for $q\leq0.5$, DeepAR consistently outperforms other models with a large effect size.
}

\subsubsection{{Discussion}}\label{sec:RQ1-discussion}
{
Given the results of the ACT case study (\autoref{sec:RQ1-act-results}), we observed that, for probabilistic prediction of both $cte$ and $he$ safety metrics, at all measured quantiles $q$, with a practical window configuration, i.e., hazard forecast horizon of \SI{3}{s} (which correlates to the minimum reaction time, as discussed in \autoref{sec:RQ1-method}) and lookback to forecast horizon ratio of three (which is similar to the ratio used by the literature in multiple forecasting problems~\cite{LIM2021TFT, SALINAS2020DeepAR}),
TFT yields significantly more accurate quantile forecasts than Seq2Seq, DeepAR, and MQCNN.
Our observations for the $he$ safety requirement is the same as $cte$, for $q\geq0.025$.
Whereas, for $q<0.025$, TFT and DeepAR both yield the highest time series prediction accuracy.
Therefore, we can conclude that for the ACT dataset, where the dataset size is large and contains numerous safety violations, TFT is the best model or one of the best models to be used for probabilistic forecasting of the safety metric values at all quantiles, given a practical window configuration.

For the ADS case study results (\autoref{sec:RQ1-ads-results}), we observed that for probabilistic safety metric prediction, again with a practical window configuration, i.e., hazard forecast horizon of \SI{3}{s} and a lookback to forecast horizon ratio of one, as discussed in \autoref{sec:RQ1-method}, TFT yields significantly more accurate predictions when $0.5<q<0.995$.
At $q=0.995$, all sequence-to-sequence models, i.e., TFT, MQCNN and Seq2Seq, yield the lowest q-Risk value.
Finally, DeepAR yields significantly more accurate predictions when $q\leq0.5$.
Therefore, for the ADS case study, where the size of the dataset is a fraction of the ACT dataset, as discussed in \autoref{sec:ADS-dataset}, TFT is the best or one of the best models to be used for safety metric forecasting when $q>0.5$.
Though DeepAR is the most accurate model when $q\leq0.5$, as discussed in \autoref{sec:RQ1-method}, given the definition of the safety metric (\autoref{eq:sm-fcn-def}),
forecasts for quantiles $q>0.5$ are more important for safety violation prediction. 
}

\begin{tcolorbox}
{
For the ACT case study,
where the size of the dataset is large,
given a practical window configuration, i.e., hazard forecast horizon of \SI{3}{s} and lookback to forecast horizon ratio of 3,
TFT is more suitable than Seq2Seq, DeepAR and MQCNN,
for probabilistic safety metric forecasting over all reported quantiles,
for both $cte$ and $he$ safety requirements.

For the ADS case study,
where the size of the dataset is small,
given a practical window configuration of $h=\SI{3}{s}$ and $cm=1$,
DeepAR is significantly more accurate than TFT, MQCNN, and Seq2Seq, for $q\leq0.5$.
However, TFT is the most accurate model,
among the evaluated models,
when $q>0.5$, which is a more important quantile range in a safety monitoring context, as discussed in \autoref{sec:RQ1-method}.

Therefore, TFT is the most accurate model, when $q>0.5$, for both case studies.
}
\end{tcolorbox}

\subsection{\texorpdfstring{RQ\textsubscript{2}}{RQ2}: Safety Violation Prediction Accuracy}

{In this section, similar to \autoref{sec:RQ1}, first we provide the details of our evaluation methodology to answer RQ\textsubscript{2} (\autoref{sec:RQ2-method}).
Then, we present the results for the ACT and ADS case studies (\autoref{sec:RQ2-act-results} and \autoref{sec:RQ2-ads-results}, respectively).
Finally, we draw conclusions from the results of both case studies (\autoref{sec:RQ2-discussion}) and present our answer to RQ\textsubscript{2}.}

\subsubsection{Methodology}\label{sec:RQ2-method}

To answer RQ\textsubscript{2}, we reuse the models trained to answer RQ\textsubscript{1} with the aim of predicting safety violations. To do so, we applied the safety violation function (\autoref{eq:viol-pred}) to the safety metric values predicted by the forecasting models, as reported in RQ\textsubscript{1}.
A non-negative safety function value implies a safety violation.
We compared the predicted safety violations with the true safety violation value of the test samples used in RQ\textsubscript{1}.
True safety violation values are calculated by applying the safety violation function to the true safety metric values of the test samples.

\paragraph{Evaluation Metric}
To report the safety violation prediction accuracy of the models, we report
\emph{Precision} (Pr=TP/(TP+FP)) and \emph{Recall} (Re=TP/(TP+FN)),
where \emph{true positives} (TP), \emph{false positive} (FP), and \emph{false negatives} (FN) are the correct, false, and missed safety violation predictions, respectively.
Note that Precision measures the fraction of correct warnings that a safety monitor raises, whereas Recall measures the fraction of safety violations that a safety monitor can successfully predict~\cite{Stocco2023ThirdEye}.
We further compute and report the F\textsubscript{$\beta$} score~\cite{baeza1999modern}
as a weighted balance between precision and recall, with $\beta=3.0$ ($\mathit{F}_3=\frac{10\cdot~\mathit{Precision}\times \mathit{Recall}}{9\cdot \mathit{Precision}+\mathit{Recall}}$), granting higher importance to Recall as compared to Precision, as false negatives have severe consequences for safety-critical systems~\cite{Stocco2023ThirdEye,2024-Grewal-ICST}.
We recall that false negatives, in the context of safety-critical systems, are safety violations that were not predicted by the safety monitor and thus could lead to system hazards.
In contrast, false positives, although an important consideration for the design of the safety monitor, lead to inconvenience or inefficiencies for the users, which are less harmful than safety violations.
For the ACT system specifically, false positives could lead to the disengagement of the autonomous taxiing operation or emergency stops, which could lead to delayed flight operations and schedules.

{
Recall that the predictions for quantiles $q>0.5$ are more important than lower quantiles, as they provide more conservative estimates of the safety metric, as discussed in \autoref{sec:RQ1-method}.
}

\subsubsection{{ACT Case Study Results}}\label{sec:RQ2-act-results}

\begin{figure*}
\begin{subfigure}{0.49\textwidth}
  \centering
  \includegraphics[width=\textwidth]{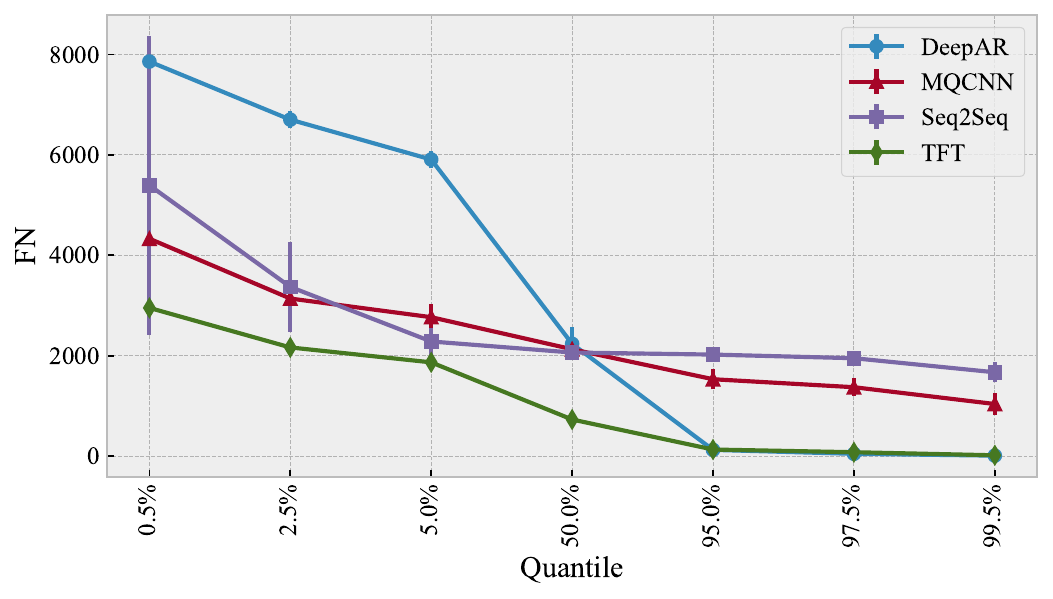}
  \caption{FN vs. q {--- ACT$_{cte}$}}
  \label{fig:rq2-fn-act-cte}
\end{subfigure}
\hfill
\begin{subfigure}{0.49\textwidth}
  \centering
  \includegraphics[width=\textwidth]{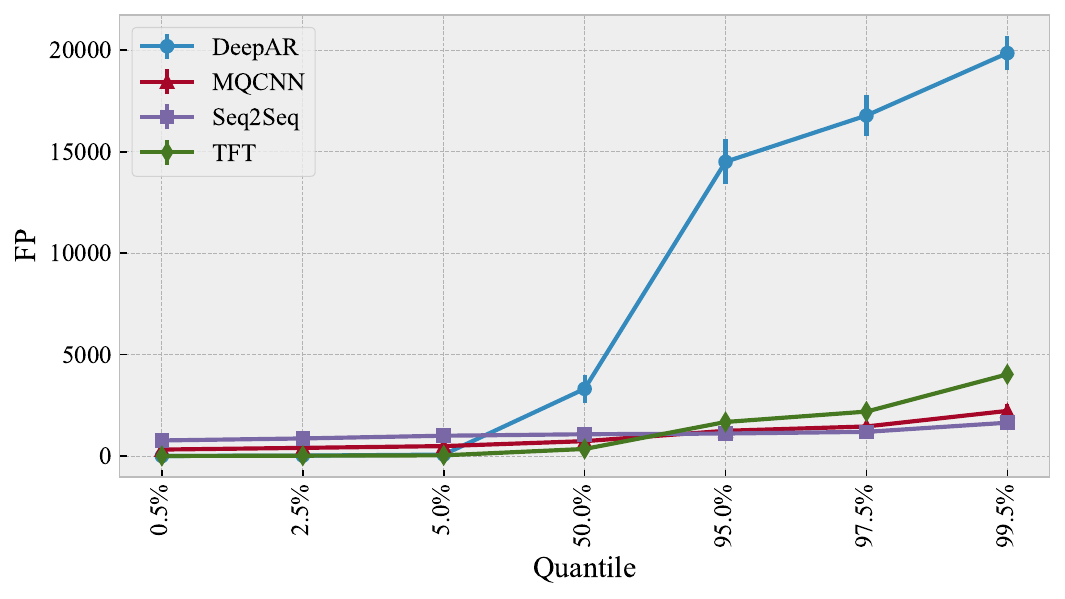}
  \caption{FP vs. q  {--- ACT$_{cte}$}}
  \label{fig:rq2-fp-act-cte}
\end{subfigure}
\hfill
\begin{subfigure}{0.49\textwidth}
  \centering
  \includegraphics[width=\textwidth]{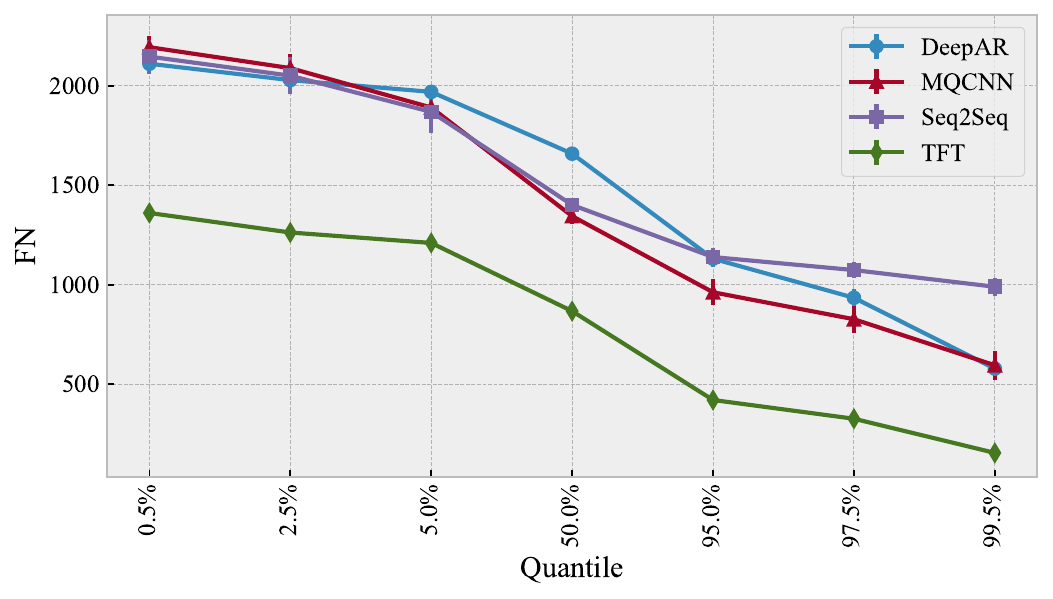}
  \caption{{FN vs. q  --- ACT$_{he}$}}
  \label{fig:rq2-fn-act-he}
\end{subfigure}
\hfill
\begin{subfigure}{0.49\textwidth}
  \centering
  \includegraphics[width=\textwidth]{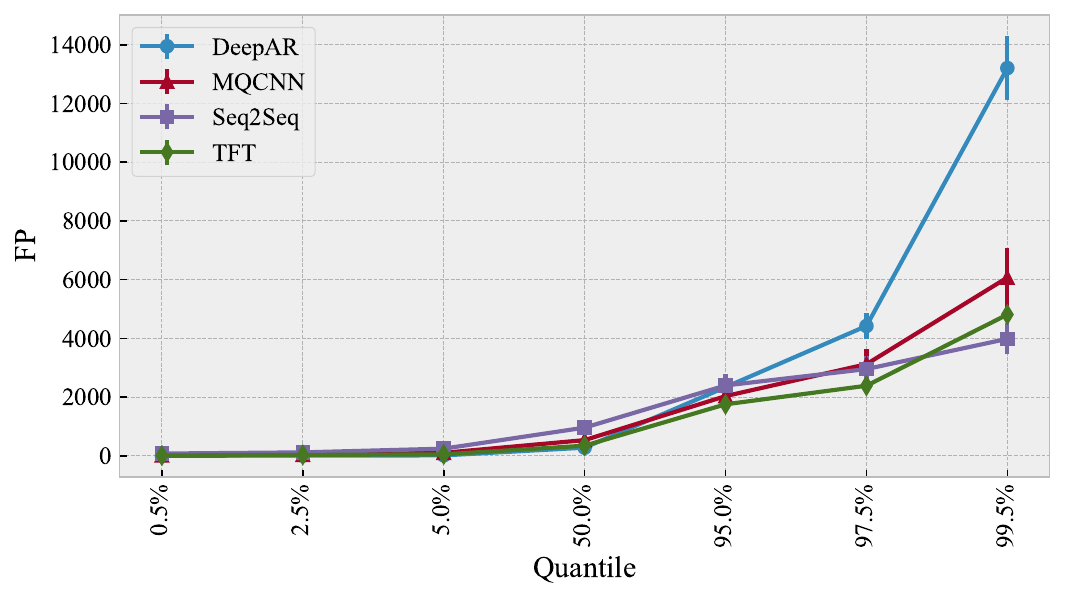}
  \caption{{FP vs. q  --- ACT$_{he}$}}
  \label{fig:rq2-fp-act-he}
\end{subfigure}

\hfill
\begin{subfigure}{0.49\textwidth}
  \centering
  \includegraphics[width=\textwidth]{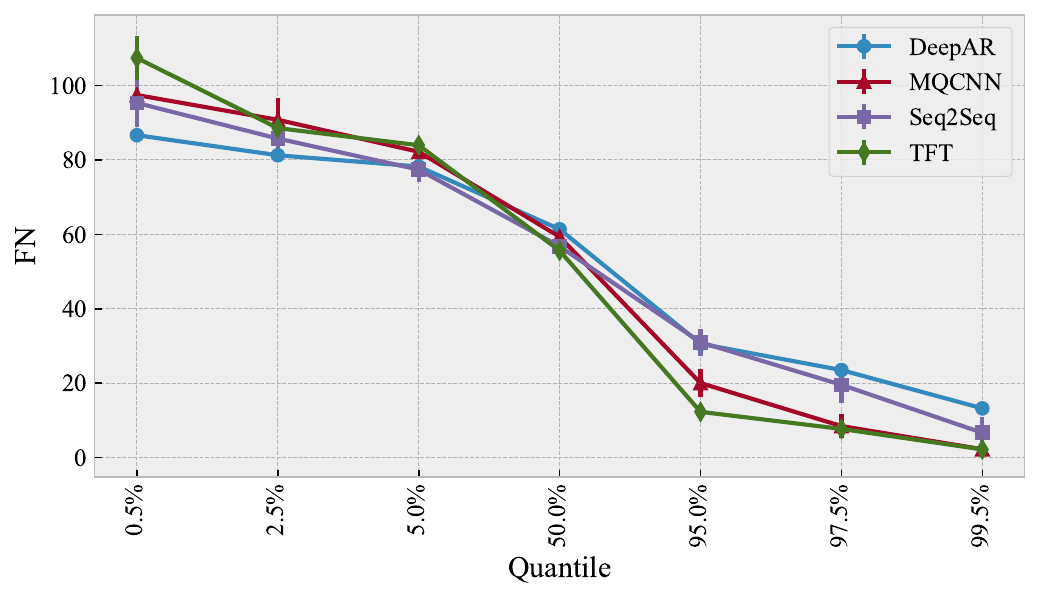}
  \caption{{FN vs. q  --- ADS$_{cte}$}}
  \label{fig:rq2-fn-ads-cte}
\end{subfigure}
\hfill
\begin{subfigure}{0.49\textwidth}
  \centering
  \includegraphics[width=\textwidth]{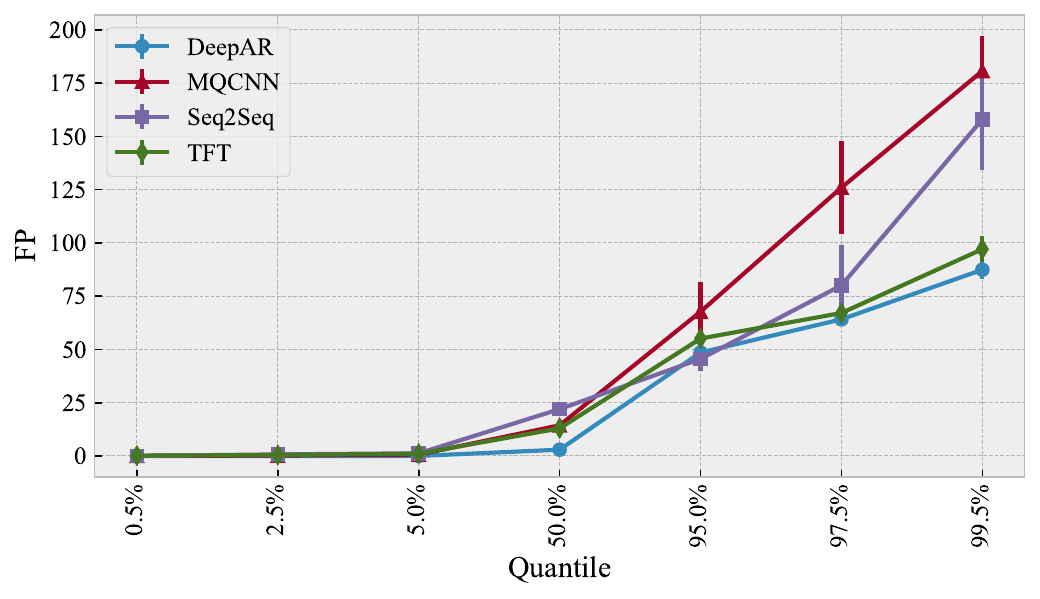}
  \caption{{FP vs. q  --- ADS$_{cte}$}}
  \label{fig:rq2-fp-ads-cte}
\end{subfigure}
\Description{Two subplots for FN and FP versus quantile.}
\caption{False Negatives (FN) and False Positives (FP) at different prediction quantiles (q){, for $cte$ and $he$ safety requirements of the ACT case study and the $cte$ safety requirement of the ADS case study, respectively}. Note that the x-axis is \emph{not} drawn to scale, in favor of a more readable presentation.}
\label{fig:rq2-fnfpvq}
\end{figure*}

\autoref{fig:rq2-fn-act-cte} and \autoref{fig:rq2-fp-act-cte} illustrate the False Negative (FN) and False Positive (FP) values for all the models at the reported quantiles, respectively.\footnote{Note that the x-axis, i.e., quantile $q$, is not drawn up to scale for better readability.}
Overall, we observe that with an increase in the prediction quantile, the FN value decreases while the FP value increases.
This is expected based on the definitions of the safety metric and the safety violation function (\autoref{eq:sm-fcn-def}
and \autoref{eq:viol-pred}, respectively).
{
Recall that,
as mentioned in \autoref{sec:RQ2-method}, the predictions at quantiles $q>0.5$ provide more conservative estimates of the safety metric values.
}
Therefore, the higher the prediction quantile, the higher the probability of the safety monitor correctly predicting safety violations (lower FN) and raising false alarms (higher FP).

As mentioned in \autoref{sec:RQ2-method}, the priority in safety-critical applications is having the lowest FN value possible (since FNs lead to system hazards),
while having a reasonably low FP value (since FPs lead to inefficiencies)
is the second priority.
Therefore, for the ACT {and ADS case studies} targeted in this paper, hereafter we focus on the results for prediction quantiles $q\geq0.5$, for which the FN and FP values are reported in \autoref{tab:rq2-act-results}.
We have provided the values at other quantiles in the supporting material (\autoref{sec:data-availability}).
\autoref{tab:rq2-act-results} reports the averages and their $95\%$ confidence intervals for the evaluation metrics, namely Pr, Re, and $\text{F}_3$. 

\begin{table}[ht]
\caption{
Various safety {violation} prediction accuracy metric values for different models and quantiles{, for $cte$ and $he$ (rows highlighted in \colorbox{lightgray}{gray}) safety requirements of the ACT case study}.
}
\centering
\resizebox{0.6\textwidth}{!}{
\begin{tabular}{lcccc}
\toprule
\multirow{2}{*}{Model}&\multicolumn{4}{c}{Average Metric Value~$\pm~0.5\times CI_{0.95}$}\\
\cmidrule(lr){2-5}
& $q=0.5$ & $q=0.95$ & $q=0.975$ & $q=0.995$ \\
\midrule
\multicolumn{5}{c}{FN}\\
\midrule
\multirow{2}{*}{Seq2Seq} & $2062.9 \pm 111.5$ & $2022.0 \pm 114.5$ & $1949.7 \pm 142.3$ & $1669.6 \pm 198.9$\\
& \cellcolor[HTML]{DCDCDC} {$1400.9 \pm 32.9$} & \cellcolor[HTML]{DCDCDC} {$1138.8 \pm 43.8$} & \cellcolor[HTML]{DCDCDC} {$1073.4 \pm 42.0$} & \cellcolor[HTML]{DCDCDC} {$989.4 \pm 44.7$}\\
\cmidrule(lr){2-5}
\multirow{2}{*}{DeepAR} &$2236.5 \pm 329.5$ & $124.5 \pm 73.9$ & $46.0 \pm 31.1$ & $10.3 \pm 7.6$\\
& \cellcolor[HTML]{DCDCDC} {$1658.8 \pm 20.3$} & \cellcolor[HTML]{DCDCDC} {$1130.5 \pm 40.0$} & \cellcolor[HTML]{DCDCDC} {$934.1 \pm 45.2$} & \cellcolor[HTML]{DCDCDC} {$577.7 \pm 46.2$}\\
\cmidrule(lr){2-5}
\multirow{2}{*}{MQCNN} &$2128.0 \pm 199.9$ & $1531.2 \pm 201.6$ & $1372.3 \pm 174.0$ & $1037.6 \pm 211.2$\\
&\cellcolor[HTML]{DCDCDC} {$1344.7 \pm 40.3$} & \cellcolor[HTML]{DCDCDC} {$961.7 \pm 65.9$} & \cellcolor[HTML]{DCDCDC} {$826.2 \pm 67.1$} & \cellcolor[HTML]{DCDCDC} {$594.6 \pm 73.2$}\\
\cmidrule(lr){2-5}
\multirow{2}{*}{TFT} &$729.2 \pm 27.2$ & $131.4 \pm 17.6$ & $78.1 \pm 11.6$ & $16.8 \pm 2.5$\\
&\cellcolor[HTML]{DCDCDC} {$866.9 \pm 20.4$} & \cellcolor[HTML]{DCDCDC} {$420.0 \pm 24.3$} & \cellcolor[HTML]{DCDCDC} {$325.7 \pm 22.3$} & \cellcolor[HTML]{DCDCDC} {$153.9 \pm 15.6$}\\
\midrule
\multicolumn{5}{c}{FP}\\
\midrule
\multirow{2}{*}{Seq2Seq} & $1073.4 \pm 129.0$ & $1108.9 \pm 150.6$ & $1191.4 \pm 175.6$ & $1642.5 \pm 291.9$\\
&\cellcolor[HTML]{DCDCDC} {$958.8 \pm 165.1$} & \cellcolor[HTML]{DCDCDC} {$2395.2 \pm 402.2$} & \cellcolor[HTML]{DCDCDC} {$2953.1 \pm 460.7$} & \cellcolor[HTML]{DCDCDC} {$3982.8 \pm 528.9$}\\
\cmidrule(lr){2-5}
\multirow{2}{*}{DeepAR} &$3311.0 \pm 684.6$ & $14501.4 \pm 1113.0$ & $16781.9 \pm 998.2$ & $19858.7 \pm 842.3$\\
&\cellcolor[HTML]{DCDCDC} {$276.4 \pm 24.8$} & \cellcolor[HTML]{DCDCDC} {$2337.9 \pm 240.0$} & \cellcolor[HTML]{DCDCDC} {$4420.1 \pm 455.4$} & \cellcolor[HTML]{DCDCDC} {$13203.1 \pm 1093.1$}\\
\cmidrule(lr){2-5}
\multirow{2}{*}{MQCNN} &$732.8 \pm 110.2$ & $1243.0 \pm 160.6$ & $1457.3 \pm 186.4$ & $2224.9 \pm 336.1$\\
&\cellcolor[HTML]{DCDCDC} {$533.7 \pm 76.3$} & \cellcolor[HTML]{DCDCDC} {$2032.3 \pm 363.5$} & \cellcolor[HTML]{DCDCDC} {$3118.9 \pm 522.3$} & \cellcolor[HTML]{DCDCDC} {$6069.3 \pm 1018.0$}\\
\cmidrule(lr){2-5}
\multirow{2}{*}{TFT} &$348.4 \pm 13.3$ & $1677.3 \pm 61.3$ & $2190.1 \pm 78.9$ & $4026.0 \pm 138.0$\\
&\cellcolor[HTML]{DCDCDC} {$344.6 \pm 19.1$} & \cellcolor[HTML]{DCDCDC} {$1752.8 \pm 73.0$} & \cellcolor[HTML]{DCDCDC} {$2385.5 \pm 102.9$} & \cellcolor[HTML]{DCDCDC} {$4808.9 \pm 221.0$}\\
\midrule
\multicolumn{5}{c}{Precision}\\
\midrule
\multirow{2}{*}{Seq2Seq} & $0.978 \pm 0.0026$ & $0.977 \pm 0.0030$ & $0.975 \pm 0.0035$ & $0.967 \pm 0.0056$\\
&\cellcolor[HTML]{DCDCDC} {$0.496 \pm 0.0304$} & \cellcolor[HTML]{DCDCDC} {$0.343 \pm 0.0254$} & \cellcolor[HTML]{DCDCDC} {$0.309 \pm 0.0237$} & \cellcolor[HTML]{DCDCDC} {$0.261 \pm 0.0233$}\\
\cmidrule(lr){2-5}
\multirow{2}{*}{DeepAR} &$0.936 \pm 0.0120$ & $0.773 \pm 0.0138$ & $0.746 \pm 0.0116$ & $0.712 \pm 0.0089$\\
&\cellcolor[HTML]{DCDCDC} {$0.693 \pm 0.0156$} & \cellcolor[HTML]{DCDCDC} {$0.338 \pm 0.0186$} & \cellcolor[HTML]{DCDCDC} {$0.242 \pm 0.0162$} & \cellcolor[HTML]{DCDCDC} {$0.118 \pm 0.0082$}\\
\cmidrule(lr){2-5}
\multirow{2}{*}{MQCNN} &$0.985 \pm 0.0022$ & $0.975 \pm 0.0031$ & $0.970 \pm 0.0036$ & $0.956 \pm 0.0062$\\
&\cellcolor[HTML]{DCDCDC} {$0.646 \pm 0.0261$} & \cellcolor[HTML]{DCDCDC} {$0.422 \pm 0.0330$} & \cellcolor[HTML]{DCDCDC} {$0.343 \pm 0.0304$} & \cellcolor[HTML]{DCDCDC} {$0.249 \pm 0.0335$}\\
\cmidrule(lr){2-5}
\multirow{2}{*}{TFT} &$0.993 \pm 0.0003$ & $0.967 \pm 0.0012$ & $0.957 \pm 0.0015$ & $0.924 \pm 0.0024$\\
&\cellcolor[HTML]{DCDCDC} {$0.804 \pm 0.0081$} & \cellcolor[HTML]{DCDCDC} {$0.515 \pm 0.0085$} & \cellcolor[HTML]{DCDCDC} {$0.451 \pm 0.0087$} & \cellcolor[HTML]{DCDCDC} {$0.308 \pm 0.0086$}\\
\midrule
\multicolumn{5}{c}{Recall}\\
\midrule
\multirow{2}{*}{Seq2Seq} &$0.958 \pm 0.0023$ & $0.959 \pm 0.0023$ & $0.960 \pm 0.0029$ & $0.966 \pm 0.0041$\\
&\cellcolor[HTML]{DCDCDC} {$0.383 \pm 0.0145$} & \cellcolor[HTML]{DCDCDC} {$0.499 \pm 0.0193$} & \cellcolor[HTML]{DCDCDC} {$0.527 \pm 0.0185$} & \cellcolor[HTML]{DCDCDC} {$0.564 \pm 0.0197$}\\
\cmidrule(lr){2-5}
\multirow{2}{*}{DeepAR} &$0.954 \pm 0.0067$ & $0.997 \pm 0.0015$ & $0.999 \pm 0.0006$ & $1.000 \pm 0.0002$\\
&\cellcolor[HTML]{DCDCDC} {$0.270 \pm 0.0089$} & \cellcolor[HTML]{DCDCDC} {$0.502 \pm 0.0176$} & \cellcolor[HTML]{DCDCDC} {$0.589 \pm 0.0199$} & \cellcolor[HTML]{DCDCDC} {$0.746 \pm 0.0203$}\\
\cmidrule(lr){2-5}
\multirow{2}{*}{MQCNN} &$0.956 \pm 0.0041$ & $0.969 \pm 0.0041$ & $0.972 \pm 0.0036$ & $0.979 \pm 0.0043$\\
&\cellcolor[HTML]{DCDCDC} {$0.408 \pm 0.0178$} & \cellcolor[HTML]{DCDCDC} {$0.577 \pm 0.0290$} & \cellcolor[HTML]{DCDCDC} {$0.636 \pm 0.0295$} & \cellcolor[HTML]{DCDCDC} {$0.738 \pm 0.0322$}\\
\cmidrule(lr){2-5}
\multirow{2}{*}{TFT} &$0.985 \pm 0.0006$ & $0.997 \pm 0.0004$ & $0.998 \pm 0.0002$ & $1.000 \pm 0.0001$\\
&\cellcolor[HTML]{DCDCDC} {$0.618 \pm 0.0090$} & \cellcolor[HTML]{DCDCDC} {$0.815 \pm 0.0107$} & \cellcolor[HTML]{DCDCDC} {$0.857 \pm 0.0098$} & \cellcolor[HTML]{DCDCDC} {$0.932 \pm 0.0069$}\\
\midrule
\multicolumn{5}{c}{F\textsubscript{3}}\\
\midrule
\multirow{2}{*}{Seq2Seq} &$0.960 \pm 0.0019$ & $0.960 \pm 0.0020$ & $0.962 \pm 0.0024$ & $0.966 \pm 0.0032$\\
&\cellcolor[HTML]{DCDCDC} {$0.390 \pm 0.0125$} & \cellcolor[HTML]{DCDCDC} {$0.472 \pm 0.0122$} & \cellcolor[HTML]{DCDCDC} {$0.486 \pm 0.0107$} & \cellcolor[HTML]{DCDCDC} {$0.498 \pm 0.0103$}\\
\cmidrule(lr){2-5}
\multirow{2}{*}{DeepAR} &$0.952 \pm 0.0049$ & $0.969 \pm 0.0012$ & $0.966 \pm 0.0015$ & $0.961 \pm 0.0015$\\
&\cellcolor[HTML]{DCDCDC} {$0.287 \pm 0.0090$} & \cellcolor[HTML]{DCDCDC} {$0.476 \pm 0.0135$} & \cellcolor[HTML]{DCDCDC} {$0.510 \pm 0.0129$} & \cellcolor[HTML]{DCDCDC} {$0.481 \pm 0.0146$}\\
\cmidrule(lr){2-5}
\multirow{2}{*}{MQCNN} &$0.959 \pm 0.0036$ & $0.969 \pm 0.0035$ & $0.972 \pm 0.0030$ & $0.976 \pm 0.0034$\\
&\cellcolor[HTML]{DCDCDC} {$0.423 \pm 0.0170$} & \cellcolor[HTML]{DCDCDC} {$0.549 \pm 0.0212$} & \cellcolor[HTML]{DCDCDC} {$0.576 \pm 0.0182$} & \cellcolor[HTML]{DCDCDC} {$0.595 \pm 0.0159$}\\
\cmidrule(lr){2-5}
\multirow{2}{*}{TFT} &$\mathbf{0.986 \pm 0.0005}$ & $\mathbf{0.994 \pm 0.0002}$ & $\mathbf{0.994 \pm 0.0001}$ & $\mathbf{0.992 \pm 0.0003}$\\
&\cellcolor[HTML]{DCDCDC} {$\mathbf{0.633 \pm 0.0084}$} & \cellcolor[HTML]{DCDCDC} {$\mathbf{0.770 \pm 0.0079}$} & \cellcolor[HTML]{DCDCDC} {$\mathbf{0.785 \pm 0.0064}$} & \cellcolor[HTML]{DCDCDC} {$\mathbf{0.774 \pm 0.0048}$}\\
\bottomrule
\end{tabular}
}
\label{tab:rq2-act-results}
\end{table}

\begin{table}[ht]
\caption{
{Various safety violation prediction accuracy metric values for different models and quantiles, for the ADS case study.}
}
\centering
\resizebox{0.6\textwidth}{!}{
\begin{tabular}{lcccc}
\toprule
\multirow{2}{*}{{Model}}&\multicolumn{4}{c}{{Average Metric Value~$\pm~0.5\times CI_{0.95}$}}\\
\cmidrule(lr){2-5}
& {$q=0.5$} & {$q=0.95$} & {$q=0.975$} & {$q=0.995$} \\
\midrule
\multicolumn{5}{c}{{FN}}\\
\midrule
{Seq2Seq} & {$56.8 \pm 1.4$} & {$31.0 \pm 3.7$} & {$19.5 \pm 4.9$} & {$6.6 \pm 4.3$} \\
{DeepAR} & {$61.4 \pm 0.7$} & {$30.6 \pm 1.2$} & {$23.5 \pm 1.5$} & {$13.2 \pm 1.8$} \\
{MQCNN} & {$59.3 \pm 2.0$} & {$20.0 \pm 3.7$} & {$8.4 \pm 3.2$} & {$2.2 \pm 1.8$} \\
{TFT} & {$55.6 \pm 1.8$} & {$12.3 \pm 1.6$} & {$7.7 \pm 1.2$} & {$2.2 \pm 0.3$} \\
\midrule
\multicolumn{5}{c}{{FP}}\\
\midrule
{Seq2Seq} & {$22.0 \pm 1.9$} & {$45.5 \pm 5.5$} & {$80.1 \pm 18.7$} & {$158.0 \pm 23.5$} \\
{DeepAR} & {$2.9 \pm 0.3$} & {$48.5 \pm 3.1$} & {$64.1 \pm 3.2$} & {$87.4 \pm 4.4$} \\
{MQCNN} & {$14.4 \pm 2.5$} & {$67.7 \pm 14.0$} & {$126.0 \pm 21.7$} & {$180.7 \pm 16.4$} \\
{TFT} & {$12.9 \pm 1.5$} & {$55.1 \pm 3.1$} & {$67.1 \pm 3.0$} & {$97.1 \pm 6.0$} \\
\midrule
\multicolumn{5}{c}{{Precision}}\\
\midrule
{Seq2Seq} & {$0.810 \pm 0.0128$} & {$0.728 \pm 0.0153$} & {$0.649 \pm 0.0380$} & {$0.505 \pm 0.0449$} \\
{DeepAR} & {$0.968 \pm 0.0035$} & {$0.711 \pm 0.0120$} & {$0.663 \pm 0.0101$} & {$0.610 \pm 0.0113$} \\
{MQCNN} & {$0.867 \pm 0.0179$} & {$0.676 \pm 0.0311$} & {$0.555 \pm 0.0405$} & {$0.460 \pm 0.0282$} \\
{TFT} & {$0.880 \pm 0.0105$} & {$0.714 \pm 0.0101$} & {$0.679 \pm 0.0085$} & {$0.605 \pm 0.0142$} \\
\midrule
\multicolumn{5}{c}{{Recall}}\\
\midrule
{Seq2Seq} & {$0.619 \pm 0.0092$} & {$0.792 \pm 0.0246$} & {$0.869 \pm 0.0329$} & {$0.955 \pm 0.0289$} \\
{DeepAR} & {$0.588 \pm 0.0047$} & {$0.795 \pm 0.0079$} & {$0.842 \pm 0.0097$} & {$0.911 \pm 0.0121$} \\
{MQCNN} & {$0.602 \pm 0.0131$} & {$0.866 \pm 0.0248$} & {$0.943 \pm 0.0216$} & {$0.985 \pm 0.0122$} \\
{TFT} & {$0.627 \pm 0.0121$} & {$0.918 \pm 0.0109$} & {$0.949 \pm 0.0080$} & {$0.985 \pm 0.0022$} \\
\midrule
\multicolumn{5}{c}{{F\textsubscript{3}}}\\
\midrule
{Seq2Seq} & {$0.634 \pm 0.0084$} & {$0.784 \pm 0.0197$} & {$0.833 \pm 0.0215$} & {$0.866 \pm 0.0144$} \\
{DeepAR} & {$0.612 \pm 0.0046$} & {$0.785 \pm 0.0062$} & {$0.820 \pm 0.0079$} & {$0.868 \pm 0.0096$} \\
{MQCNN} & {$0.620 \pm 0.0119$} & {$0.838 \pm 0.0169$} & {$0.874 \pm 0.0099$} & {$0.880 \pm 0.0028$} \\
{TFT} & {$\mathbf{0.645 \pm 0.0113}$} & {$\mathbf{0.892 \pm 0.0086}$} & {$\mathbf{0.912 \pm 0.0057}$} & {$\mathbf{0.927 \pm 0.0031}$} \\
\bottomrule
\end{tabular}
}
\label{tab:rq2-ads-results}
\end{table}

Overall, Precision for all models drops as q increases, whereas Recall increases{, for both $cte$ and $he$ safety requirements}.
This general trend is in line with the FN and FP trends above. Since estimates become more conservative, more safety violations are correctly predicted (Recall~$\uparrow$) while the proportion of false alarms increases (Precision~$\downarrow$).
{For the $he$ safety requirement, since the proportion of time steps including a safety violation is lower than that of the $cte$ safety requirement (\autoref{sec:ACT-dataset}), we expect and observe that the precision scores of the models recorded for $he$ are lower than the scores recorded for  $cte$.}
{In the case of the $cte$ safety requirement, n}ote that TFT and DeepAR reach a Recall value of 1.0 at a q = 0.995, thus indicating all of the safety violations are correctly predicted, which is also confirmed by the low corresponding FN values; Seq2Seq and MQCNN, on the other hand, yield the lowest Recall scores (high FN values).
However, we observe that DeepAR has the lowest Precision among the models for $\text{q}\geq0.5$, indicating a higher fraction of false alarms, which is also confirmed by the fact that the FP value for DeepAR is an order of magnitude greater than that of other models.
In contrast, TFT has a precision above 0.92, for $\text{q}\geq0.5$, which makes it a more suitable choice for safety monitoring than DeepAR.
{In the case of the $he$ safety requirement, we observe that TFT consistently yields the highest Precision and Recall scores for $\text{q}\geq0.5$, whereas the Recall of other models is $20-40\%$ lower.}
The superiority of TFT over other models is further confirmed by the reported F\textsubscript{3} scores, where TFT consistently has the highest F\textsubscript{3} scores for $\text{q}\geq0.5${, for both the $cte$ and $he$ safety requirements}.
Note that the highest F\textsubscript{3} scores for each quantile are highlighted in bold.
Our visual observations are supported by the statistical comparison results reported in \autoref{tab:rq2-stat-test}.
Columns $A$ and $B$ indicate the DL-forecasting models being compared.
Given a significance level of $\alpha=0.01$,
we observe that the differences between TFT and the other models are significant for all quantiles.
Furthermore, for all quantiles, $\hat{A}_{AB}$ is greater than $0.71$ when $B=\text{TFT}$, indicating that the difference between TFT and other models is large.

\begin{table*}
  \centering
  \small
  \caption{Statistical comparison of F\textsubscript{3} score values for different DL-based forecasters at quantiles q$\geq 0.5${, for $cte$ and $he$ safety requirements of the ACT case study and the $cte$ safety requirement of the ADS case study, respectively}.}
  \label{tab:rq2-stat-test}
  \resizebox{0.85\textwidth}{!}{
  \begin{tabular}{cccccccccc}
  \toprule
    \multicolumn{2}{c}{Comparison}&\multicolumn{8}{c}{F\textsubscript{3} score}
    \\
    \cmidrule(lr){1-2}\cmidrule(lr){3-10}
    \multirow{2}{2em}{$A$}&\multirow{2}{2em}{$B$}
    &\multicolumn{2}{c}{$q=0.5$}&\multicolumn{2}{c}{$q=0.95$}&\multicolumn{2}{c}{$q=0.975$}&\multicolumn{2}{c}{$q=0.995$}
    \\
    \cmidrule(lr){3-4}\cmidrule(lr){5-6}\cmidrule(lr){7-8}\cmidrule(lr){9-10}
    &&$p$&$\hat{A}_{AB}$&$p$&$\hat{A}_{AB}$&$p$&$\hat{A}_{AB}$&$p$&$\hat{A}_{AB}$\\

    \midrule
    TFT & Seq2Seq & $\num{3.02e-11}$ & $\num{1.00}$ & $\num{3.02e-11}$ & $\num{1.00}$ & $\num{3.02e-11}$ & $\num{1.00}$ & $\num{3.02e-11}$ & $\num{1.00}$ \\
    TFT & DeepAR & $\num{3.02e-11}$ & $\num{1.00}$ & $\num{3.02e-11}$ & $\num{1.00}$ & $\num{3.02e-11}$ & $\num{1.00}$ & $\num{3.02e-11}$ & $\num{1.00}$ \\
    TFT & MQCNN & $\num{3.02e-11}$ & $\num{1.00}$ & $\num{3.02e-11}$ & $\num{1.00}$ & $\num{3.02e-11}$ & $\num{1.00}$ & $\num{3.02e-11}$ & $\num{1.00}$ \\
    Seq2Seq & DeepAR & $\num{1.84e-02}$ & $\num{0.68}$ & $\num{1.16e-07}$ & $\num{0.10}$ & $\num{1.60e-03}$ & $\num{0.26}$ & $\num{3.39e-02}$ & $\num{0.66}$ \\
    Seq2Seq & MQCNN & $\num{7.62e-01}$ & $\num{0.48}$ & $\num{2.28e-05}$ & $\num{0.18}$ & $\num{1.25e-05}$ & $\num{0.17}$ & $\num{3.37e-05}$ & $\num{0.19}$ \\
    DeepAR & MQCNN & $\num{3.64e-02}$ & $\num{0.34}$ & $\num{1.37e-01}$ & $\num{0.39}$ & $\num{4.71e-04}$ & $\num{0.24}$ & $\num{7.69e-08}$ & $\num{0.10}$ \\
    \midrule
    &&\multicolumn{8}{c}{{ACT$_{he}$}}\\
    \cmidrule(lr){3-10}
    {TFT}&{Seq2Seq}& {$\num{3.02e-11}$} & {$\num{1.00}$} & {$\num{3.02e-11}$} & {$\num{1.00}$} & {$\num{3.02e-11}$} & {$\num{1.00}$} & {$\num{3.02e-11}$} & {$\num{1.00}$}\\
    {TFT} & {DeepAR} &{$\num{3.02e-11}$} & {$\num{1.00}$} & {$\num{3.02e-11}$} & {$\num{1.00}$} & {$\num{3.02e-11}$} & {$\num{1.00}$} & {$\num{3.02e-11}$} & {$\num{1.00}$}\\
    {TFT} & {MQCNN}&{$\num{3.02e-11}$} & {$\num{1.00}$} & {$\num{3.02e-11}$} & {$\num{1.00}$} & {$\num{3.02e-11}$} & {$\num{1.00}$} & {$\num{3.02e-11}$} & {$\num{1.00}$}\\
    {Seq2Seq} & {DeepAR}&{$\num{8.99e-11}$} & {$\num{0.99}$} & {$\num{5.59e-01}$} & {$\num{0.46}$} & {$\num{5.26e-04}$} & {$\num{0.24}$} & {$\num{3.64e-02}$} & {$\num{0.66}$}\\
    {Seq2Seq} & {MQCNN}&{$\num{3.34e-03}$} & {$\num{0.28}$} & {$\num{3.01e-07}$} & {$\num{0.11}$} & {$\num{1.55e-09}$} & {$\num{0.05}$} & {$\num{1.96e-10}$} & {$\num{0.02}$}\\
    {DeepAR} & {MQCNN}&{$\num{6.70e-11}$} & {$\num{0.01}$} & {$\num{8.20e-07}$} & {$\num{0.13}$} & {$\num{6.53e-07}$} & {$\num{0.13}$} & {$\num{7.39e-11}$} & {$\num{0.01}$}\\
    \midrule
    &&\multicolumn{8}{c}{{ADS$_{cte}$}}\\
    \cmidrule(lr){3-10}
    {TFT} & {Seq2Seq} & {$\num{1.41e-01}$} & {$\num{0.61}$} & {$\num{2.92e-09}$} & {$\num{0.95}$} & {$\num{2.19e-08}$} & {$\num{0.92}$} & {$\num{1.61e-11}$} & {$\num{1.00}$} \\
    {TFT} & {DeepAR} & {$\num{2.58e-05}$} & {$\num{0.82}$} & {$\num{3.01e-11}$} & {$\num{1.00}$} & {$\num{3.01e-11}$} & {$\num{1.00}$} & {$\num{4.96e-11}$} & {$\num{0.99}$} \\
    {TFT} & {MQCNN} & {$\num{9.88e-03}$} & {$\num{0.69}$} & {$\num{4.98e-07}$} & {$\num{0.88}$} & {$\num{1.98e-08}$} & {$\num{0.92}$} & {$\num{7.85e-12}$} & {$\num{1.00}$} \\
    {Seq2Seq} & {DeepAR} & {$\num{2.12e-04}$} & {$\num{0.78}$} & {$\num{5.89e-01}$} & {$\num{0.46}$} & {$\num{2.77e-01}$} & {$\num{0.58}$} & {$\num{1.52e-01}$} & {$\num{0.61}$} \\
    {Seq2Seq} & {MQCNN} & {$\num{6.25e-02}$} & {$\num{0.64}$} & {$\num{9.79e-05}$} & {$\num{0.21}$} & {$\num{3.25e-02}$} & {$\num{0.34}$} & {$\num{6.87e-01}$} & {$\num{0.47}$} \\
    {DeepAR} & {MQCNN} & {$\num{2.97e-01}$} & {$\num{0.42}$} & {$\num{1.49e-06}$} & {$\num{0.14}$} & {$\num{1.74e-08}$} & {$\num{0.08}$} & {$\num{6.60e-03}$} & {$\num{0.30}$} \\
    \bottomrule
\end{tabular}}
\end{table*}

\subsubsection{{ADS Case Study Results}}\label{sec:RQ2-ads-results}
{
Similar to the ACT case study results (\autoref{sec:RQ2-act-results}), \autoref{fig:rq2-fn-ads-cte} and \autoref{fig:rq2-fp-ads-cte}, indicate that 
FN decreases and FP increases with increase in prediction quantile.
As explained in \autoref{sec:RQ2-act-results}, we will focus on the results for prediction quantiles $q\geq0.5$.
We report the averages and $95\%$ confidence intervals of the FN, FP, Precision, Recall and F\textsubscript{3} score values in \autoref{tab:rq2-ads-results}.

Overall, in line FP and FN trends, Precision for all models drops when $q$ increases, while Recall increases, as discussed in \autoref{sec:RQ2-act-results}.
Note that TFT and MQCNN yield the highest Recall score of $0.985$ at $q=0.995$, while for the same quantile, MQCNN yields the lowest Precision score.
We further observe for $q\geq0.5$, TFT is the second best performing model in terms of Precision score while yielding the highest Recall score.
This results in TFT achieving the highest F\textsubscript{3} score over all quantiles $q\geq0.5$.
Moreover, we observe that DeepAR yields the lowest Recall and F\textsubscript{3} score over all quantiles $q\geq0.5$.

This case study thus confirms the superiority of TFT over Seq2Seq, MQCNN, and DeepAR, when $q>0.5$, in terms of F\textsubscript{3} score, based on the results of our statistical analysis, reported in \autoref{tab:rq2-stat-test}.
We further observe that, when $q>0.5$, TFT is significantly better than other models with a high effect size.
However, for $q=0.5$, we observe that TFT is not significantly better than Seq2Seq, though it is still outperforming DeepAR and MQCNN significantly.
Moreover, the effect size of the difference between the F\textsubscript{3} scores of TFT and MQCNN when $q=0.5$, i.e., $\hat{A}_{AB}$, when $A=\text{TFT}$ and $B=\text{MQCNN}$, is $0.64<\hat{A}_{AB}=0.69<0.71$, indicating a medium effect size.
}

\subsubsection{{Discussion}}\label{sec:RQ2-discussion}
{
Based on the results of the ACT case study (\autoref{sec:RQ2-act-results}),
where the dataset is large and includes numerous safety violations,
we conclude that, for probabilistic prediction of the safety violation, at all measured quantiles q, with again a hazard forecast horizon of \SI{3}{s} and a lookback to forecast horizon ratio of 3, as discussed in \autoref{sec:RQ1-act-results}, TFT is significantly more accurate than Seq2Seq, DeepAR, and MQCNN, for both $cte$ and $he$ safety requirements.

Based on the ADS case study results (\autoref{sec:RQ2-ads-results}),
where the size of the dataset is much smaller than the ACT dataset,
we observe that, given the hazard forecast horizon of \SI{3}{s} (equal to the minimum reaction time) and a lookback to forecast horizon ratio of 1, as discussed in \autoref{sec:RQ1-method}, TFT is significantly more accurate that Seq2Seq, DeepAR, and MQCNN with a large effect size, when $q>0.5$.
At $q=0.5$, TFT and Seq2Seq both yield the most accurate predictions,
while significantly outperforming DeepAR and MQCNN with a large and medium effect size, respectively.
Therefore, we conclude that TFT is the most suitable model,
among all evaluated models,
for safety violation prediction using safety metric forecasting,
for all reported quantiles $q\geq0.5$.
Predictions for $q\leq0.5$ are in any case not sufficiently accurate regardless of the model employed and therefore comparisons for such q values are not of practical utility. 
}

\begin{tcolorbox}
{
For the ACT case study,
where the size of the dataset is large,
given a hazard forecast horizon of \SI{3}{s}
and a lookback to a forecast horizon ratio of 3 for safety violation prediction (minimum reaction time),
TFT is significantly more accurate, with a large effect size, than Seq2Seq, DeepAR, and MQCNN for all quantiles $\text{q}\geq0.5$, for both the $cte$ and $he$ safety requirements.

For the ADS case study,
where the datset size is much smaller,
given a practical window configuration, i.e., $h=\SI{3}{s}$ and $cm=1$,
TFT is the most suitable model,
among all evaluated models,
for probabilistic safety metric forecasting,
for all quantiles $q\geq0.5$. Predictions for $q\leq0.5$ are poor for all models. 

Therefore, when $q > 0.5$, TFT is consistently the best model for both case studies.
}
\end{tcolorbox}

\subsection{\texorpdfstring{RQ\textsubscript{3}}{RQ3}: Prediction Accuracy Sensitivity Analysis}\label{sec:RQ3}

{
In \autoref{sec:RQ3_method}, we provide the details of our evaluation methodology as it relates to answering RQ\textsubscript{3}.
As discussed in \autoref{sec:RQ1-method}, due to the low number of samples available in the dataset of the ADS case study, we were only able to study the effect of varying window sizes on prediction accuracy for the two safety requirements of the ACT case study, for which we report the results in \autoref{sec:RQ3-results}.
Finally, we present our answer to RQ\textsubscript{3}, based on the results presented in \autoref{sec:RQ3-results}.
}

\subsubsection{Methodology}\label{sec:RQ3_method}

We have answered RQ\textsubscript{1} and RQ\textsubscript{2} based on the minimum reaction time (\SI{3}{s}) for hazard forecast horizon and a commonly used lookback horizon that is three times longer (\SI{12}{s}).
However, the size of the hazard forecast and lookback horizons are design choices of the system developers.
To investigate the impact of window configuration, for each forecasting model, in terms of safety metric and violation accuracy, inference latency, and computations resource usage, we also assessed the tuned models with different hazard forecast horizons and lookback-to-forecast horizon values.

Particularly, we selected the hazard forecast horizon from \{3, 12\} and the ratio of lookback to forecast horizon window size, also known as \emph{context multiplier} ($cm$), from \{1, 3, 9\}.
Thus, we studied the following forecast horizon window size and context multiplier combinations ($h$, $cm$):
$(3,1)$, $(3,3)$, $(3,9)$, $(12,1)$, $(12,3)$, and $(12,9)$.
Note that the smallest total window size\footnote{$\text{Total window size} = h + (cm \times h)$} is $3 + 1\times3 = \text{\SI{6}{s}}$, whereas the largest total window size is $12 + 9\times12 = \SI{120}{s}$.
In our preliminary experiments, we observed that increasing the forecast horizon increases the likelihood that samples include safety violations.
We further observed that for forecast horizons longer than \SI{12}{s}, the distribution of test samples becomes highly imbalanced, leading to biased comparisons of safety violation prediction accuracy metrics between short ($h=\SI{3}{s}$) and very long ($h>\SI{12}{s}$) forecast horizons.
Furthermore, we would not have been able to study the effect of varying $cm$ on very large forecast horizons since their total window size on higher $cm$ values would become longer than the maximum training sample length.
Thus, for the evaluation of RQ\textsubscript{3}, we did not include forecast horizons larger than \SI{12}{s}.
The results for a representative instance of our preliminary experiments with long forecast horizons, i.e., a forecast horizon of \SI{36}{s} and context multiplier of 1, are included in our supporting material (see \autoref{sec:data-availability}).

\subsubsection{Results}\label{sec:RQ3-results}

As discussed in \autoref{sec:RQ2-act-results}, we are interested in the predictions at quantiles that primarily detect as many hazards as possible while having a low number of false alarms.
Due to the safety-critical nature of our problem, we only focus here on predictions at $\text{q}=0.995$ since it is the most conservative measured prediction quantile (\autoref{sec:RQ2-act-results}).
Recall that when comparing models, a lower q-Risk value implies a more accurate model at predicting safety metric values.

\begin{figure*}
\begin{subfigure}{0.9\textwidth}
  \centering
  \includegraphics[width=\textwidth]{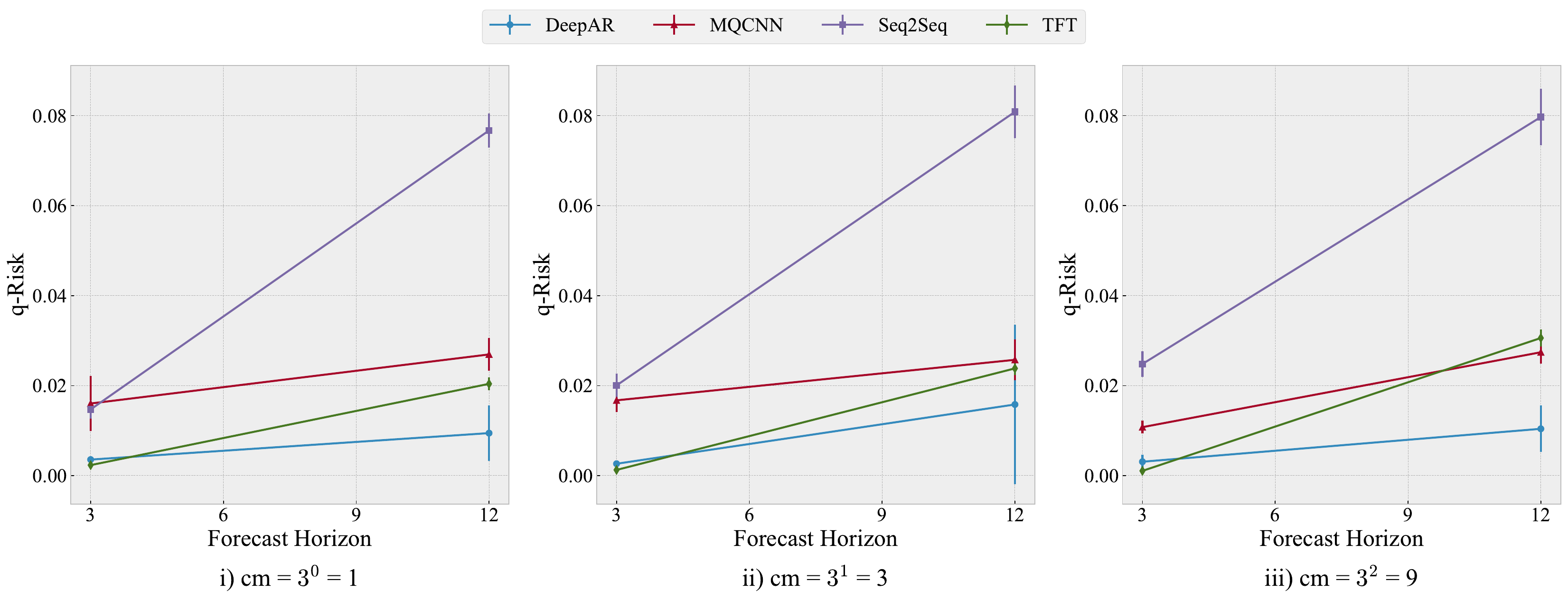}
  \caption{{ACT$_{cte}$}}
  \label{fig:rq3-qrisk_act_cte}
\end{subfigure}
\hfill
\begin{subfigure}{0.9\textwidth}
  \centering
  \includegraphics[width=\textwidth]{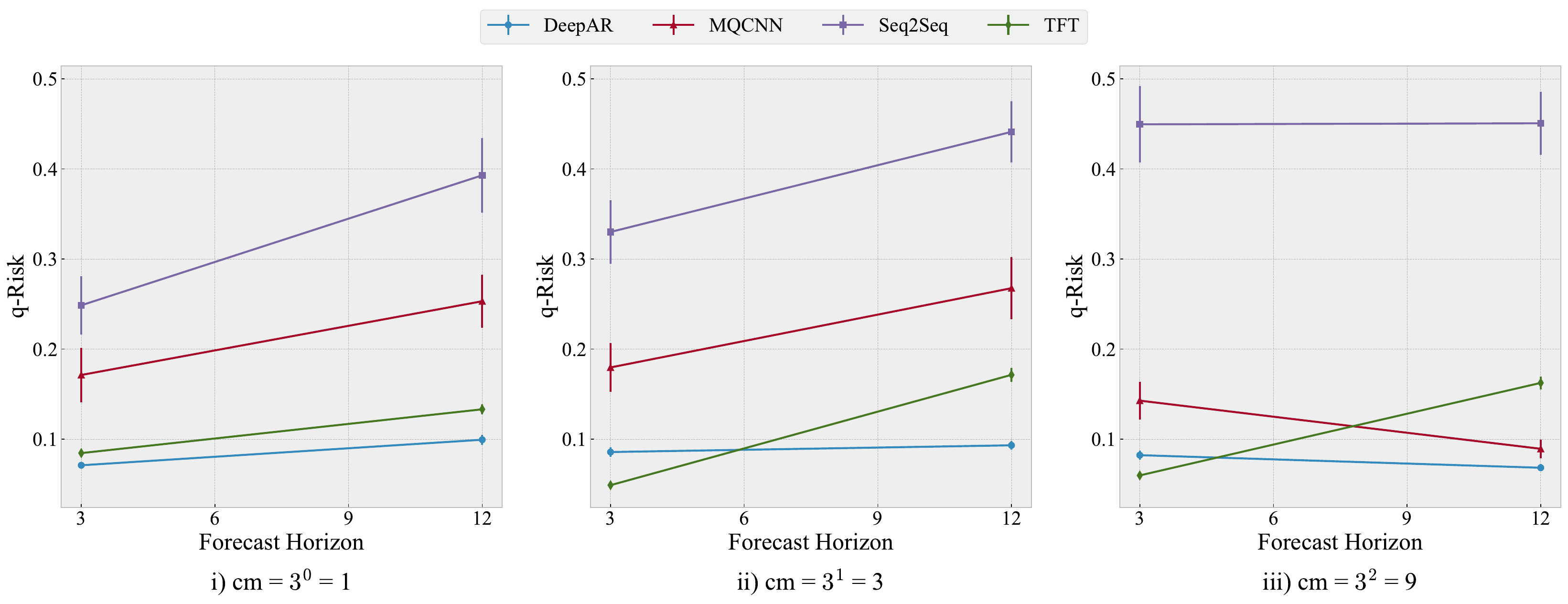}
  \caption{{ACT$_{he}$}}
  \label{fig:rq3-qrisk_act_he}
\end{subfigure}
\Description{Three subplots of q-Risk versus forecast horizon, each for a fixed cm value of 1, 3 and 9. All methods show increase in q-Risk with increase in forecast horizon.}
\caption{Safety metric prediction accuracy metrics for various window configurations of DeepAR, MQCNN, Seq2Seq and TFT{, for the $cte$ and $he$ safety requirements, respectively}.}
\label{fig:rq3-qrisk}
\end{figure*}

From {\autoref{fig:rq3-qrisk_act_cte}}, we observe that for {the $cte$ safety requirement, given} a fixed $cm$ value, increasing the forecast horizon {$h$ }leads to higher q-Risk values and thus lower accuracy.
Further, we can also see that, in contrast, for a fixed forecast horizon, increasing $cm$ does not significantly improve q-Risk.
We observe that TFT outperforms all models (i.e., has the lowest q-Risk value) at the forecast horizon of 3, across all $cm$ values.
At longer forecast horizons ($h=12$), DeepAR outperforms all other models.
Moreover, note that due to the iterative forecasting architecture of DeepAR, it is prone to accumulating forecasting errors.
Thus, its confidence interval increases significantly with the increase in forecast horizon.
For instance, at $cm=3$, the confidence interval of DeepAR includes the q-Risk values for both TFT and MQCNN.

{
For the $he$ safety requirement (\autoref{fig:rq3-qrisk_act_he}), similar to $cte$, we observe that given a fixed $cm$ value, increasing $h$ leads to higher q-Risk values, except for MQCNN at $cm=9$, where q-Risk decreases when increasing the forecast horizon.
}

\begin{figure*}
\begin{subfigure}{0.9\textwidth}
  \centering
  \includegraphics[width=\textwidth]{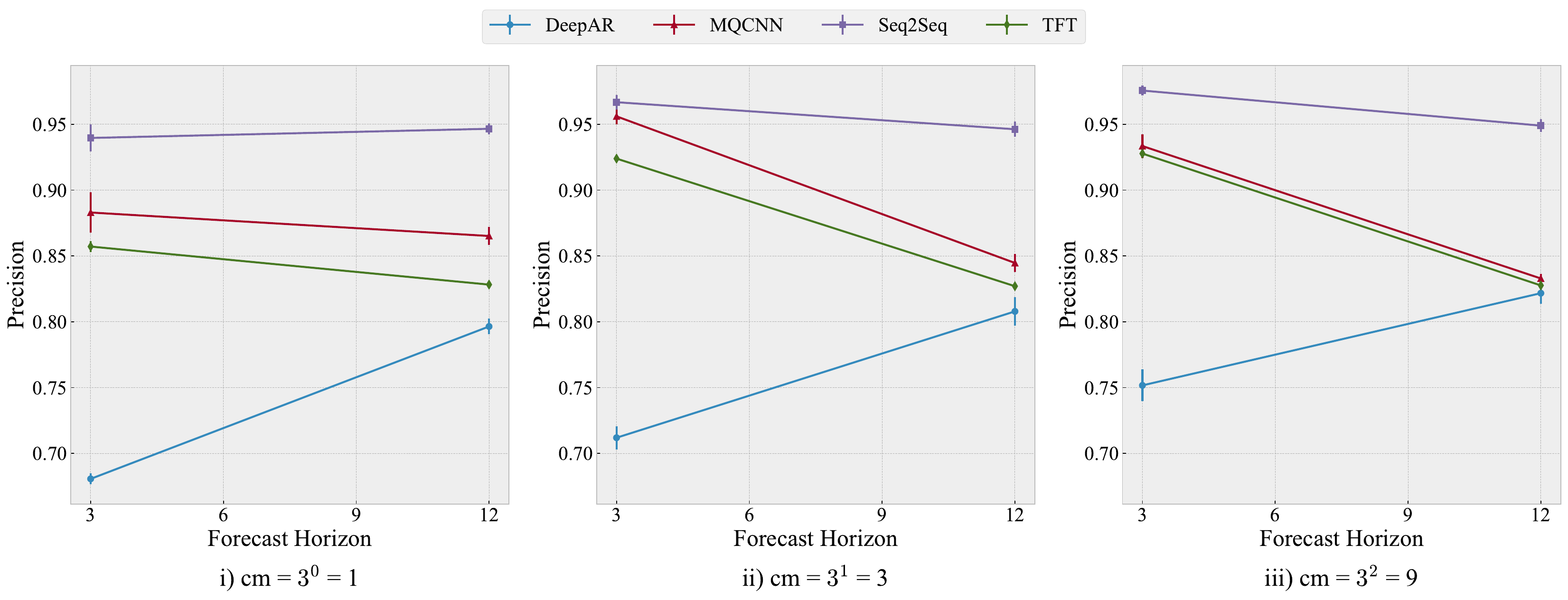}
  \caption{Precision {--- ACT$_{cte}$}}
  \label{fig:rq3-precision-act-cte}
\end{subfigure}
\hfill
\begin{subfigure}{0.9\textwidth}
  \centering
  \includegraphics[width=\textwidth]{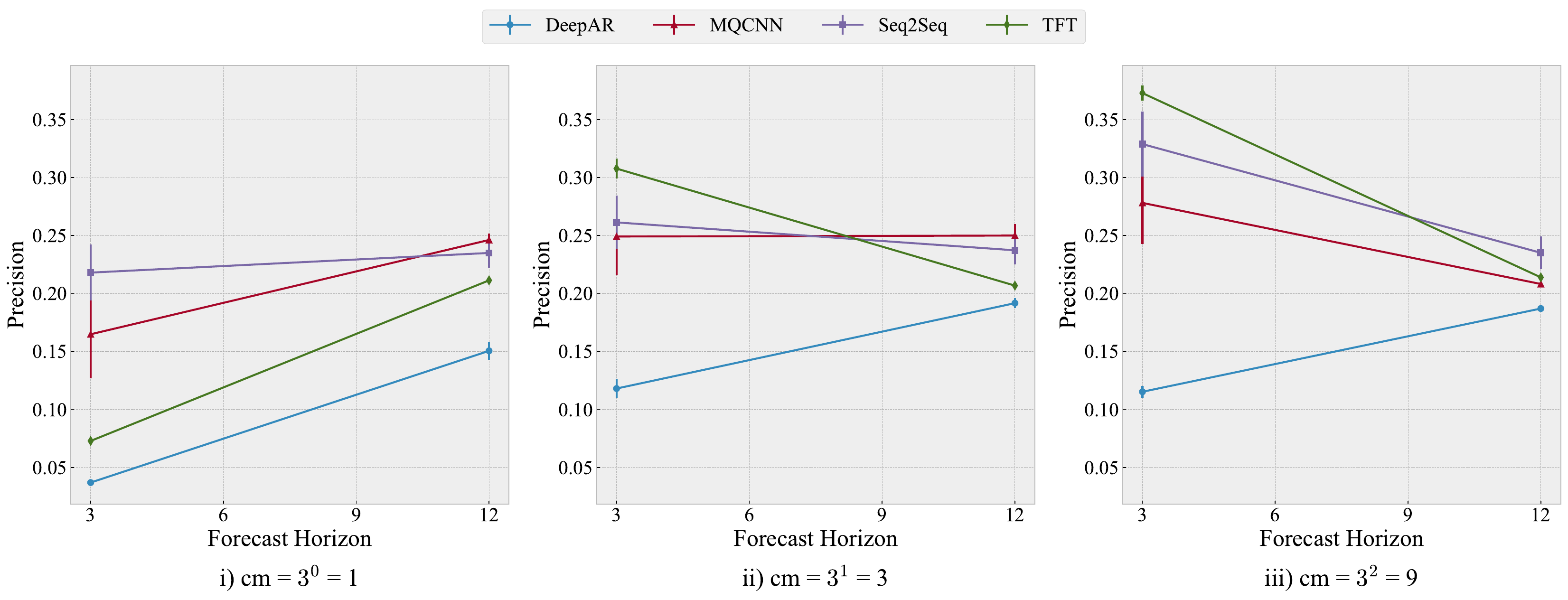}
  \caption{{Precision --- ACT$_{he}$}}
  \label{fig:rq3-precision-act-he}
\end{subfigure}
\Description{A grid of ...}
\caption{{Precision score measurements for DeepAR, MQCNN, Seq2Seq, and TFT, in various window configurations, for the $cte$ (a) and $he$ (b) safety requirements.}}
\label{fig:rq3-precision}
\end{figure*}

In terms of precision {for the $cte$ safety requirement} (\autoref{fig:rq3-precision-act-cte}), we observe that when increasing forecast horizon, given a constant $cm$ value, the precision of models with a sequence-to-sequence architecture (TFT, MQCNN, and Seq2Seq)  drops while DeepAR's precision increases, most particularly at $cm=1$.
{A similar trend is observed for the $he$ safety requirement (\autoref{fig:rq3-precision-act-he}) with the difference that the models reach a lower precision score than similar models evaluated on the $cte$ safety requirement data. 
This can be explained by the lower proportion of samples including safety violations for $he$, when compared to $cte$ (as discussed in \autoref{sec:RQ1-act-results}).}
Further note that{, for both $cte$ and $he$, given} a constant forecast horizon, increasing $cm$ can slightly improve the precision of the model, most particularly for DeepAR.

\begin{figure*}
\begin{subfigure}{0.9\textwidth}
  \centering
  \includegraphics[width=\textwidth]{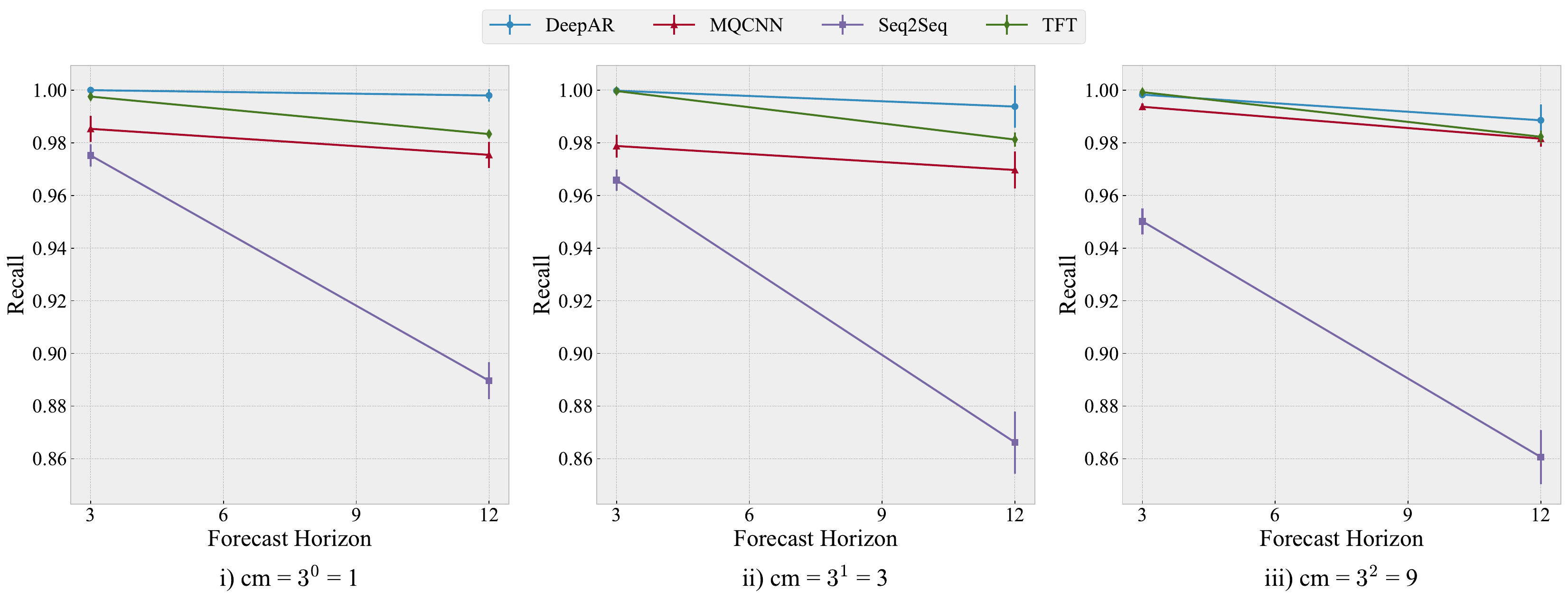}
  \caption{Recall {--- ACT$_{cte}$}}
  \label{fig:rq3-recall-act-cte}
\end{subfigure}
\hfill
\begin{subfigure}{0.9\textwidth}
  \centering
  \includegraphics[width=\textwidth]{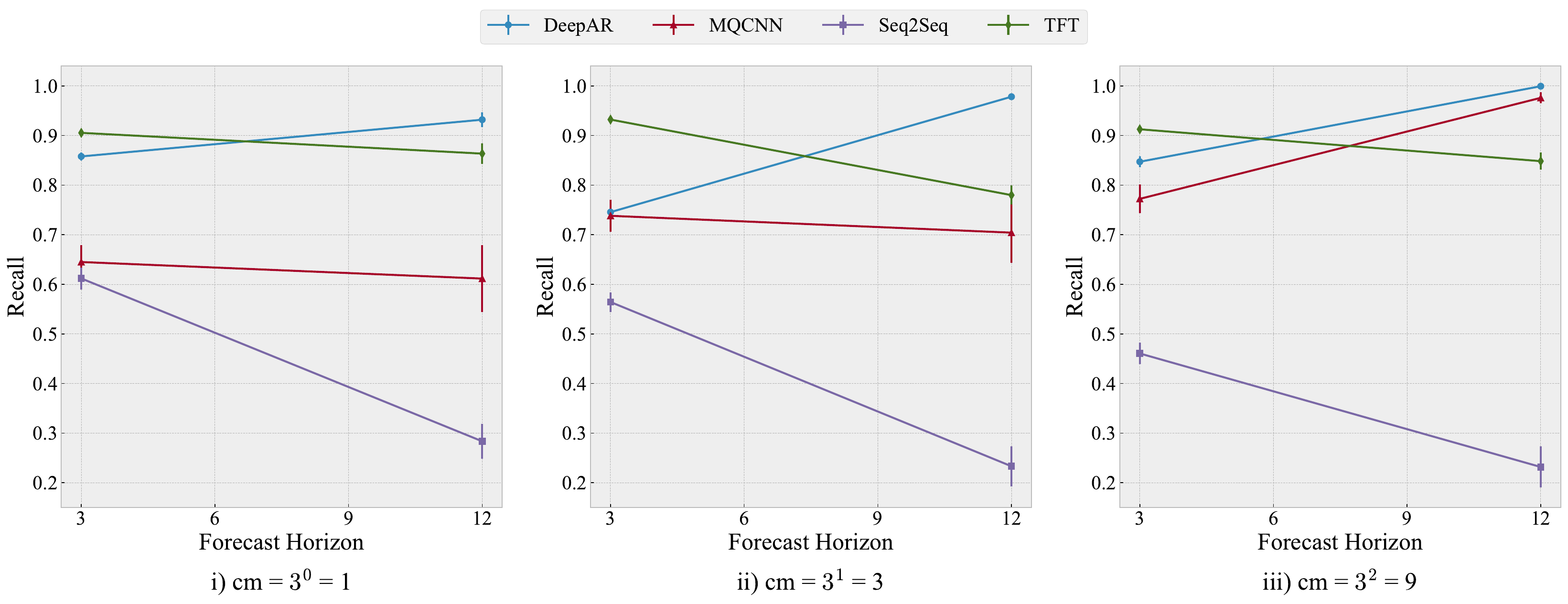}
  \caption{{Recall --- ACT$_{he}$}}
  \label{fig:rq3-recall-act-he}
\end{subfigure}
\Description{A grid of ...}
\caption{{Recall score measurements for DeepAR, MQCNN, Seq2Seq, and TFT, in various window configurations, for the $cte$ (a) and $he$ (b) safety requirements.}}
\label{fig:rq3-recall}
\end{figure*}

In terms of recall {for the $cte$ safety requirement} (\autoref{fig:rq3-recall-act-cte}), we observe that all models yield a lower recall when increasing forecast horizon while increasing $cm$ does not significantly improve recall given the same forecast horizon.
Note that TFT and DeepAR similarly reach the highest recall at $h=3$.
Whereas, at $h=12$, TFT experiences a larger drop in recall and is outperformed by DeepAR.
Concerning the effect of $cm$, we observe that MQCNN experiences the most improvement when $cm$ changes from 3 to 9.
At the same time, we observe that the recall score of DeepAR and Seq2Seq drops when $cm$ increases given a fixed forecast horizon.
Finally, the recall score of TFT is not significantly affected by changes in $cm$ values.
{Similarly, for the $he$ safety requirement, TFT reaches the highest recall score at $h=3$. 
However, instead of all models experiencing a drop in recall score with increases in forecast horizon, the recall score for DeepAR increases in all $cm$.
MQCNN also experiences a recall score increase with an increase in $h$ at $cm=9$.
To conclude, the effect of $cm$ is similar to that observed for the $cte$ safety requirement, with the exception of the differences mentioned above.}

\begin{figure*}
\begin{subfigure}{0.9\textwidth}
  \centering
  \includegraphics[width=\textwidth]{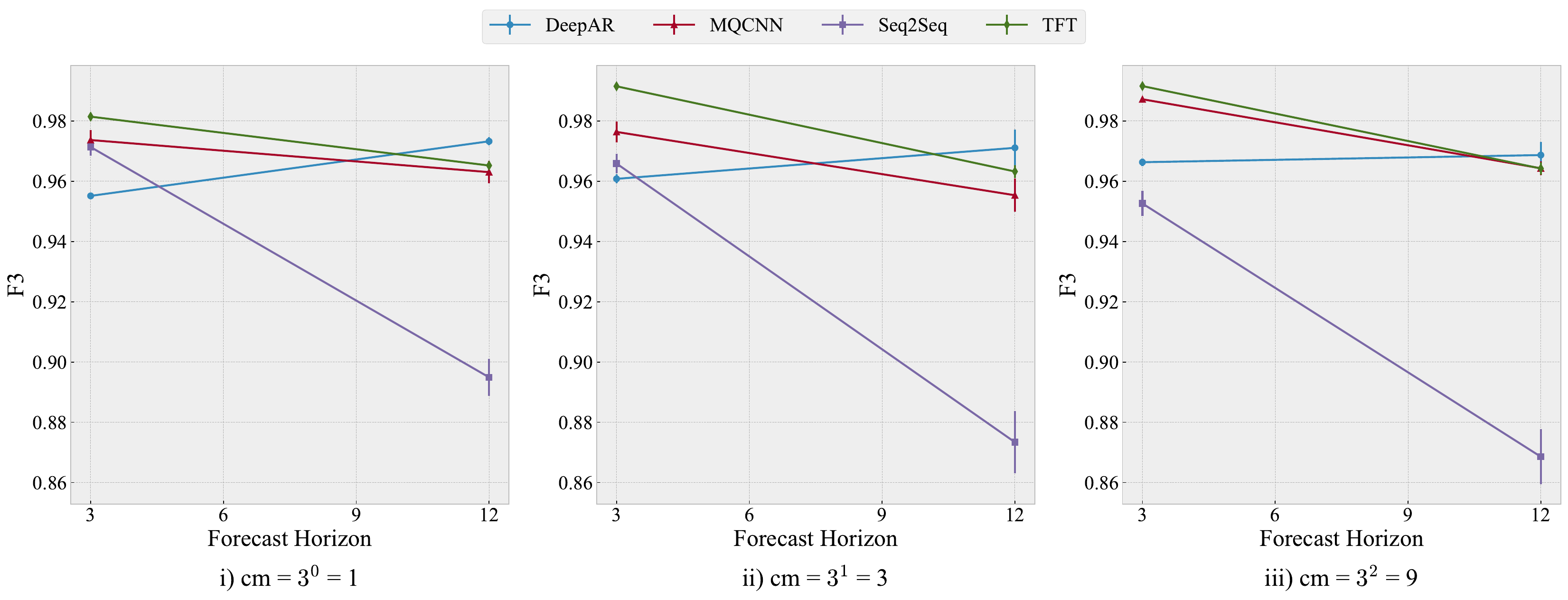}
  \caption{F\textsubscript{3} score {--- ACT$_{cte}$}}
  \label{fig:rq3-f3-act-cte}
\end{subfigure}
\hfill
\begin{subfigure}{0.9\textwidth}
  \centering
  \includegraphics[width=\textwidth]{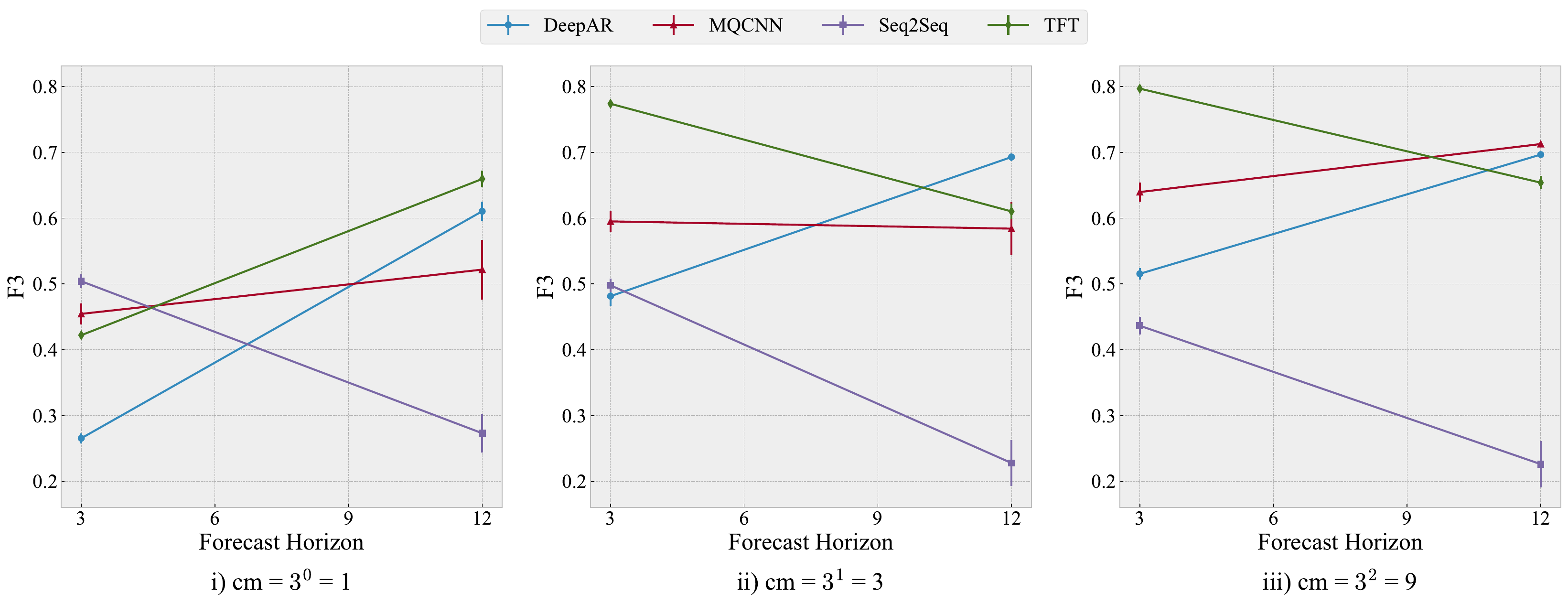}
  \caption{{F\textsubscript{3} score --- ACT$_{he}$}}
  \label{fig:rq3-f3-act-he}
\end{subfigure}
\Description{A grid of 6 F3 score subplots, 3 for the cte and 3 for the he safety requirements of the ACT cases study.}
\caption{{F\textsubscript{3} score measurements for DeepAR, MQCNN, Seq2Seq, and TFT, in various window configurations, for the $cte$ (a) and $he$ (b) safety requirements.}}
\label{fig:rq3-f3}
\end{figure*}

In terms of overall safety violation prediction accuracy, i.e., F\textsubscript{3} (\autoref{fig:rq3-f3}), {for both $cte$ and $he$ safety requirements,} we observe that TFT yields the highest F\textsubscript{3} score when $h=3$ for all values of $cm${, except for $he$ when $cm=1$.
However, we observe that F\textsubscript{3} is much lower than the scores reached for other $cm$ values, suggesting that the small lookback horizon does not contain sufficient information for the models to generate accurate forecasts.}
Note that for the same window configuration, DeepAR yields the lowest F\textsubscript{3} score, which is mainly due to its very low precision.
Nevertheless, given the increase in precision and a smaller decrease in recall {for $cte$, as well as an increase for $he$)}, the F\textsubscript{3} score of DeepAR for configurations with $h=12$ at any $cm$ value rises to become the best model.
However, note that the highest F\textsubscript{3} score at $h=12$ is still lower than the highest F\textsubscript{3} score reached by TFT at $h=3$, suggesting that increasing the forecast horizon $h$ leads to lower safety violation prediction accuracy in general.


\begin{tcolorbox}
Given a hazard forecast horizon of \SI{3}{s}, {for both $cte$ and $he$ safety requirements of the ACT case study,}
TFT yields the most accurate safety metric and safety violation predictions, with an improving accuracy when the lookback horizon length increases ($cm \times h$).
{For the $he$ safety requirement at $cm=1$, the lookback horizon does not contain sufficient information for any model to generate accurate forecasts.}
We further conclude that for prediction horizons longer than the minimum reaction time, i.e., $h=\SI{12}{s}$), regardless of the $cm$ value, DeepAR yields the most accurate predictions.
\end{tcolorbox}

\subsection{\texorpdfstring{RQ\textsubscript{4}}{RQ4}: Latency and Memory Overhead}

{
In section \autoref{sec:RQ4-method}, we provide the details of our evaluation methodology for answering RQ\textsubscript{4}.
As discussed in the beginning of \autoref{sec:RQ3}, we were only able to evaluate the effect of varying window sizes, on runtime performance, for the two safety requirements of the ACT case study (we present the results in \autoref{sec:RQ4-results}).
We also measured the runtime performance of the ADS case study, when $h=\SI{3}{s}$ and $cm=1$, which is also discussed in \autoref{sec:RQ4-results}.
We finally provide our answer to RQ\textsubscript{4} based on the results discussed in \autoref{sec:RQ4-results}.
}

\subsubsection{Methodology}\label{sec:RQ4-method}

To answer RQ\textsubscript{4}, we queried each of the models trained in RQ\textsubscript{3} over the whole test set {for both $cte$ and $he$ safety requirements of the ACT case study,} and collected the average latency (in milliseconds \textit{ms}) and peak GPU memory\footnote{Recall that, as mentioned in \autoref{sec:eval-hardware}, we use a GPU to train the models and generate predictions.} usage during inference (in megabytes, \textit{MB}).
Furthermore, we measure the model size in terms of GPU memory usage (in megabytes, \textit{MB}), by measuring the difference in GPU memory usage before and after a model is loaded to the GPU.
Note that the base ML libraries are preloaded in the GPU and their GPU memory usage is not reported.
{For the ADS case study, we applied the above process to the models that were trained in RQ\textsubscript{1} ($h=\SI{3}{s}$ and $cm=1$).}

\subsubsection{Results}\label{sec:RQ4-results}
\begin{figure*}
\begin{subfigure}{0.9\textwidth}
  \centering
  \includegraphics[width=\textwidth]{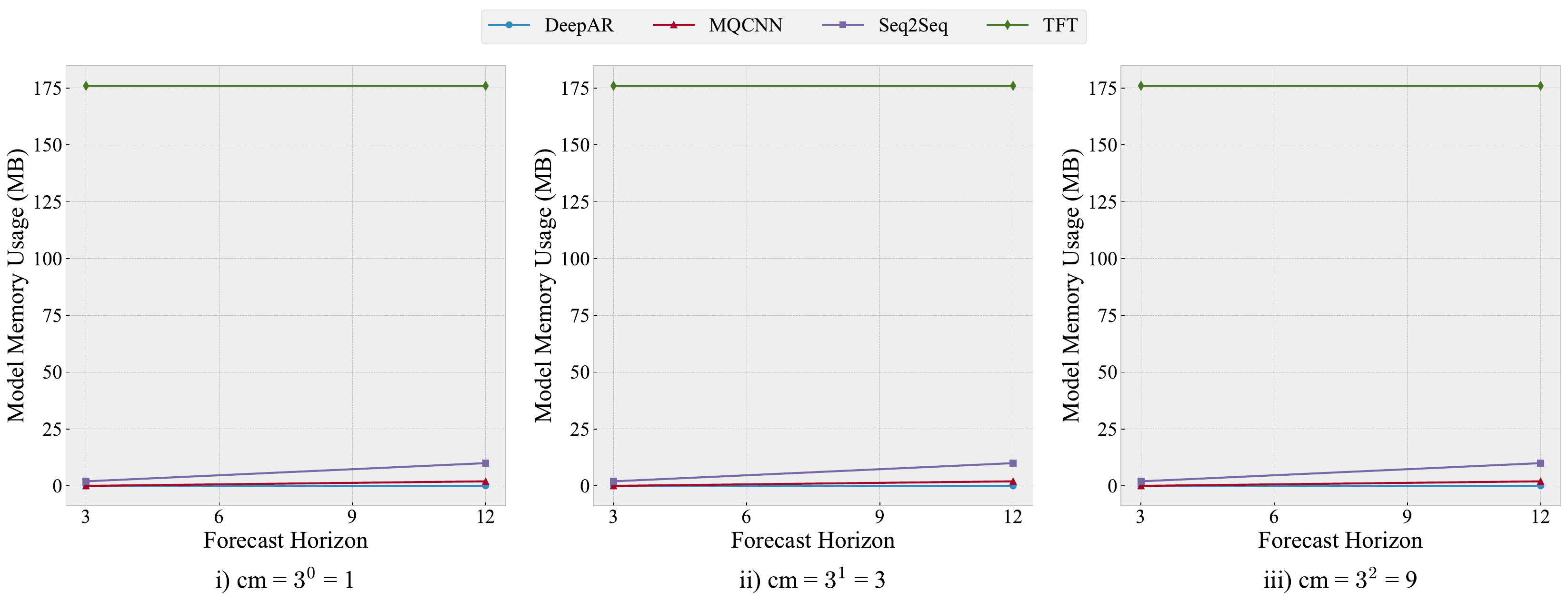}
  \caption{Model Memory Usage (MB)}
  \label{fig:rq4-model-mem}
\end{subfigure}
\hfill
\begin{subfigure}{0.9\textwidth}
  \centering
  \includegraphics[width=\textwidth]{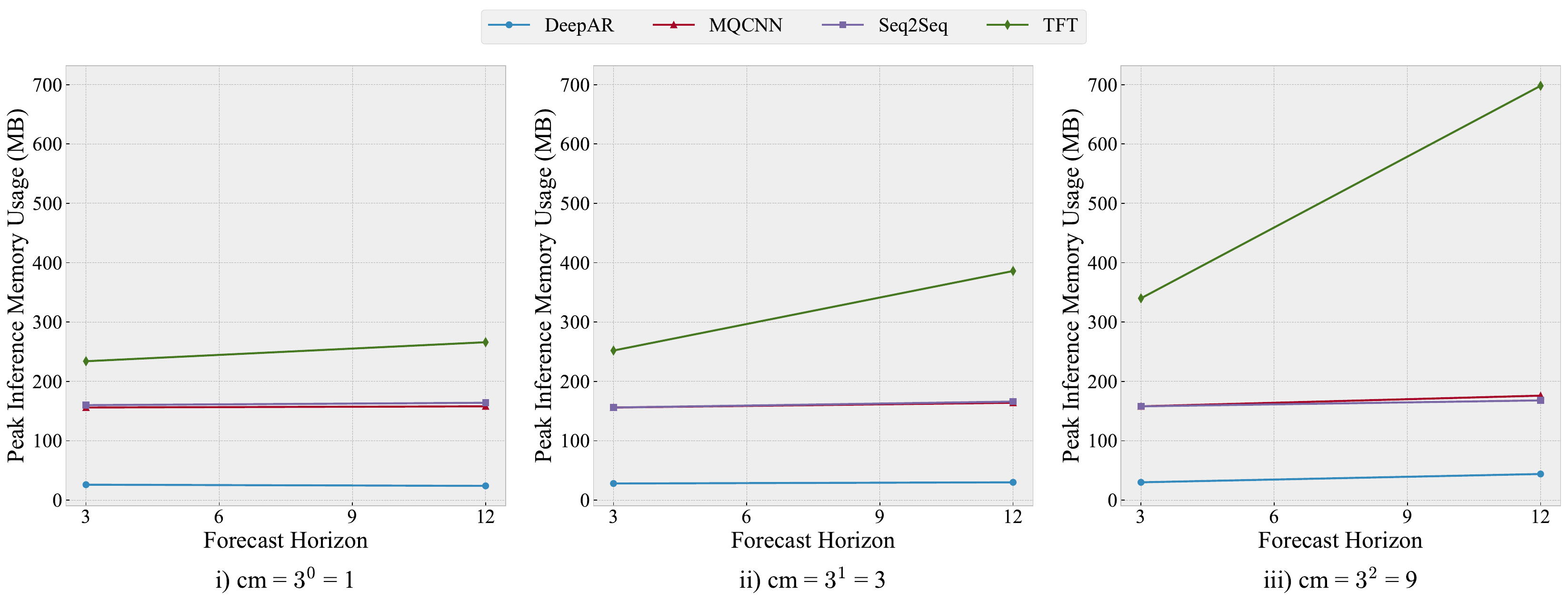}
  \caption{Peak Inference Memory Usage (MB)}
  \label{fig:rq4-peak-mem}
\end{subfigure}
\hfill
\begin{subfigure}{0.9\textwidth}
  \centering
  \includegraphics[width=\textwidth]{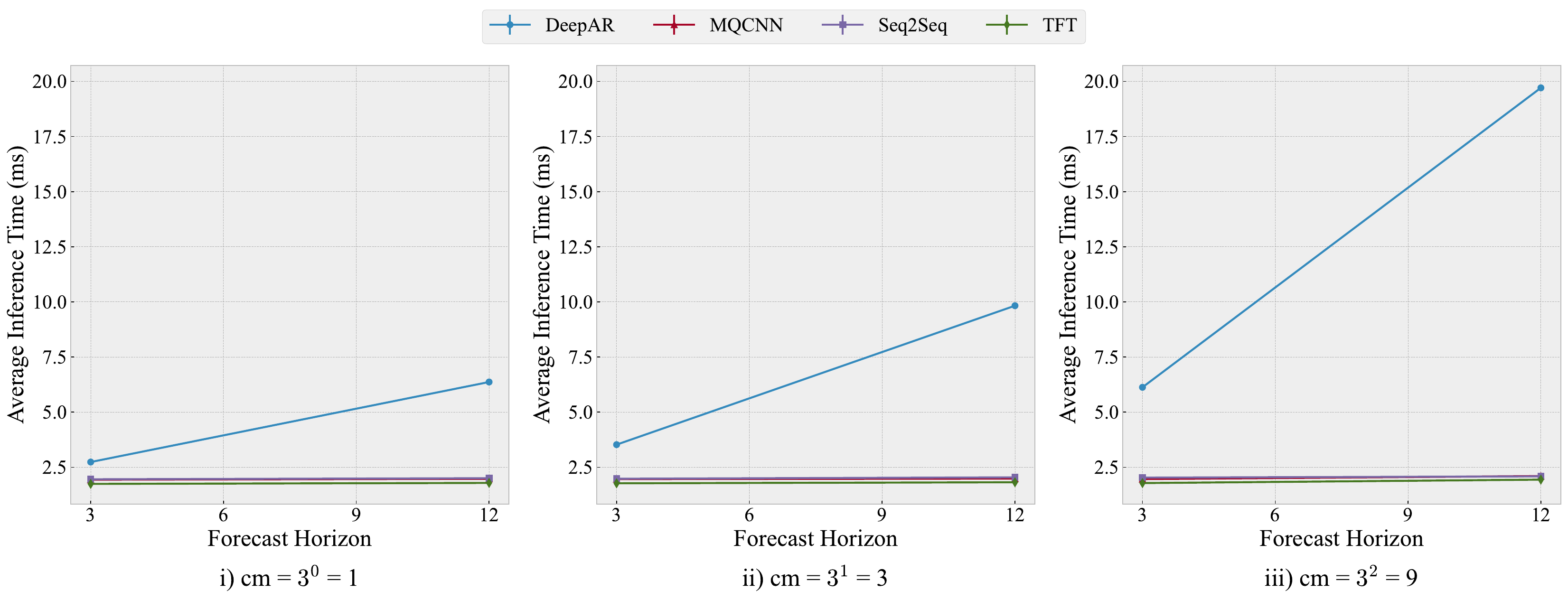}
  \Description{A grid of nine subplot, 3 for each of model memory usage, peak inference memory usage and average inference latency.}
  \caption{Average Inference Latency (ms)}
  \label{fig:rq4-avg-lat}
\end{subfigure}
\caption{The plot for a) model memory usage, b) peak inference memory usage, and c) average inference latency for DeepAR, MQCNN, Seq2Seq, and TFT models at different window configurations{, for the $cte$ safety requirement of the ACT case study, which is similar to the results for the $he$ safety requirement. The above results at forecast horizon of $\SI{3}{s}$ and $cm=1$ are also similar to the results for the ADS case study.}}
\label{fig:rq4-v-predLen}
\end{figure*}

{The RQ\textsubscript{4} evaluation results for the $cte$ safety requirement of the ACT case study are similar to those for its $he$ safety requirement.
This is to be expected, as the size of the input and output vectors for the same model do not change between the two safety requirements.
The only difference between them is due to differences in the hyperparameter values selected to best address each safety requirement, which in turn can change the number of learnable parameters in the model.
However, as observed, such change has not substantially changed the results.
Therefore, although we present the results for the $cte$ safety requirement of the ACT case study (\autoref{fig:rq4-v-predLen}) in the paper, our discussion of the results and conclusions similarly hold for the $he$ safety requirement.
Moreover, we have observed that the ADS case study results, are not substantially different from the ACT case study results for the same window configuration, i.e., when $h=\SI{3}{s}$ and $cm=1$.
Thus, the discussion of the results and conclusions for the $cte$ safety requirement of the ACT case study, given a similar window configuration, i.e., when $h=\SI{3}{s}$ and $cm=1$, similarly holds for the ADS case study.
The figures and results for the ADS case study, and the $he$ safety requirement of the ACT case study, are available in our replication package (\autoref{sec:data-availability}).}

As shown in \autoref{fig:rq4-model-mem}, all models in all window configurations have low GPU memory usage. TFT, which has the highest usage of all, only consumes around 175 MB.
We further observe that an increase in context multiplier does not lead to a significant increase in model memory usage, while an increase in forecast horizon leads to slight increases in MQCNN and Seq2Seq model memory usage.

From \autoref{fig:rq4-peak-mem}, we observe that an increase in forecast horizon leads to increased peak memory usage during inference.
When increasing $cm$, the peak memory usage of the models does not increase significantly for the same forecast horizon, except for TFT where the rate of increase in peak memory usage with increasing forecast horizon grows at higher $cm$ values.
Nevertheless, the largest peak memory usage during inference, which we observe for TFT when $h=\SI{12}{s}$ and $cm=9$, is 700 MB, only consuming $17.5\%$ of the available memory of an NVIDIA Jetson Nano GPU, the \emph{least} powerful embedded GPU made by NVIDIA.
Therefore, we conclude that all the evaluated models, for all $h$ and $cm$ combinations, yield practical GPU memory usage in terms of model size and peak inference memory usage.

Finally, \autoref{fig:rq4-avg-lat} suggests that the average inference latency for sequence-to-sequence forecasting models (MQCNN, Seq2Seq, and TFT) does not significantly change when increasing the forecast horizon or context multiplier.
However, as expected, DeepAR's inference latency linearly increases with forecast horizon.
Furthermore, when increasing $cm$, DeepAR's prediction latency increases for the same forecast horizon.
Thus, for longer prediction horizons or higher context multipliers, DeepAR is prone to having a high inference latency which could render its use prohibitive in the context of safety-critical systems.
In general, for a safety monitor to be effective it should be able to predict safety violations and raise an alarm before the planning or control modules update their command.
In our ACT example, the system does not include a planner and the controller directly generates control commands based on the perception system outputs.
Thus, in our ACT case study, the use of a safety monitor is only meaningful if its average inference latency is less than the controller cycle time, i.e., the time period between two generated control commands.
For practical reference, the design requirement for the maximum cycle time of planning and control modules at the Indy Autonomous Challenge~\cite{IAC}, is \SI{10}{ms}~\cite{jung2023IAC}.\footnote{The Indy Autonomous Challenge is an international challenge where teams from universities develop autonomous racing vehicles with the ultimate goal of improving the safety and performance of autonomous driving technology~\cite{IAC}.}
{
In autonomous aviation, \citet{paredes2024neuromorphic} and \citet{navrdi2022latency} report an average vision cycle latency of \SI{12}{ms} and \SI{13}{ms}, respectively on an NVIDIA Jetson GPU.
In another example, for a real-time vision-based drone system developed by \citet{farrukh2023flyos}, the authors empirically measure and report that the average latency for the vision pipeline, where the safety monitor should have an average \SI{16}{ms} or less to be useful.
}
Note that DeepAR with a forecast horizon of 12, reaches the maximum latency of \SI{10}{ms} at $cm=3$ and largely exceeds it at $cm=9$.
In contrast, sequence-to-sequence models maintain a constant average inference latency of approximately \SI{2}{ms} for all $h$ and $cm$ values.
This is expected since sequence-to-sequence models generate their forecast for all forecast horizon timesteps at once.

\begin{tcolorbox}
{For both $cte$ and $he$ safety requirements in the ACT case study, a}ll models in all configurations yield practical model size and peak inference memory usage.
Furthermore, although MQCNN, Seq2Seq, and TFT exhibit a constant and very low inference latency, DeepAR's inference latency significantly increases with the forecast horizon and context multiplier, which can render DeepAR impractical for longer forecast horizons at $cm > 1$.

{For the ADS case study, all models, in a practical window configuration of $h=\SI{3}{s}$ and $cm=1$, yield practical model size, peak inference memory usage and inference latency.}
\end{tcolorbox}

\subsection{Discussion}\label{sec:discussion}

\paragraph{Safety Monitoring via Safety Metric Forecasting}
Overall, the results of our study suggest that safety metric forecasting, given learned component outputs and scenarios, is effective for safety monitoring. Indeed, the models, {when evaluated on a dataset with a balanced distribution of safety violations, i.e., the $cte$ safety requirement of the ACT case study, }have yielded F\textsubscript{3} scores above $95\%$ for all $cm$ and $h$ combinations, except for Seq2Seq at $h=\SI{12}{s}$.\footnote{Recall that by safety monitoring, we refer to runtime monitoring of learned components and the system operational context to predict a system safety requirement violation, which is, different from predicting when a learned component might mispredict.}
However, the fact that all the evaluated models have high F\textsubscript{3} scores{, for the $cte$ safety requirement in the ACT case study,} does \emph{not} mean that all of them are equally accurate in predicting safety violations.
This is highlighted in \autoref{tab:rq2-act-results}, where $1$-$2\%$ differences in F\textsubscript{3} scores{, for the $cte$ safety requirement, }translate into two orders of magnitude increase in the number of false negatives (FN).
{Furthermore, for both the $he$ safety requirement of the ACT case study, where the distribution of safety violations in the dataset contains less safety violations compared to $cte$, and the ADS case study, where the size of the dataset is substantially smaller than that of the ACT case study, we observe that at least one model yields an F\textsubscript{3} score of $77\%$ or more, for $q>0.5$.
This suggests that training the DL-forecasting models to achieve high safety violation prediction accuracy, collecting a large number of diverse safety violations in the dataset is crucial.
}

Moreover, we observe that the ranking of evaluated models in terms of safety metric accuracy (q-Risk) matches the ranking of models in terms of recall scores, suggesting that q-Risk values could be indicators of the recall scores for safety violation prediction (compare \autoref{fig:rq3-qrisk} with \autoref{fig:rq3-recall}).
However, we observe that the same order does not hold for precision and F\textsubscript{3} scores.
For instance, MQCNN and Seq2Seq have higher (q-Risk) values than DeepAR for $h=\SI{3}{s}$ (\autoref{fig:rq3-qrisk}), whereas their precision and F\textsubscript{3} scores (\autoref{fig:rq3-precision} and \autoref{fig:rq3-f3}, respectively) are significantly higher.
Thus, we conclude that although q-Risk values, which are readily available after training the models and testing them on the test dataset, might be an indicator of the recall score for safety violation prediction, they are not indicators of precision and overall accuracy (F\textsubscript{3}) scores for safety violation prediction.
Thus, in practice, one needs to compute precision and F\textsubscript{3} scores before choosing a model for runtime deployment, and not only rely on q-Risk scores.

Our results further illustrate that the use of DL-based probabilistic forecasting methods, especially those with sequence-to-sequence architecture, leads to low inference latency while consuming feasible computing resources in terms of model size and peak memory usage during inference.

Furthermore, the results confirm the superiority of probabilistic forecasting over point forecasting for use in safety monitoring.
This conclusion is drawn based on our empirical results and the fact that
point forecast predictions
correspond to the median ($q=0.5$) value of the probability distribution predicted by probabilistic forecasting methods~\cite{BENIDIS2022DL4TS}.
Our empirical results show that using values from the tail-end of the forecast probability distribution ($q\geq0.95$ in our case) leads to more accurate safety metric and safety violation predictions than predictions for $q=0.5$.

\paragraph{Window Configurations.}
We explored the effect of different combinations of varying hazard forecast horizons and context multipliers (\emph{window configurations}), on prediction accuracy (RQ\textsubscript{3}) and runtime performance (RQ\textsubscript{4}){, on the ACT case study only, due to the limited size of the dataset used for the ADS case study (as discussed in \autoref{sec:RQ1-method})}.
Given all the results discussed in \autoref{sec:RQ3-results} and \autoref{sec:RQ4-results}, we conclude that for {both $cte$ and $he$ safety requirements in }the ACT safety monitoring problem, TFT is the best model to be used for predicting imminent safety violations, i.e., $h=\SI{3}{s}$, for all $cm$ values.
We further suggest that high $cm$ values ($cm=9$) be used as they improve overall safety violation accuracy.
Nevertheless, as the peak inference memory of TFT increases with the increase in $cm$, a lower $cm$ might also be considered depending on the available GPU memory onboard the learning-enabled autonomous system.
The results for $h=\SI{12}{s}$ further highlight that, although DeepAR has superior prediction accuracy than other models, it is not as accurate as TFT for $h=\SI{3}{s}$. 
Furthermore, the high average inference latency of DeepAR at $h=\SI{12}{s}$ prohibits it from being used as a safety monitor of the learned component, as discussed in \autoref{sec:RQ4-results}.
{
However, if model optimizations
and specialized inference hardware can reduce the DeepAR's inference latency to an acceptable range, it can be considered for predicting longer horizon safety violations given its good prediction accuracy.
Nevertheless}, considering both \emph{accuracy} and \emph{inference latency}, using TFT on shorter forecast horizons than \SI{12}{s}, is a better option.

{
\paragraph{
Challenging Scenarios}

Although the trained safety metric forecasters yield high overall accuracy in predicting safety violations, 
it is important to identify the scenarios during which the safety monitor is more likely to mispredict safety violations.
Characterizing such scenarios will allow the developers to generate more relevant execution data which can be used to train the safety monitors further and increase their safety violation prediction accuracy.
Moreover, knowing the scenarios under which the safety monitor is expected to yield lower safety violation prediction accuracy would allow a system to be vigilant during the run-time of such scenarios and intervene in the automated operation of the system, if necessary. 

One potential method relies on fitting a regression tree~\cite{loh2011cart} to the safety violation prediction results (such as the ones provided in \autoref{sec:RQ2-act-results} and \autoref{sec:RQ2-ads-results}).
Concretely, a regression tree is fitted to a dataset whose features are the scenario parameters, and target variable is the F\textsubscript{3} score that the safety monitor yields for the corresponding scenario.
A notable benefit of a regression tree is that it allows the extraction of explainable rules that specify the part of the scenario space where the safety monitor yields a lower accuracy.

As an example, to explore the feasibility of explaining variation in safety violation prediction accuracy, we have fitted a regression tree to the safety violation prediction accuracy results of the safety monitor based on the TFT forecasting model, at the prediction quantile $q=0.995$, for the $cte$ safety requirement of the ACT case study (\autoref{sec:RQ2-act-results}).
Therefore, the features of the regression tree are the ACT scenario parameters, i.e., \textit{time of day}, \textit{cloud cover}, and \textit{starting $cte$ and $he$ of the aircraft}~\footnote{The detailed description and value ranges for the scenario parameters are provided in our replication package (\autoref{sec:data-availability}).}, while the target variable is the F\textsubscript{3} score of the TFT model at $q=0.995$.
Using grid search, we fitted regression trees by exploring combinations of values for its hyperparameters, i.e., \textit{maximum depth} and \textit{minimum number of samples per leaf node}.
We computed the average \textit{mean squared error} (MSE) for each model using 10-fold cross validation~\cite{fushiki2011kfold} and selected the most accurate tree, i.e., the one with the lowest average MSE over all ten folds
($MSE=\num{3e-4}$).
The computed measure of determination ($R^2$) for the most accurate model is $0.68$, indicating that most of the variance in F\textsubscript{3} is explained by the tree.
We have provided the dataset used to train the regression tree, its preprocessing details, the model selection and cross-validation script, as well as the selected regression tree (with detailed accuracy metrics) in our replication package (see \autoref{sec:data-availability}).

Based on the most accurate regression tree, we observe that the following three rules characterize part of the scenario space
where the safety monitor yields its lowest F\textsubscript{3} score:

\begin{itemize}
    \item $\textsf{time\_of\_day}\in\{\textsf{afternoon}\}
    \;
    \land 
    \;\textsf{cloud\_cover}\in\{\textsf{moderate}, \textsf{high}\}
    \implies
    F_3 = 0.952
    $
    \item $\textsf{time\_of\_day}\in\{\textsf{afternoon}\}
    \;
    \land
    \;\textsf{cloud\_cover}\in\{\textsf{none}, \textsf{low}\}
    \implies
    F_3 = 0.973
    $
    \item $\textsf{time\_of\_day}\in\{\textsf{morning}\}
    \;
    \land
    \;
    \textsf{he}_\textsf{start} \in [\SI{-10}{\degree}, \SI{-7.5}{\degree}]
    \implies
    F_3 = 0.984
    $
\end{itemize}

Given the above rules, we observe that time of day, cloud cover conditions and starting $he$ values are the most important features explaining variations in safety violation prediction accuracy across scenarios.
We further conclude, based on the above rules, that the lowest accuracy scores are observed during  
the afternoon,
when the sky is moderately or highly cloudy ($\textsf{cloud\_cover}\in\{\textsf{moderate}, \textsf{high}\}$).
As mentioned earlier, knowing the scenario subspace where prediction is challenging, specified by the above rules, can help the developer understand where more scenarios can be generated to re-train the safety monitor and potentially increase its prediction accuracy for low-accuracy scenarios.
Moreover, the user can take the uncertainty of the safety monitor predictions into account at runtime, e.g., by being vigilant during scenarios where safety violation prediction accuracy is low and intervening when necessary.

}

\subsection{Threats to Validity}\label{sec:threats}

In this section, we discuss potential threats to the validity of our study, namely internal, external, conclusion, and construct validity~\cite{val-1,val-2,val-3}.

\paragraph{Internal Validity}
Internal validity is concerned with the accuracy of the cause-and-effect relationships established by the experiments.
Due to the limitations of the GluonTS library at the time of our evaluation, we had to use models for evaluation that were implemented in different ML frameworks.
Concretely, DeepAR, MQCNN, and Seq2Seq models were implemented in MXNet while the TFT model was implemented in PyTorch, as the MXNet implementation of the TFT model was faulty and the MXNet models for MQCNN and Seq2Seq were not available.
The use of different ML frameworks could impact the internal validity of the results.
However, we conducted preliminary experiments on a model implemented in both frameworks\footnote{At the time of our evaluation, only similar DeepAR implementations were available in both frameworks.} to compare the impact of differences in ML framework on safety metric forecasting accuracy and runtime performance (memory and time overhead). 
We found that the differences in accuracy and runtime performance metrics were less than the standard deviation of the measurements and negligible.

\paragraph{External Validity}
External validity is concerned with the generalizability of our results.
One notable factor to consider is that in this study, we relied only on a specific ACT system (TinyTaxiNet) and simulation platform (X-Plane){, for the ACT case study, and relied on a specific ADS system (Dave-2) and a simulation platform (Udacity simulator).}
However, X-Plane is a widely used high-fidelity simulator, and TinyTaxiNet was the best open-source ACT available at the time of our evaluation.
Through preliminary experiments, we confirmed the superior $cte$ estimation accuracy of TinyTaxiNet against two other pre-trained models that were available on the NASA ULI X-Plane Simulator project repository on GitHub~\cite{NASA-ULI}.
{Regarding the ADS case study, Dave-2 is a widely used lane keeping ADS, and Udacity is a popular simulator used for closed-track simulation of ADS.}
Nonetheless, further studies involving other learning-enabled autonomous systems in aviation and autonomous driving, as well as other domains, such as autonomous agriculture, and manufacturing, are required.
We should however keep in mind that experiments such as the ones reported here entail substantial computations and extensive calendar time{, i.e., 7500+ hours of GPU computation which were performed over 42 calendar days, thanks to having access to multiple GPUs on the Digital Research Alliance of Canada compute clusters.
Another relevant factor is that due to X-Plane's capabilities, we were only able to set the weather statically, i.e., without sudden changes that rarely happen.
The same static weather has been used in the ADS simulation used by our study~\cite{Stocco2023ThirdEye}.
Nevertheless, both X-Plane and Udacity simulator are widely used high-fidelity simulators, as mentioned above.
Moreover, given that the maximum duration of a scenario execution, in both the ACT and ADS cases studies, are less than $\SI{4}{min}$ and $\SI{2}{min}$, respectively, assuming a static weather over the execution is not unreasonable.
An additional factor potentially impacting our proposed method's generalizability, is the fact that we assume that the monitored safety metrics are directly measurable (e.g., $cte_{act}$ can be measured directly using a GPS) or can be estimated during the system operation (e.g., Time-To-Collision or TTC in the case of autonomous driving is estimated based the relative distance and velocity between the ADS and the object in front of it~\cite{MINDERHOUD2001TTC}, which can be measured using a front radar on the ADS).
However, this is not a restrictive assumption as safety requirements, similar to any type of requirement, should have already been defined by the system developers and safety engineers, such that they are \emph{measurable}~\cite{ReqEngStandard}.
Therefore, the safety requirements are expected to rely on metrics that can be measured or estimated to assess their satisfaction or violation.
}
Another factor that could impact the generalizability of our results relates to the fact that we have not evaluated the performance results of our models on embedded hardware similar to the one that might be used during the operation of real ACT {and ADS }systems.
Nevertheless, our analysis revealed that the memory demand by the models is quite low and in line with what is reported in resource-constrained environments.
{Regarding the average inference latency measurements, model optimizations such as model quantization~\cite{roth2024embeddedNN}, can potentially improve the average inference latency of models, especially DeepAR such that its latency falls below runtime constraints.
However, we have performed our latency measurements using cloud-based NVIDIA V100 GPUs,
in an inference setting,
e.g., disabling gradient calculations~\cite{Pyt-InferMode, mxn-predictMode},
where
the
inference latency 
is expected to be comparable to or lower than
that of embedded GPUs.
}

\paragraph{Conclusion Validity}
Conclusion validity relates to the conclusions that can be drawn from the collected data and their statistical significance.
We followed the widely accepted rule-of-thumb of 30 repetitions for the experiments and we report every statistical value with its confidence interval.

\paragraph{Construct Validity}
Construct validity is concerned with the degree to which the measured variables in the study represent the underlying concept being studied.
As discussed in \autoref{sec:RQ1-method}, q-Risk is a widely used metric to measure the accuracy of time series predictions (forecast safety metric values in our case) for a specific prediction quantile~\cite{LIM2021TFT, SALINAS2020DeepAR, lim2021time, BENIDIS2022DL4TS}, since it provides a summation of quantile loss (QL) over the forecast horizon for all predictions, normalized over all samples in the test set.
As discussed in \autoref{sec:RQ2-method}, we have used precision and recall which are widely used as metrics to capture the accuracy of the models in terms of missed safety violations and false prediction, respectively.
Furthermore, similar to \cite{Stocco2023ThirdEye}, we have used F\textsubscript{3} score as an aggregate metric to compare the overall safety violation prediction accuracy of safety monitors while capturing the relative importance of false negatives and false positives.
As discussed in \autoref{sec:RQ4-method}, the performance overhead introduced by the safety monitor can be refined to space and time overheads.
Since models are loaded in the GPU, GPU memory usage is a metric that successfully captures the space overhead of the models as model size and peak memory usage during inference.
Whereas, average prediction latency effectively captures the time overhead of the model at runtime.

\subsection{Data Availability}\label{sec:data-availability}

The evaluated DL-based probabilistic forecasting models have been implemented in Python.
{
We made the aforementioned implementations,
the instructions to set up the ACT case study,
the detailed description of the scenario parameters used to generate data,
the generated ACT dataset,
the raw and preprocessed ADS dataset,
and the detailed evaluation results, for both the ACT and ADS case studies,}
available in our replication package~\cite{replication_package}.

\section{Conclusion and Future Work}\label{sec:conclusion}

In this paper, we proposed a method for safety monitoring of learned components in autonomous systems via probabilistic safety metric forecasting.
We address the practical challenges of lacking access to internal information of the learned component and the system having limited operational resources, by using state-of-the-art DL-based probabilistic time series forecasters. They rely on scenarios and learned component output values to provide predictions of the safety metric probability distribution with acceptable inference latency and memory usage.
{We apply these forecasters to widely used case studies in autonomous aviation and autonomous driving, namely ACT and lane keeping ADS, respectively,}
where we run extensive experiments to evaluate the safety metric and violation prediction accuracy, inference latency, and computation resource usage of state-of-the-art models, with a varying lookback and hazard forecast horizons while comparing them against a very competitive baseline (DeepAR).
Our evaluation results suggest that probabilistic forecasting of safety metrics, given learned component outputs and scenarios, is effective for safety monitoring.
Moreover, the evaluation results show that using Temporal Fusion Transformer (TFT) for predicting imminent safety violations ($h=\SI{3}{s}$), for all lookback horizons, leads to the most accurate predictions with acceptable inference latency while consuming reasonable computational resources.

As part of future work, we plan to apply our proposed safety monitoring method to other learning-enabled autonomous systems in various domains such as automated driving, agriculture, and manufacturing.
Furthermore, we plan to investigate whether using search-based methods that identify the hazard boundary of a learned component, e.g., MLCSHE~\cite{sharifi2023mlcshe}, reduces the size of the dataset required to train an accurate safety monitor.
{
Finally, in the future, we plan to further investigate the impact of re-training the safety monitor, using additional scenario generated according the regression tree analysis results (\autoref{sec:discussion}), on its safety violation prediction accuracy. 
}

\begin{acks}
We thank Corina P{\u{a}}s{\u{a}}reanu for her feedback in the early stages of the work and her pointer to the TinyTaxiNet model.
We would also like to thank Nathan Aschbacher and Frederic Risacher for their constructive feedback on the work. 
This work was partially supported by funding from the Natural Sciences and Engineering Research Council of Canada (NSERC), through the Canada Research Chairs and discovery programs, Ontario Graduate Scholarship, Mitacs Accelerate Program, and Auxon Corporation.
{This research was enabled in part by support provided by British Columbia Digital Research Infrastructure (\url{https://www.bc.net}), Compute Ontario (\url{https://www.computeontario.ca}), and the Digital Research Alliance of Canada (\url{https://alliancecan.ca}). Lionel Briand was partly funded by the Science Foundation Ireland grant 13/RC/2094-2. Andrea Stocco was supported by the Bavarian Ministry of Economic Affairs, Regional Development, and Energy.
}
\end{acks}

\bibliographystyle{ACM-Reference-Format}
\bibliography{references}

\end{document}